\documentclass[11pt]{article}

\usepackage[preprint]{acl}

\usepackage{times}
\usepackage{latexsym}

\usepackage[T1]{fontenc}
\usepackage[utf8]{inputenc}

\usepackage{microtype}
\usepackage{inconsolata}

\usepackage{graphicx}

\usepackage{amsmath, amssymb}
\usepackage{mathtools}
\usepackage{xcolor}
\usepackage{enumitem}
\usepackage{caption}
\usepackage{subcaption}
\usepackage{wrapfig}
\usepackage{cleveref}
\usepackage{tcolorbox}
\tcbuselibrary{skins,breakable,listings}

\usepackage{listings}
\usepackage{multirow}
\usepackage{booktabs}
\usepackage{tcolorbox}
\usepackage{makecell}
\usepackage[table]{xcolor}

\usepackage{xspace} % for safe spacing after macros
\usepackage{CJKutf8}

\newcommand{\sparagraph}[1]{\vspace{3pt}\par\noindent\textbf{#1}}

\title{Why Do Multilingual Reasoning Gaps Emerge \\in Reasoning Language Models? }

\author{
\makecell{Deokhyung Kang\textsuperscript{1},
Seonjeong Hwang\textsuperscript{1},
Daehui Kim\textsuperscript{2},
\\Hyounghun Kim\textsuperscript{1,3},
Gary Geunbae Lee\textsuperscript{1,3}}
\\
\textsuperscript{1}Graduate School of Artificial Intelligence, POSTECH\\
\textsuperscript{2}Agentic AI Lab, KT\\
\textsuperscript{3}Department of Computer Science and Engineering, POSTECH\\
\texttt{\small\{deokhk, seonjeongh, h.kim, gblee\}@postech.ac.kr, daehui.kim@kt.com}
}

\begin{document}
\maketitle
\begin{abstract}
Reasoning language models (RLMs) achieve strong performance on complex reasoning tasks, yet they still exhibit a \textit{multilingual reasoning gap}, performing better in high-resource languages than in low-resource ones. While recent efforts have been made to address this gap, its underlying causes remain largely unexplored. In this work, we show that this gap primarily stems from failures in language \textit{understanding}—specifically, the model’s inability to translate multilingual inputs into the language dominating its reasoning traces (typically English). As identifying understanding failures can enable targeted mitigation of the gap, we evaluate a range of detection methods and find that understanding failures are detectable to a meaningful extent, with supervised approaches performing best. Building on this, we propose \textit{Selective Translation}, a strategy that incorporates an English translation into the initial reasoning trace when an understanding failure is detected. Experimental results using Qwen3-4B show that Selective Translation substantially bridges the multilingual reasoning gap, achieving near full-translation performance while translating only about 20\% of inputs. Together, our results show that failures in language understanding are the primary driver of the multilingual reasoning gap and can be detected and selectively mitigated, clarifying its origin and suggesting a path toward more equitable multilingual reasoning.\footnote{Our code and data are publicly available at \href{https://github.com/deokhk/RLM_analysis}{here}.}

\end{abstract}

\section{Introduction}
Recent reasoning language models (\textbf{RLMs}), such as OpenAI’s o1~\cite{jaech2024openai} and DeepSeek-R1~\cite{guo2025deepseek}, have achieved remarkable performance on complex reasoning tasks. By generating long reasoning traces (intermediate steps) before producing a final response, they substantially outperform conventional large language models~\cite{muennighoff2025s1simpletesttimescaling, yu2025dapo}. Despite these advances, RLMs still exhibit a \textit{multilingual reasoning gap}, performing much better on queries in high-resource languages (e.g., English) than in low-resource languages~\cite{wang2025polymath}. While recent studies~\cite{zhao2025less, bajpai2025multilingual, yoo2025code} have explored various methods to narrow this gap, a systematic understanding of \textbf{why these gaps arise in RLMs} remains unexplored — a key question for developing principled approaches toward equitable multilingual reasoning.

\begin{figure}[!t]
  \centering
  \includegraphics[width=\columnwidth]{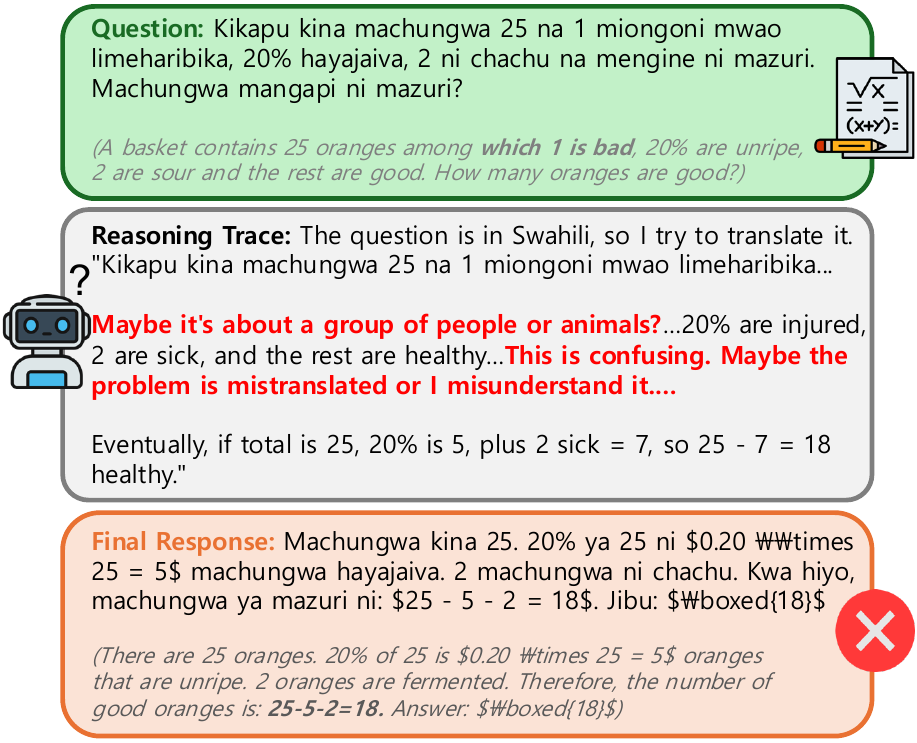}
  \caption{Understanding failure in Qwen3-4B: the model shows confusion when interpreting the Swahili input (e.g., “This is confusing…”) and ignores the “1 bad orange” condition, leading to an incorrect answer.}
  \label{fig:intro}
\end{figure}

In this work, we investigate this question by framing RLMs’ multilingual reasoning as a three-stage process, grounded in observations from prior works~\cite{yong2025crosslingual, qi2025models, tam2025language}. As shown in Figure~\ref{fig:intro}, when given a math problem in Swahili, the model first translates the input meaning into the dominant language of its reasoning trace (English), a stage we refer to as (i) language \textit{understanding}. It then (ii) \textit{reasons} within that English-centric trace before (iii) \textit{generating} the final response back in Swahili. Building on this framework, we quantify the contribution of failures in each stage to the multilingual reasoning gap through a stage-wise attribution analysis, using stage-specific interventions designed to control for such failures. We evaluate multiple open-source RLMs, including the Qwen3 family~\cite{yang2025qwen3} and gpt-oss-20b~\cite{agarwal2025gpt}, across 10 languages ranging from high- to low-resource settings. 
Our analysis reveals a consistent pattern: \textbf{understanding failures are the dominant source of the multilingual reasoning gap} in most cases. This finding suggests that if such failures can be detected, we can prevent them by abstaining from answering or mitigate the gap by translating the input into a high-resource language before reasoning.

Accordingly, we next investigate \textbf{whether such understanding failures can be detected}. We observe that models often leave recognizable signals of misunderstanding within their reasoning traces (Figure~\ref{fig:intro}). Inspired by this, we adapt detection methods originally developed for identifying undesirable behaviors such as hallucination and jailbreak, which similarly arise when the model fails to satisfy its intended objective. 
Using various detection methods—ranging from LLM-based monitoring~\cite{baker2025monitoring} and self-reflection~\cite{xiong2024can} to token-probability signals~\cite{manakul-etal-2023-selfcheckgpt} and supervised detectors~\cite{azaria2023internal, chan2025can}—we find that understanding failures are detectable across multiple approaches, with supervised approaches performing best.

Finally, we propose \textbf{Selective Translation}, which incorporates an English translation into the initial reasoning trace only when an understanding failure is detected. It efficiently narrows the multilingual reasoning gap, improving average accuracy from 81.1 to 88.0 on the Polymath-Low~\cite{wang2025polymath} benchmark with Qwen3-4B~\cite{yang2025qwen3}, closely matching full translation (89.4) while translating only 20\% of inputs. These gains, achieved while intervening on only a small fraction of inputs, demonstrate the reliability of the detector and the promise of understanding failure detection for equitable multilingual reasoning.

We summarize our main contributions as follows:
\textbf{(i)} We provide the first systematic analysis of the multilingual reasoning gap in RLMs and show that understanding failures are the dominant source of this gap, providing a basis for developing principled approaches toward equitable multilingual reasoning.
\textbf{(ii)} We systematically evaluate a range of methods for understanding failure detection and show that such failures can be automatically detected.
\textbf{(iii)} We demonstrate the effectiveness of understanding failure detection via Selective Translation in efficiently narrowing multilingual reasoning gaps.

\section{Related Work}
\paragraph{Reasoning language models (RLMs).}
Recent RLMs such as DeepSeek-R1~\citep{guo2025deepseek} and Qwen3~\citep{yang2025qwen3} are trained to produce long chains of thought and achieve strong performance on challenging reasoning tasks~\citep{muennighoff2025s1simpletesttimescaling,jaech2024openai}. Recently, their multilingual reasoning capabilities have gained increasing attention. \citet{yong2025crosslingual} shows that scaling inference compute on an English-centric RLM yields clear multilingual gains, contrary to earlier findings on smaller models~\citep{son-etal-2025-linguistic}. Prior work~\citep{yong2025crosslingual,tam2025language} also finds that RLMs tend to produce reasoning traces dominated by high-resource languages, which we refer to as the reasoning language. \citet{park2025cross} attributes this to cross-lingual collapse, where GRPO~\citep{shao2024deepseekmath} training encourages models to revert to English-dominant reasoning traces due to accuracy-centric reward signals. In response, several studies have explored controlling the reasoning language via language-forcing prefixes~\citep{yong2025crosslingual,qi2025models,tam2025language} or language-consistency rewards~\citep{park2025cross}. However, these approaches often degrade accuracy or require costly target-language reasoning data, particularly for low-resource languages. Therefore, we focus on a setting where the model reasons in a high-resource language (English in our case). 
\paragraph{Multilingual reasoning gap.}
Despite this, RLMs still exhibit a multilingual reasoning gap~\cite{yang2025qwen3}. Recent work has attempted to bridge this gap through approaches such as representation editing~\cite{zhao2025less}, tailored prompting~\cite{yoo2025code}, and prefix tuning~\cite{bajpai2025multilingual}. However, the underlying causes of this gap remain largely unexplored. Therefore, we tackle this question by systematically identifying the sources of the multilingual reasoning gap and investigating methods to mitigate them. Appendix~\ref{sec:appendix_additional_related_work} provides an extended discussion on multilingual reasoning and abstention to further contextualize our work within the broader literature.
\section{Why Does the Multilingual Reasoning Gap Emerge?}\label{sec:gap_analysis}
\subsection{Multilingual Reasoning Process}
\label{sec:mf_process}

\begin{figure}[!t]
    \centering
    \includegraphics[width=\columnwidth]{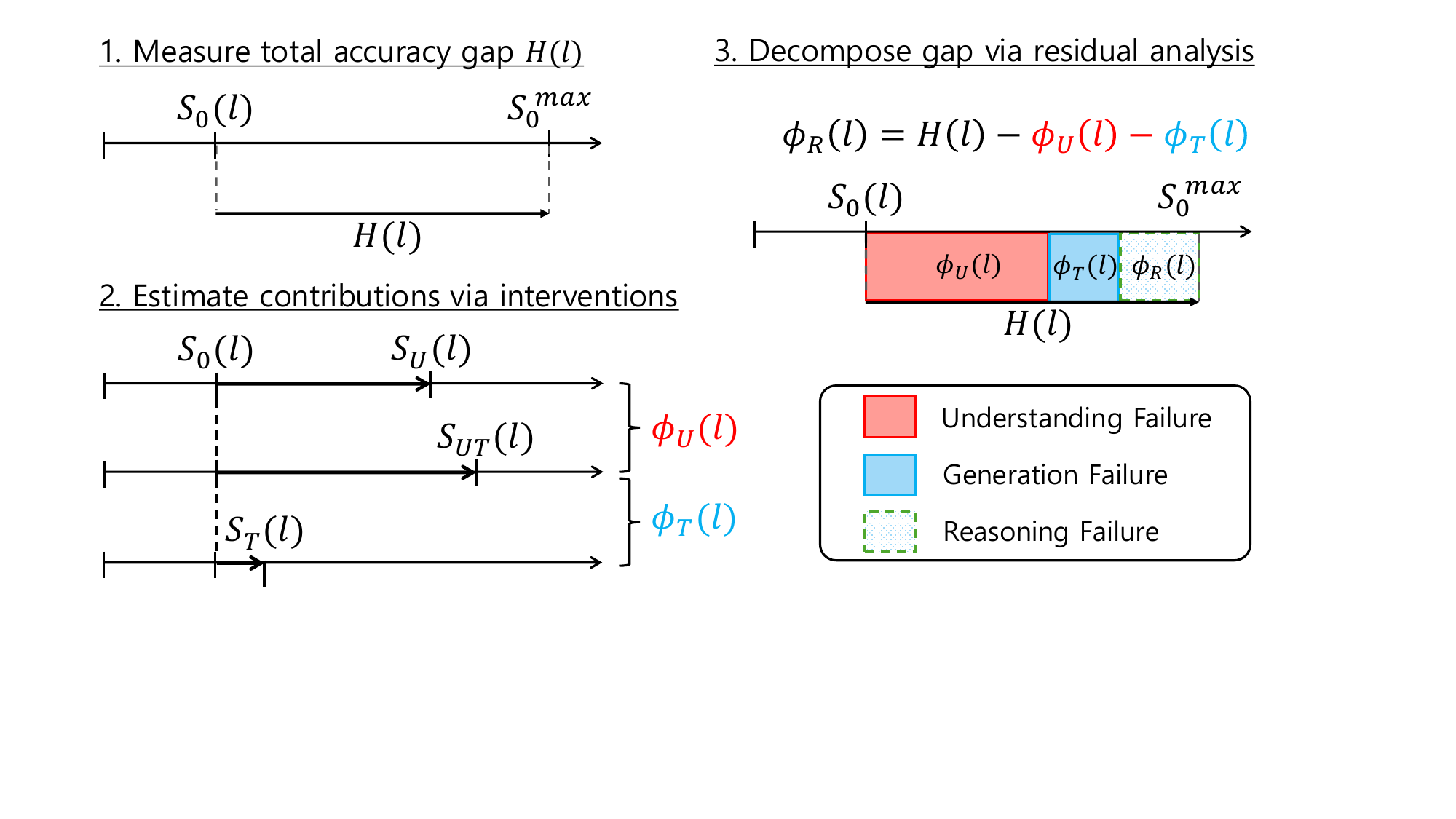}
    \caption{
    Stage-wise attribution of the multilingual reasoning gap. 
    The total gap $H(l)$ is decomposed into contributions from understanding ($\phi_U$), generation ($\phi_T$), and reasoning ($\phi_R$) using targeted interventions followed by residual analysis.
    }
    \label{fig:attribution_intuition}
\end{figure}

To analyze why the multilingual reasoning gap emerges, we draw on prior findings on how RLMs process multilingual inputs. Prior work shows that reasoning traces are typically dominated by high-resource languages such as English or Chinese, while final responses tend to align with the input language~\citep{yong2025crosslingual, tam2025language}. During the early stages of reasoning, models often begin by internally translating the input into the dominant language of their reasoning traces~\cite{yong2025crosslingual}. We interpret \textit{this translation as the model’s language \textbf{understanding}} process, which enables subsequent reasoning to be carried out in the dominant language. Building on this view, we conceptualize multilingual reasoning as a three-stage process: (i) \textbf{understanding} the input, (ii) \textbf{reasoning} in the dominant language, and (iii) \textbf{generating} the final response in the input language. We hypothesize that failures may arise at each stage (e.g., input misinterpretation, degraded reasoning despite successful understanding, or generation errors), thereby contributing to the multilingual reasoning gap.

\subsection{Stage-wise Attribution Analysis}\label{sec:method_gap}

We perform a stage-wise attribution analysis to identify which stages drive the multilingual reasoning gap. We use targeted interventions for the understanding and generation stages and measure how much the gap is reduced when failures at each stage are controlled for. We then attribute the remaining gap to reasoning stage, as directly controlling reasoning failures is difficult. Figure~\ref{fig:attribution_intuition} illustrates the overall process.

\sparagraph{Preliminary.}
Given an input $x=(x_{1},\dots,x_{n})$, an RLM first generates a reasoning trace $r=(r_{1},\dots,r_{k})$ conditioned on $x$, and then produces a final response $y=(y_{1},\dots,y_{m})$ based on both $x$ and $r$. The reasoning trace $r$ is a long chain of thought that includes intermediate steps and often a candidate final answer~\citep{guo2025deepseek, chen2025towards}; in practice, it typically begins with \texttt{<think>} and ends with \texttt{</think>}.

\sparagraph{Step 1: Measuring the Total Gap.}
For language $l$, let $S_0(l)$ denote the RLM’s accuracy under the \textbf{Base} setting, i.e., the original model evaluated on the input language without any intervention. Let $S_0^{\max} = \max_{l \in \mathcal{L}} S_0(l)$ be the best performance across all languages. We define the multilingual reasoning gap for language $l$ as:
\[
H(l) = S_0^{\max} - S_0(l).
\]
\sparagraph{Step 2: Estimating Contributions via Interventions.}
To attribute the sources of the gap, we introduce two interventions:

\sparagraph{(1) Understanding Intervention (U).}
As described in Section~\ref{sec:mf_process}, during the understanding stage, an RLM translates the meaning of the input $x_{l}$ into a dominant language within its reasoning trace~$r$. To assess failures in this stage, we introduce the \textit{understanding intervention}, which provides an explicit translation of $x_{l}$ via a fixed prefix $\pi(x_{\mathrm{dom}})$ at the beginning of the reasoning trace:
\begin{tcolorbox}
[colback=gray!5!white,colframe=black!75!black,width=\columnwidth,boxrule=0.6pt,arc=2pt]
\normalsize
\texttt{$\pi$}($x_{dom}$) = "Okay, let's see. I understand the
question as: '\{$x_{dom}$\}'. Let's solve the
problem based on this understanding."
\vspace{-2pt}
\end{tcolorbox}
Here, $x_{\mathrm{dom}}$ denotes a \textbf{dataset-provided reference (gold) English translation} of $x_l$. We use the English translation to align with the model’s dominant reasoning language\footnote{In our experiments, reasoning traces are predominantly in English; see Appendix~\ref{appendix:language_distribution}.},
and a dataset-provided reference translation to ensure that the intervention is not affected by translation noise. The model then generates $r$ conditioned on $(x_{l},\pi(x_{\mathrm{dom}}))$ instead of only $x_{l}$. The performance gain quantifies the extent of \textit{understanding failures}.

\sparagraph{(2) Answer Extraction from Reasoning Trace (T).}
To assess failures in response generation, we extract the final answer directly from the reasoning trace $r$ instead of the final response $y$, thereby isolating errors introduced during the \textit{generation} stage. We apply the same answer extraction logic to both $r$ and $y$, as detailed in Appendix~\ref{appendix:eval_details}.

Let $S_U(l)$, $S_T(l)$, and $S_{UT}(l)$ denote the accuracies under \textbf{w/ U}, \textbf{w/ T}, and \textbf{w/ U+T}, respectively. Each intervention explains part of the gap through the performance gain it induces over the \textbf{Base} setting. Because the effects of U and T may interact, we attribute their contributions using a Shapley decomposition~\cite{shorrocks2013decomposition}, a principled and order-invariant approach for stage-wise attribution:
\begin{equation*}
\small
\begin{aligned}
\phi_U(l)
&= \max\Big\{0, \tfrac{1}{2}\big[(S_U(l)-S_0(l)) + (S_{UT}(l)-S_T(l))\big]\Big\}, \\
\phi_T(l)
&= \max\Big\{0, \tfrac{1}{2}\big[(S_T(l)-S_0(l)) + (S_{UT}(l)-S_U(l))\big]\Big\}.
\end{aligned}
\end{equation*}
\sparagraph{Step 3: Decomposing the Gap via Residual Analysis.}
After estimating the contributions of understanding and generation, we decompose the remaining gap via residual analysis. Specifically, we attribute the unexplained portion of the gap to reasoning:
\[
\phi_R(l) = H(l) - \phi_U(l) - \phi_T(l) \quad (\ge 0).
\]
To analyze the relative importance of each stage, we normalize these contributions to obtain stage-wise shares:
\[
\begin{aligned}
\text{U-share}(l) &= \tfrac{\phi_U(l)}{H(l)} \quad (\text{Understanding}) \\
\text{R-share}(l) &= \tfrac{\phi_R(l)}{H(l)} \quad (\text{Reasoning}) \\
\text{G-share}(l) &= \tfrac{\phi_T(l)}{H(l)} \quad (\text{Generation})
\end{aligned}
\]
with U-share($l$) $+$ R-share($l$) $+$ G-share($l$) $=1$.
\sparagraph{Aggregation Across Languages.}
To obtain a dataset-level view of stage contributions, we aggregate these shares across languages. Since the magnitude of the gap varies by language, we weight each language by its gap $H(l)$ (i.e., its headroom), thereby assigning greater weight to languages with larger performance gaps:

\begin{align*}
\text{Weighted-Share}_{\star} 
&= \frac{\sum_{l \in \mathcal{L}} H(l) \cdot \text{Share}_{\star}(l)}
        {\sum_{l \in \mathcal{L}} H(l)}, \\
&\quad \star \in \{\text{U}, \text{R}, \text{G}\}.
\end{align*}
Finally, to focus on statistically meaningful differences across languages, we only consider languages whose \textbf{Base} performance is significantly lower than that of the best-performing language (e.g., English) according to a Welch’s t-test (p<0.05)~\cite{welch1947generalization}.

\subsection{Experimental Settings}\label{sec:gap_setup}
\paragraph{Models.} We primarily evaluate two recent RLMs from distinct model families: Qwen3-4B~\cite{yang2025qwen3} and gpt-oss-20b~\cite{agarwal2025gpt}. 
We select these models as they are publicly available, achieve state-of-the-art performance at a comparable scale, and support multiple languages. 
To examine generalizability across model scales, we also evaluate Qwen3-1.7B, 8B, and 14B.

\paragraph{Evaluation datasets.}
We evaluate models on two multilingual reasoning benchmarks: \textbf{Polymath}~\cite{wang2025polymath} and \textbf{MMLU-ProX-Lite}~\cite{xuanmmluprox}. Polymath spans mathematical reasoning across different difficulty levels, and we focus on the low, medium, and high levels, ranging from K-12 mathematics to challenging competition problems from AIME~\cite{aops_aime}. For STEM reasoning, we use STEM-related categories from MMLU-ProX-Lite, a subset from MMLU-ProX for efficient evaluation. We evaluate models on a set of typologically diverse languages with varying resource availability, consistently across both benchmarks. 
Following the taxonomy of~\citet{joshi-etal-2020-state}, we group them into:
\begin{itemize}
    \item \textbf{High-resource languages}: English (en), German (de), Spanish (es), Arabic (ar), Japanese (ja), Korean (ko)
    \item \textbf{Mid-resource languages}: Thai (th), Bengali (bn)
    \item \textbf{Low-resource languages}: Swahili (sw), Telugu (te)
\end{itemize}
This yields 125 test samples per language for each difficulty level on Polymath and 257 samples per language on MMLU-ProX-Lite. See Appendix~\ref{appendix:dataset_details} for detailed dataset descriptions and evaluation prompts.
\paragraph{Evaluation metrics and settings.}
We report task accuracy averaged over three runs with different random seeds, sampled with temperature~$=0.6$, top-$p=0.95$, and top-$k=20$, and a maximum generation length of 32,768 tokens.
Correctness is evaluated using \textsc{math-verify}~\cite{kydlicek2025mathverify} for Polymath and string-based matching for MMLU-ProX-Lite. Further evaluation details are provided in Appendix~\ref{appendix:eval_details}.
\subsection{Results}
\begin{table*}[!t]
\centering
\footnotesize
\setlength{\tabcolsep}{7pt}
\renewcommand{\arraystretch}{0.7}
\begin{tabular}{l|l|cccccc|cc|cc|c}
\toprule
Dataset & Method & en & de & es & ar & ja & ko & th & bn & sw & te & \textbf{Avg} \\
\midrule
\multirow{4}{*}{Polymath-Low}
 & Base   & 96.5 & 88.0 & 93.9 & 89.6 & 85.3 & 90.7 & 85.1 & 83.2 & 29.3 & 69.9 & \textbf{81.1} \\
 & w/ T   & 96.0 & 87.5 & 93.3 & 89.3 & 85.1 & 89.6 & 85.1 & 82.9 & 31.7 & 70.9 & \textbf{81.1} \\
 & w/ U   & 95.2 & \cellcolor[rgb]{0.88,0.95,0.88}89.6 & 94.7 & \cellcolor[rgb]{0.75,0.88,0.98}92.5 & \cellcolor[rgb]{0.75,0.88,0.98}91.5 & \cellcolor[rgb]{0.88,0.95,0.88}93.1 & \cellcolor[rgb]{0.75,0.88,0.98}92.0 & \cellcolor[rgb]{0.75,0.88,0.98}94.4 & \cellcolor[rgb]{0.75,0.88,0.98}88.0 & \cellcolor[rgb]{0.75,0.88,0.98}87.7 & \cellcolor[rgb]{0.75,0.88,0.98}\textbf{91.9} \\
 & w/ U+T & 95.2 & \cellcolor[rgb]{0.88,0.95,0.88}89.6 & 94.4 & \cellcolor[rgb]{0.88,0.95,0.88}92.3 & \cellcolor[rgb]{0.75,0.88,0.98}91.2 & \cellcolor[rgb]{0.88,0.95,0.88}92.8 & \cellcolor[rgb]{0.75,0.88,0.98}92.3 & \cellcolor[rgb]{0.75,0.88,0.98}94.4 & \cellcolor[rgb]{0.75,0.88,0.98}90.1 & \cellcolor[rgb]{0.75,0.88,0.98}89.3 & \cellcolor[rgb]{0.75,0.88,0.98}\textbf{92.2} \\
\midrule
\multirow{4}{*}{Polymath-Medium}
 & Base   & 74.7 & 70.9 & 72.0 & 68.8 & 70.9 & 70.9 & 67.7 & 64.8 & 45.6 & 64.8 & \textbf{67.1} \\
 & w/ T   & 75.2 & 70.9 & 71.7 & 69.1 & 70.7 & 70.9 & 67.7 & 65.6 & 45.9 & 65.1 & \textbf{67.3} \\
 & w/ U   & 74.7 & 72.0 & 74.1 & \cellcolor[rgb]{0.88,0.95,0.88}73.3 & \cellcolor[rgb]{0.88,0.95,0.88}75.2 & 73.6 & 70.4 & 68.5 & \cellcolor[rgb]{0.75,0.88,0.98}67.5 & \cellcolor[rgb]{0.75,0.88,0.98}70.9 & \cellcolor[rgb]{0.75,0.88,0.98}\textbf{72.0} \\
 & w/ U+T & 74.7 & 72.0 & 74.4 & \cellcolor[rgb]{0.75,0.88,0.98}73.3 & \cellcolor[rgb]{0.75,0.88,0.98}75.5 & 73.3 & 70.4 & \cellcolor[rgb]{0.88,0.95,0.88}70.4 & \cellcolor[rgb]{0.75,0.88,0.98}68.8 & \cellcolor[rgb]{0.75,0.88,0.98}71.2 & \cellcolor[rgb]{0.75,0.88,0.98}\textbf{72.4} \\
\midrule
\multirow{4}{*}{Polymath-High}
 & Base   & 53.9 & 51.2 & 48.8 & 52.0 & 50.1 & 52.8 & 45.3 & 44.8 & 28.8 & 40.8 & \textbf{46.9} \\
 & w/ T   & 54.1 & 50.9 & 49.6 & 52.5 & 50.4 & 53.6 & 45.6 & 45.6 & 29.3 & 40.8 & \textbf{47.3} \\
 & w/ U   & 54.9 & 51.2 & 52.0 & 51.7 & 52.5 & 53.3 & \cellcolor[rgb]{0.88,0.95,0.88}50.9 & \cellcolor[rgb]{0.88,0.95,0.88}49.1 & \cellcolor[rgb]{0.75,0.88,0.98}50.4 & \cellcolor[rgb]{0.75,0.88,0.98}50.1 & \cellcolor[rgb]{0.75,0.88,0.98}\textbf{51.6} \\
 & w/ U+T & 54.9 & 51.7 & 52.3 & 51.7 & 52.5 & 53.1 & \cellcolor[rgb]{0.88,0.95,0.88}51.5 & \cellcolor[rgb]{0.88,0.95,0.88}49.6 & \cellcolor[rgb]{0.75,0.88,0.98}50.7 & \cellcolor[rgb]{0.75,0.88,0.98}50.1 & \cellcolor[rgb]{0.75,0.88,0.98}\textbf{51.8} \\
\midrule
\multirow{4}{*}{MMLU-ProX-Lite}
 & Base   & 77.0 & 77.8 & 76.5 & 73.3 & 75.0 & 74.6 & 73.9 & 74.6 & 53.6 & 71.1 & \textbf{72.7} \\
 & w/ T   & 76.9 & 77.4 & 76.5 & 73.3 & 74.8 & 74.4 & 73.8 & 74.6 & 53.4 & 71.1 & \textbf{72.6} \\
 & w/ U   & \cellcolor[rgb]{0.88,0.95,0.88}78.5 & 77.4 & 77.7 & \cellcolor[rgb]{0.75,0.88,0.98}78.0 & \cellcolor[rgb]{0.75,0.88,0.98}78.5 & 77.2 & \cellcolor[rgb]{0.75,0.88,0.98}78.1 & \cellcolor[rgb]{0.75,0.88,0.98}78.2 & \cellcolor[rgb]{0.75,0.88,0.98}73.5 & \cellcolor[rgb]{0.75,0.88,0.98}77.2 & \cellcolor[rgb]{0.75,0.88,0.98}\textbf{77.4} \\
 & w/ U+T & 78.1 & 77.3 & 78.0 & \cellcolor[rgb]{0.75,0.88,0.98}78.0 & \cellcolor[rgb]{0.75,0.88,0.98}78.5 & \cellcolor[rgb]{0.88,0.95,0.88}77.7 & \cellcolor[rgb]{0.75,0.88,0.98}78.2 & \cellcolor[rgb]{0.75,0.88,0.98}78.2 & \cellcolor[rgb]{0.75,0.88,0.98}74.7 & \cellcolor[rgb]{0.75,0.88,0.98}77.4 & \cellcolor[rgb]{0.75,0.88,0.98}\textbf{77.6} \\
\bottomrule
\end{tabular}
\caption{
Accuracy comparison of methods across languages on Qwen3-4B.
\textbf{Base} denotes the original model without intervention, \textbf{w/ T} applies Answer Extraction from Reasoning Trace, \textbf{w/ U} applies Understanding Intervention, and \textbf{w/ U + T} applies both. Pairwise comparisons against the Base use Welch’s t-test:\colorbox[rgb]{0.75,0.88,0.98}{blue} cells mark statistically significant improvements (p < 0.05), and \colorbox[rgb]{0.88,0.95,0.88}{green} cells indicate notable improvements (p < 0.1). The Average row is \textbf{bolded}. Understanding intervention significantly improves performance, especially for low-resource languages.}
\label{tab:main_result}
\end{table*}

\sparagraph{Understanding intervention significantly improves performance.}
We first examine how each intervention affects task accuracy. Table~\ref{tab:main_result} reports multilingual benchmark results on Qwen3-4B with and without the proposed interventions. The Understanding Intervention (\textbf{w/ U}) yields consistent and significant improvements, especially in low-resource languages such as Swahili (sw) and Telugu (te). For example, in Polymath-Low, it improves Swahili from \(29.3 \to 88.0\).
By contrast, the Answer Extraction from Reasoning Trace (\textbf{w/ T}) shows little change from the \textbf{Base}. These results suggest that the multilingual gap in RLMs mainly arises from failures in \textit{Understanding}. To verify this more systematically, we perform the stage-wise attribution analysis (Section~\ref{sec:method_gap}) to quantify the contribution of each stage.
\begin{figure*}[!th]
  \centering
  \includegraphics[width=\textwidth]{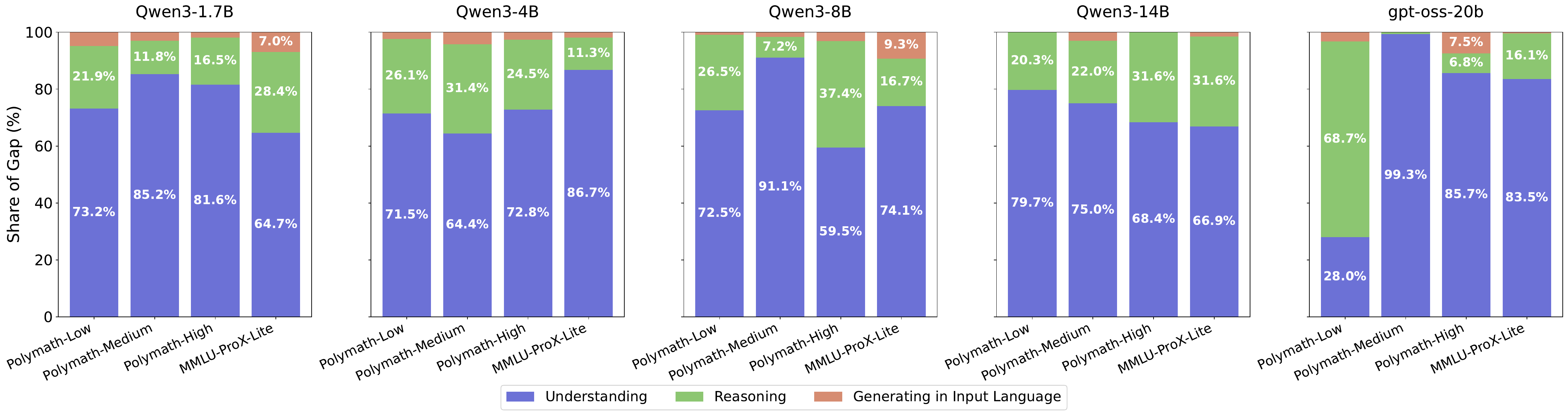}
  \caption{Weighted shares of \textit{Understanding}, \textit{Reasoning}, and \textit{Generation} in the input language to the overall multilingual reasoning gap. Across models and datasets, failures in \textit{Understanding} generally dominate the gap.}
  \label{fig:gap-agg}
\end{figure*}
\sparagraph{Understanding failures dominate the multilingual reasoning gap, regardless of reasoning difficulty.}
\begin{table}[t]
\centering
\resizebox{\linewidth}{!}{%
\begin{tabular}{l|cc|cc}
\toprule
\multirow{2}{*}{\textbf{Dataset}} & \multicolumn{2}{c}{\textbf{Qwen3-4B}} & \multicolumn{2}{c}{\textbf{gpt-oss-20b}} \\
\cmidrule(lr){2-3}\cmidrule(lr){4-5}
 & Base & w / U & Base & w / U \\
\midrule
Low    & $0.82_{\pm 0.21}$ & $0.95_{\pm 0.03}$ & $0.91_{\pm 0.05}$ & $0.94_{\pm 0.03}$ \\
Medium & $0.89_{\pm 0.11}$ & $0.96_{\pm 0.04}$ & $0.97_{\pm 0.04}$ & $0.99_{\pm 0.02}$ \\
High   & $0.85_{\pm 0.14}$ & $0.95_{\pm 0.02}$ & $0.92_{\pm 0.05}$ & $0.98_{\pm 0.03}$ \\
\bottomrule
\end{tabular}}
\caption{Average reasoning performance ratio (mean\textsubscript{$\pm$SD}) across Polymath splits of different difficulty (Low, Medium, High).}
\vspace{-12pt}
\label{tab:intv_effect_ratio}
\end{table}

Figure~\ref{fig:gap-agg} presents the results of the stage-wise attribution analysis on the multilingual reasoning gap. Failures in the \textit{understanding} stage account for most of the gap, while the \textit{generation} stage contributes only marginally.\footnote{ Appendix~\ref{sec:appendix_prefix_variants} further shows that our findings generalize across different understanding prefix variants.} The language-specific attribution analysis (Appendix~\ref{sec:appendix_language_gap}) reveals that this trend is especially pronounced in low-resource languages, while the impact of understanding failures is reduced in high-resource languages. However, since the multilingual reasoning gap is largely driven by low-resource languages, \textit{understanding} remains the primary bottleneck overall. Moreover, the share of \textit{Reasoning} remains relatively small and shows no consistent trend across different reasoning difficulty levels (Polymath-Low, Medium, and High).

To further validate this findings, we assess whether reasoning difficulty continues to affect performance once understanding failures are resolved. 
Specifically, we compute the average \textbf{Reasoning Performance Ratio}—the average ratio of each language’s accuracy to that of the best-performing language under the \textbf{Base} setting—before and after applying the Understanding Intervention (\textbf{w/ U}). 
As shown in Table~\ref{tab:intv_effect_ratio}, this ratio increases to nearly~1.0 across all Polymath splits for both models, demonstrating that once understanding is resolved, the multilingual reasoning gap collapses regardless of reasoning difficulty.
\sparagraph{Multilingual reasoning strongly correlates with translation ability.}
\begin{figure}[!t]
    \centering
    \includegraphics[width=\columnwidth]{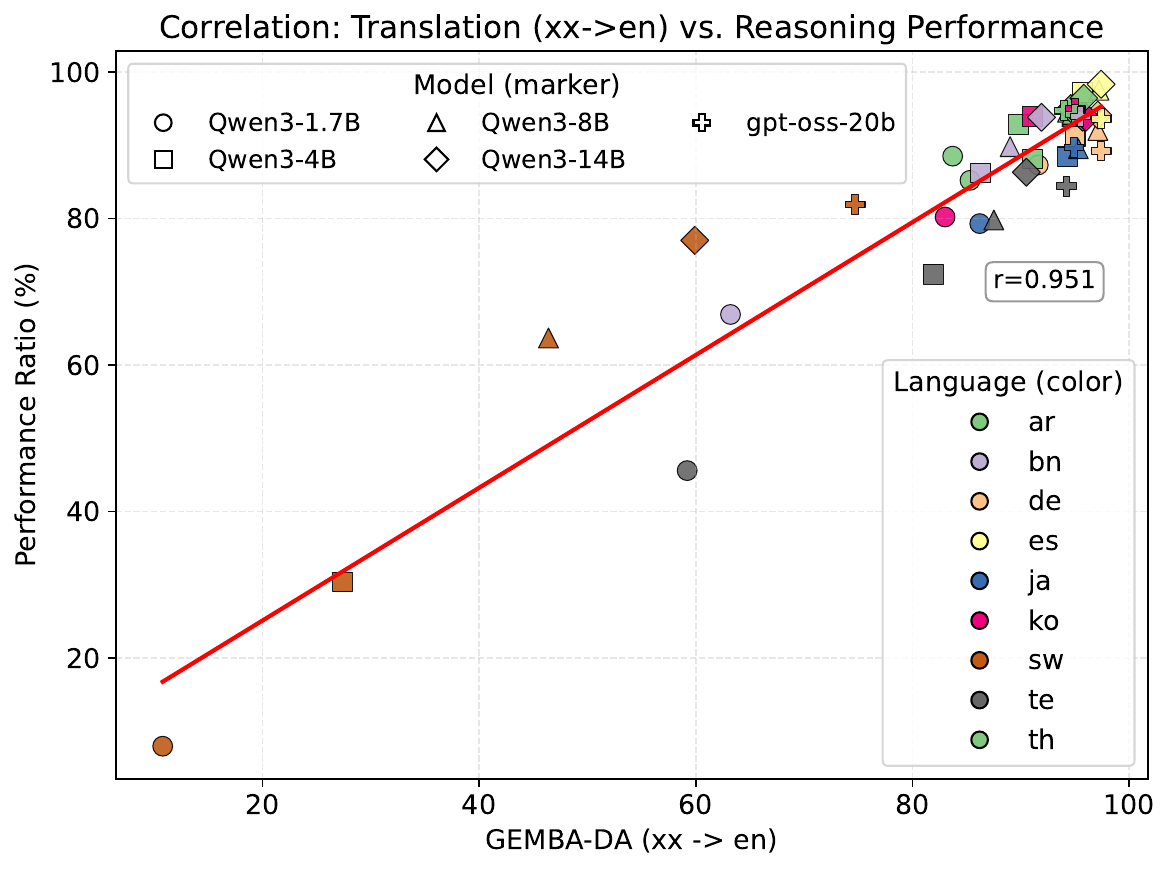}
    \caption{Scatter plot of Reasoning Performance Ratio on Polymath-Low vs. Translation quality on FLORES-200
    (measured by GEMBA-DA in the \texttt{xx} $\rightarrow$ \texttt{en} direction). A global linear trend line shows a strong positive Pearson correlation ($r=0.951$)}
    \label{fig:perf-vs-trans}
\vspace{-8pt}
\end{figure}
Having identified \textit{understanding} failures as the dominant source of multilingual reasoning gaps, we next investigate what affects this process. As defined in Section~\ref{sec:mf_process}, the \textit{understanding} stage translates the input into the reasoning language (typically English). This implies that \textit{understanding}, and consequently multilingual reasoning ability, may depend on how well the model \textit{translates the input \textbf{into} English}. To test this, we measure RLMs’ \texttt{xx}$\rightarrow$\texttt{en} translation quality on FLORES-200~\cite{nllb2022} using GEMBA-DA~\cite{kocmi2023large}, an LLM-based direct assessment metric with gpt-4.1~\cite{openai2025gpt4.1} as the judge. Scores range from 0 (“no meaning preserved”) to 100 (“perfect meaning and grammar”). Figure~\ref{fig:perf-vs-trans} shows reasoning performance ratio on Polymath-Low versus translation quality across ten languages and five models, revealing a strong correlation ($r=0.951$). This result demonstrates that languages more faithfully translated into English also yield stronger reasoning performance, indicating that \textit{understanding} is a key factor in multilingual reasoning.

\section{Detecting Understanding Failures}
From Section~\ref{sec:gap_analysis}, we found that \textit{understanding failure} is the main source of multilingual reasoning gaps.
This raises a key question: \textbf{can we detect when the model fails to understand the input?} To answer this, we define the detection task (Section~\ref{sec:u_task_definition}), present methods and setups (Section~\ref{sec:u_method}, ~\ref{sec:u_setup}), and report experimental results (Section~\ref{sec:u_results_and_analysis}). 
\subsection{Task Definition}\label{sec:u_task_definition}
We formulate understanding failure detection as a binary classification task that operates under the \textbf{Base} setting—i.e., without any interventions. Given the reasoning model’s \textit{input and output signals produced in the Base setting} for a datapoint, we determine whether the model has failed to understand the input (\texttt{label = 1}) or has correctly understood it (\texttt{label = 0}). Ground-truth labels are defined using the model predictions in Section~\ref{sec:gap_analysis}. Specifically, for each language $l$ and dataset $D_l$, we define two sets: $I_{\text{Base}}$ denotes the indices of datapoints correctly answered by the model under the Base setting, and $I_{\text{U}}$ denotes those correctly answered under the Understanding Intervention (w/ U). We restrict our attention to the subset $I_{\text{Base}} \cup I_{\text{U}}$ to exclude datapoints beyond the model’s inherent reasoning capability. Within this subset, we assign labels as follows:
\[
\resizebox{\columnwidth}{!}{$
y_i =
\begin{cases}
1, & \text{if } i \in I_{\text{U}} \setminus I_{\text{Base}} \quad (\text{understanding failure}) \\
0, & \text{if } i \in I_{\text{Base}} \quad (\text{understood})
\end{cases}
$}
\]

This formulation isolates errors attributable to understanding by focusing on cases where the model fails under the Base setting but succeeds once understanding is resolved.

\subsection{Methods}\label{sec:u_method}
As shown in Figure~\ref{fig:intro}, models often leave clear cues in their reasoning traces (e.g., “This is confusing…”) when they fail to understand the input, suggesting that such failures produce detectable signals (see Appendix~\ref{sec:appendix_uf_case_study} for more examples). Motivated by these observations, we adapt detection methods for undesirable behaviors such as hallucination and jailbreak to the task of \textit{understanding failure detection}, as these behaviors also arise from failures to meet the model’s intended objective. We consider three types of approaches.
\sparagraph{LLM–based approaches.}
These methods rely on prompting either external large language models to assess understanding or the reasoning model itself to self-reflect on its own understanding. First, a zero-shot \underline{LLM-based detector}, based on GPT-4.1-mini, is prompted to judge whether the model correctly understood the input given the reasoning trace, following prior work on behavior monitoring~\cite{baker2025monitoring, chan2025can}. Second, a \underline{self-reflection} method~\cite{kadavath2022language, xiong2024can} prompts the reasoning model itself to explicitly reflect on whether it understood the input after producing its reasoning trace.
\sparagraph{Token-probability–based approaches.}
These approaches analyze token-level probabilities over the full reasoning trace to quantify the model’s uncertainty, which we hypothesize may reflect failures in the \textit{understanding} stage. We hypothesize that when the model fails to correctly interpret the input, this may manifest as lower token-level confidence throughout the reasoning process. Following prior work on hallucination detection~\cite{manakul-etal-2023-selfcheckgpt}, we compute two confidence-based signals, the \underline{average} and \underline{minimum per-token confidence}, and classify samples as understanding failures when their values fall below a calibrated threshold. We further use the \underline{Input negative log-likelihood (NLL)} as a proxy for input familiarity and hypothesize that higher NLL may be associated with understanding failures.
\sparagraph{Supervised approaches.}
Supervised detectors are trained to predict understanding failure labels (defined in Section~\ref{sec:u_task_definition}) from textual or hidden-state features. The first is a fine-tuned \underline{mmBERT~\cite{marone2025mmbert} detector}, adapted from prior behavioral monitoring work~\cite{chan2025can}. It takes the query and its reasoning trace as input. The second is a \underline{prober}, a two-layer perceptron that takes as input the final-layer hidden state corresponding to the last token of the reasoning trace, following prior probing methods~\cite{azaria2023internal, zhang2025reasoning}. 

Finally, we define a \underline{random baseline} that predicts “not understood” with probability equal to the proportion of such labels in the calibration data for each language. This serves as a performance floor, quantifying performance achievable using only language-specific label distribution priors.

\begin{table*}[!ht]
\centering
\renewcommand{\arraystretch}{0.85}
\resizebox{\textwidth}{!}{
\begin{tabular}{llcccccc}
\toprule
\multirow{2}{*}{\textbf{Dataset}} & \multirow{2}{*}{\textbf{Method}} 
& \multicolumn{3}{c}{\textbf{Qwen3-4B}} 
& \multicolumn{3}{c}{\textbf{gpt-oss-20b}} \\
\cmidrule(lr){3-5} \cmidrule(lr){6-8}
 &  & \textbf{Balanced acc ↑} & \textbf{F1 ↑} & \textbf{PR-AUC ↑} 
 & \textbf{Balanced acc ↑} & \textbf{F1 ↑} & \textbf{PR-AUC ↑} \\
\midrule

\multirow{8}{*}{Polymath-Low}                         
& Random baseline       & 66.2 ± 2.6  & 41.0 ± 3.6  & -           & 53.4 ± 4.4  & 12.0 ± 7.1  & -           \\
\cmidrule(lr){2-8}

& Avg confidence        & 77.8 ± 1.9  & 50.7 ± 1.1  & 54.3 ± 1.9  & \underline{70.1 ± 1.1}  & 24.7 ± 1.6  & 18.6 ± 3.5  \\
& Min confidence        & 71.4 ± 3.9  & 42.4 ± 3.0  & 33.6 ± 1.9  & 65.3 ± 5.0  & 24.4 ± 3.5  & 17.1 ± 4.7  \\
& Input NLL            & 64.9 ± 0.1  & 39.8 ± 0.7  & 32.8 ± 1.8  & 57.3 ± 2.9  & 16.3 ± 2.2  & 12.1 ± 1.3  \\
\cmidrule(lr){2-8}
& Self-reflection       & 61.6 ± 2.0  & 36.7 ± 4.9  & -           & 50.7 ± 0.7  & 2.6 ± 2.7   & -           \\
& LLM-based detector & 71.7 ± 1.0  & 55.7 ± 1.4  & -           & 51.9 ± 0.5  & 7.8 ± 1.6   & -           \\
\cmidrule(lr){2-8}
& mmBERT detector       & \underline{85.2 ± 1.2}  & \textbf{65.9 ± 6.2}  & \underline{72.6 ± 1.2}  & 66.2 ± 3.6  & \textbf{34.8 ± 1.5}  & \textbf{31.7 ± 3.0}  \\
& Prober               & \textbf{85.5 ± 1.3}  & \underline{63.7 ± 3.2}  & \textbf{75.7 ± 1.5}  & \textbf{75.4 ± 2.4}  & \underline{34.5 ± 7.7}  & \underline{30.3 ± 1.7}  \\
\midrule[0.4pt]
\vspace{-0.5em} \\[-0.8em]   
\midrule[0.4pt]
\multirow{8}{*}{MMLU-ProX-Lite}            
& Random baseline       & 55.4 ± 3.5  & 20.6 ± 4.5  & -           & 50.5 ± 0.1  & 10.3 ± 0.9  & -           \\
\cmidrule(lr){2-8}

& Avg confidence        & 64.4 ± 5.3  & \underline{34.6 ± 6.5}  & 31.0 ± 2.6  & 52.2 ± 1.5  & 15.3 ± 1.5  & 8.8 ± 1.5   \\
& Min confidence        & \underline{67.6 ± 1.0}  & 33.0 ± 1.8  & 24.5 ± 2.2  & 63.8 ± 1.5  & 21.4 ± 1.5  & 16.3 ± 3.1  \\
& Input NLL            & 55.0 ± 2.5  & 17.8 ± 8.1  & 22.2 ± 1.9  & 53.9 ± 0.5  & 15.3 ± 0.7  & 9.6 ± 1.2   \\
\cmidrule(lr){2-8}

& Self-reflection       & 59.3 ± 1.1  & 29.4 ± 2.6  & -           & 53.9 ± 0.4  & 14.4 ± 1.3  & -           \\
& LLM-based detector & 55.6 ± 0.4  & 20.1 ± 1.0  & -           & 51.1 ± 0.3  & 4.8 ± 1.2   & -           \\
\cmidrule(lr){2-8}
& mmBERT detector       & 59.7 ± 0.3  & 30.1 ± 1.3  & \underline{35.8 ± 1.7}  & \textbf{61.9 ± 6.0}  & \underline{25.4 ± 6.0}  & \textbf{24.5 ± 5.2}  \\
& Prober              & \textbf{77.3 ± 1.0}  & \textbf{44.5 ± 3.0}  & \textbf{42.6 ± 1.8}  & \underline{60.1 ± 1.1}  & \textbf{26.0 ± 3.4}  & \underline{20.6 ± 3.4}  \\
\bottomrule
\end{tabular}
}
\caption{Performance of understanding failure detection methods reported as mean ± stdev. Best performance across methods is highlighted in bold and the second best is underlined. Supervised approaches achieve the best performance overall. Per-language results are provided in Appendix~\ref{sec:appendix_language_specific_detection}.}
\label{tab:understanding_detection}
\end{table*}

\subsection{Experimental Settings}\label{sec:u_setup}
We evaluate understanding detection approaches on Polymath-Low~\cite{wang2025polymath} and MMLU-ProX-Lite~\cite{xuanmmluprox}, using the same set of languages as in Section~\ref{sec:gap_setup}.
For each dataset, we provide the detection methods with detection signals produced by Qwen3-4B and gpt-oss-20b in the \textbf{Base} setting.

\sparagraph{Evaluation metrics.}
Since these models already perform reasonably well in the evaluated languages, over 86\% of the samples are labeled as \textit{understood} (label = 0).
To properly assess detection methods under class imbalance, we treat \textit{not understood} (label = 1) as the positive class and report the following metrics: \textbf{Balanced accuracy} averages the true-positive and true-negative rates, which is informative in imbalanced settings. \textbf{F1} is the harmonic mean of precision and recall. \textbf{PR-AUC} measures the area under the precision–recall curve, providing a threshold-independent evaluation. Metrics are computed by aggregating samples across languages, and we report the average over three  runs.

\sparagraph{Calibration data.}
For Polymath-Low, we use MGSM~\cite{shilanguage} as calibration data, excluding samples already included in Polymath-Low. For MMLU-ProX-Lite, we use the validation split of MMLU-ProX-Lite. This data is used for threshold calibration in token-probability–based approaches and as training data for supervised approaches. Additional implementation details on methods and data statistics are in Appendix~\ref{appendix:uf_details}.

\subsection{Results and Analysis}\label{sec:u_results_and_analysis}
\sparagraph{Supervised approaches achieve the best detection performance.} 
Table~\ref{tab:understanding_detection} reports the performance of understanding failure detection methods. Supervised approaches achieve the highest detection performance and substantially outperform the random baseline, indicating that they capture meaningful signals rather than merely exploiting language-specific label distributions. Some Token-probability–based approaches (Avg, Min confidence) perform moderately but generally fall behind, while Input NLL and self-reflection signals are less effective. The LLM-based detector also shows limited performance, indicating that understanding failure detection remains challenging even for recent LLMs without task-specific supervision. 

\begin{table*}[!t]
\centering
\resizebox{\textwidth}{!}{
\begin{tabular}{l|l|cccccc|cc|cc|c}
\toprule
& \textbf{Method} & en & de & es & ar & ja & ko & th & bn & sw & te & \textbf{Avg. (Translator usage)} \\
\midrule
\multirow{3}{*}{\textbf{Polymath-Low}} 
& Base                  & 96.5 & 88.0 & 93.9 & 89.6 & 85.3 & 90.7 & 85.1 & 83.2 & 29.3 & 69.9 & \textbf{81.1 (0.0\%)} \\
& Selective translation & 
\makecell{96.3 \\ (1.6\%)} &
\makecell{88.3 \\ (3.7\%)} &
\makecell{94.4 \\ (3.7\%)} &
\makecell{90.4 \\ (4.8\%)} &
\makecell{86.1 \\ (13.1\%)} &
\makecell{91.5 \\ (5.9\%)} &
\makecell{88.3 \\ (9.6\%)} &
\makecell{86.7 \\ (26.1\%)} &
\makecell{81.3 \\ (86.4\%)} &
\makecell{77.1 \\ (37.9\%)} &
\textbf{88.0 (19.3\%)}  \\
\cmidrule(lr){2-13}
& Full translation      & 96.0 & 88.3 & 93.3 & 90.9 & 87.5 & 92.5 & 89.6 & 90.4 & 85.3 & 80.5 & \textbf{89.4 (100.0\%)} \\
\midrule
\multirow{3}{*}{\textbf{MMLU-ProX-Lite}} 
& Base                  & 77.0 & 77.8 & 76.5 & 73.3 & 75.0 & 74.6 & 73.9 & 74.6 & 53.6 & 71.1 & \textbf{72.7 (0.0\%)} \\
& Selective translation &
\makecell{77.3 \\ (5.2\%)} &
\makecell{76.9 \\ (10.0\%)} &
\makecell{77.2 \\ (8.0\%)} &
\makecell{74.1 \\ (17.9\%)} &
\makecell{75.9 \\ (19.2\%)} &
\makecell{75.9 \\ (14.4\%)} &
\makecell{73.9 \\ (18.4\%)} &
\makecell{74.1 \\ (27.0\%)} &
\makecell{65.1 \\ (55.8\%)} &
\makecell{72.8 \\ (31.6\%)} &
\textbf{74.3 (20.8\%)} \\
\cmidrule(lr){2-13}
& Full translation      & 79.0 & 77.8 & 78.3 & 76.0 & 77.8 & 77.4 & 76.1 & 76.4 & 71.6 & 74.8 & \textbf{76.5 (100.0\%)} \\
\bottomrule
\end{tabular}
}
\caption{Performance of translation strategies with Qwen3-4B. 
\textbf{Avg. (Translator usage)} reports average accuracy across languages, with the overall translator usage shown in parentheses. 
For \textbf{Selective translation}, per-language translator usage (\%) is shown below each accuracy score. Appendix~\ref{sec:appendix_st_oss_20b} provides results on gpt-oss-20b.}
\label{tab:selective_translation_results}
\end{table*}

\sparagraph{Generalization to Unseen Languages.}
\begin{figure}[!t]
    \centering
    \includegraphics[width=\columnwidth]{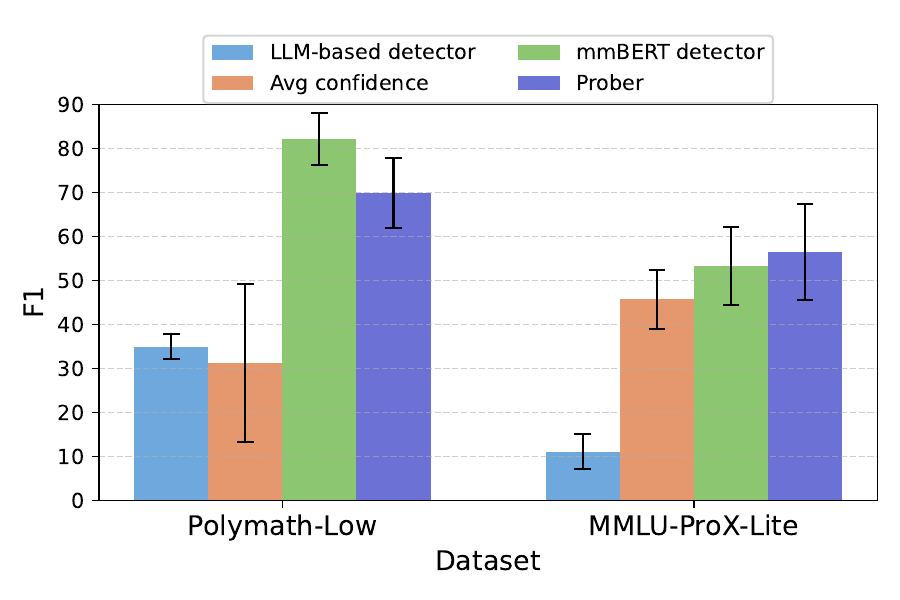}
    \caption{F1 scores for understanding failure detection on \textit{unseen} languages (fr, mr, wo) across two benchmarks on Qwen3-4B.}
    \label{fig:unseen_language}
\end{figure}
We next examine whether supervised approaches can generalize beyond the languages they were trained on. To this end, we use French (fr), Marathi (mr), and Wolof (wo), spanning high-resource (fr) to low-resource (mr, wo) languages. Figure~\ref{fig:unseen_language} shows results. We observe that both the mmBERT detector and the prober consistently outperform the average-confidence baseline and the LLM-based detector across unseen languages. This shows that the superiority of supervised classifiers generalizes beyond training languages, highlighting robustness in multilingual settings.

\section{Selective Translation}\label{sec:u_selective_translation}
Having established that understanding failures can be reliably detected, we investigate how such detection can be used to address the multilingual reasoning gap. To this end, we propose \textit{Selective Translation}, a simple yet effective strategy that incorporates an English translation of the input into the initial reasoning trace only when an understanding failure is detected.\footnote{We use GPT-4.1 for translation; see Appendix~\ref{sec:appendix_selective_translation} for implementation details.} Table~\ref{tab:selective_translation_results} presents the results of this approach using the \textit{prober} as a detector. Selective Translation is particularly effective in low-resource languages, where understanding failures are frequent. By intervening on cases that would otherwise fail, it substantially reduces the multilingual reasoning gap—improving average accuracy from 81.1 to 88.0 on Polymath-Low and from 72.7 to 74.3 on MMLU-ProX-Lite, closely approaching full translation (89.4 and 76.5)—while requiring translation for only about 20\% of inputs on average. These results show that the detector accurately identifies when translation is needed and enables efficient bridging of the multilingual reasoning gap.\footnote{Appendix~\ref{sec:appendix_efficiency} analyzes latency and detector overhead, showing that the detector itself introduces negligible latency, while ST incurs modest latency overall.}
\begin{figure}[t]
    \centering
    \includegraphics[width=\linewidth]{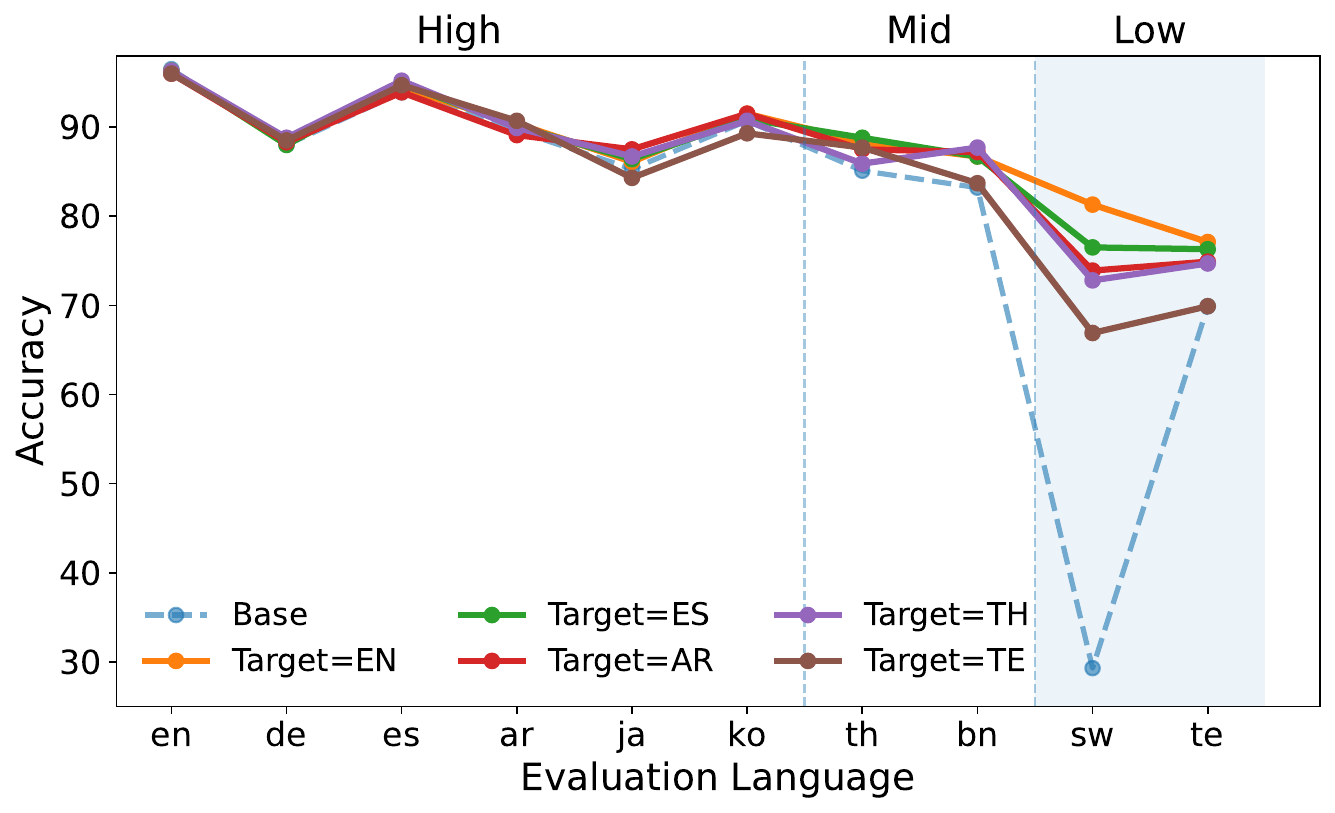}
    \caption{Accuracy of Selective Translation with different translation target languages across evaluation languages on Polymath-Low with Qwen3-4B.}
    \label{fig:st_target_language}
\end{figure}
\sparagraph{Selective Translation with Non-English Translation Targets.}
Prior work has shown that multilingual reasoning can benefit from using non-English target languages \cite{gao2025thinkingmultilinguallyempowerllm}. In our setting, we use English as the default translation target, as reasoning traces in RLMs are strongly English-centric. To validate this choice, we evaluate selective translation (ST) with non-English targets (es, ar, th, te). Figure~\ref{fig:st_target_language} shows that English performs best overall, while performance degrades as the target shifts toward lower-resource languages (e.g., te). This degradation is more pronounced on low-resource evaluation languages (e.g., sw, te), where ST is frequently applied. These results suggest that using lower-resource targets introduces additional understanding failures, supporting our design choice of using English as the default translation target.
\sparagraph{Early Detection of Understanding Failures.}
\begin{figure}[!t]
    \centering
    \includegraphics[width=\columnwidth]{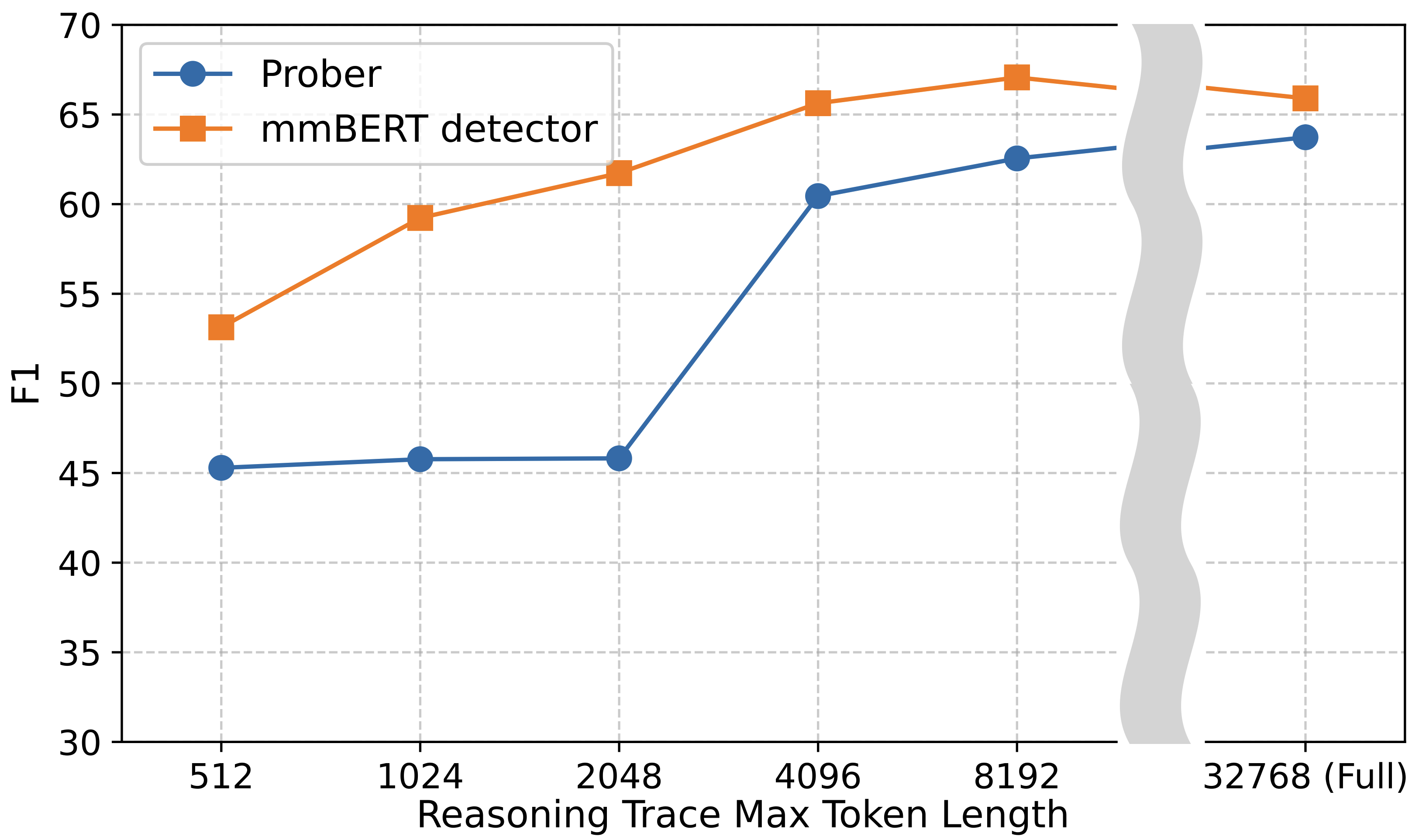}
    \caption{F1 score of understanding failure detection on Polymath-Low with Qwen3-4B, measured with varying the maximum reasoning trace token length.}
    \label{fig:early_detection}
\end{figure}
Finally, we investigate whether understanding failures can be detected \textbf{early}, before the model completes its full reasoning trace. Each detector is trained on reasoning traces truncated to different lengths, and Figure~\ref{fig:early_detection} shows how detection performance improves as more tokens become available. Notably, both the mmBERT detector and the prober achieve performance comparable to the full-trace setting with only 4,096 tokens, indicating that reliable detection does not require observing the entire reasoning process. Appendix~\ref{sec:appendix_st_early_detection} further shows that Selective Translation remains effective under this early-detection setting. Together, these results highlight the promise of early detection for improving the efficiency of Selective Translation.
\section{Conclusion}
We presented the first systematic analysis of the multilingual reasoning gap in RLMs and showed that it primarily stems from \textit{understanding} failures. We further demonstrated that such failures can be detected from behavioral signals and effectively mitigated through \textit{Selective Translation}, which achieves near–full-translation performance while using translation selectively. Future work could explore integrating detection and mitigation directly into model training.
\section*{Limitations}
Our experiments focus on mathematical and STEM reasoning tasks. While we consider problems with varying difficulty levels and task types within these domains, verifying whether our findings generalize to other domains such as commonsense reasoning would help assess the broader applicability of our findings.

From the language perspective, our study includes ten typologically and resource-wise diverse languages. While this set does not cover all language families, we believe our conclusions capture general multilingual trends. Nevertheless, verifying them on additional and lower-resource languages could further strengthen our claims.

Finally, our analysis centers on scenarios where the model predominantly reasons in English—the dominant internal language observed in most reasoning language models, especially when processing inputs from low-resource languages. This setting is particularly suitable for decomposing multilingual reasoning into distinct stages, as the understanding stage naturally emerges when the model internally translates the input into English before reasoning. In contrast, models that natively reason in other languages (e.g., Russian) may not exhibit a clear understanding stage; however, such cases mostly involve high-resource languages that contribute less to the overall multilingual reasoning gap. Therefore, we focus on the English reasoning setting. Nonetheless, investigating how multilingual reasoning operates in such non-English reasoning models would be a valuable direction for future work.

\section*{Ethical Considerations}
In our research, we use datasets such as Polymath~\cite{wang2025polymath} and MMLU-ProX-Lite~\cite{xuanmmluprox}, which are licensed under Apache 2.0 and MIT, respectively. The models used in our research—Qwen3~\cite{yang2025qwen3} (1.7B/4B/8B/14B) and gpt-oss-20b~\cite{agarwal2025gpt}—are also licensed under Apache 2.0. All datasets and models were used strictly for research purposes, and no artifacts were utilized beyond the scope of the study. We use ChatGPT and GitHub Copilot for writing and coding assistance.
\section*{Acknowledgments}
This research was supported by Culture, Sports and Tourism R\&D Program through the Korea Creative Content Agency grant funded by the Ministry of Culture, Sports and Tourism in 2025 (Project Name: Development of an AI-Based Korean Diagnostic System for Efficient Korean Speaking Learning by Foreigners, Project Number: RS-2025-02413038, Contribution Rate: 45\%); by the Smart HealthCare for Police Officers Program(www.kipot.or.kr) through the Korea Institutes of Police Technology(KIPoT) funded by the Korean National Police Agency(KNPA, Korea)(No. RS-2022-PT000186, Contribution Rate: 45\%); and by the Institute of Information \& communications Technology Planning \& Evaluation (IITP) grant funded by the Korea government(MSIT) (No.RS-2019-II191906, Artificial Intelligence Graduate School Program(POSTECH), Contribution Rate: 10\%).

\bibliography{custom}

\appendix
\clearpage
\section{Language Distributions}\label{appendix:language_distribution}
\subsection{Computing Language Distributions}
We estimate the language composition of reasoning traces and final responses by aggregating sentence-level language identification results from the fastText~\cite{joulin2016bag, joulin2016fasttext} language identification model.
Given a text sample (either a reasoning trace or a final response), we first use regular expressions to remove LaTeX expressions, code blocks, and formula-like spans (e.g., \texttt{\$...\$}, \texttt{\textbackslash begin\{equation\}}...\texttt{\textbackslash end\{equation\}}, or backtick code), ensuring that only natural-language segments remain for reliable detection. The cleaned text is then segmented into sentences based on punctuation,
and those shorter than 10 characters are discarded to maintain reliable predictions from the fastText model.
Each valid sentence is subsequently classified by the fastText language identification model (\texttt{lid.176.ftz}).
Finally, we compute the overall language distribution of a text $t$ for all detected languages $l$ as:
\begin{align*}
P(l) &= \frac{\text{\# of sentences predicted as } l}
{\text{total number of valid sentences in } t},
\end{align*}
resulting in a normalized distribution that captures the relative prevalence of each language.

\subsection{Results}
\begin{figure*}[!ht]
    \centering
    \begin{subfigure}{\textwidth}
        \centering
        \includegraphics[width=1.0\textwidth]{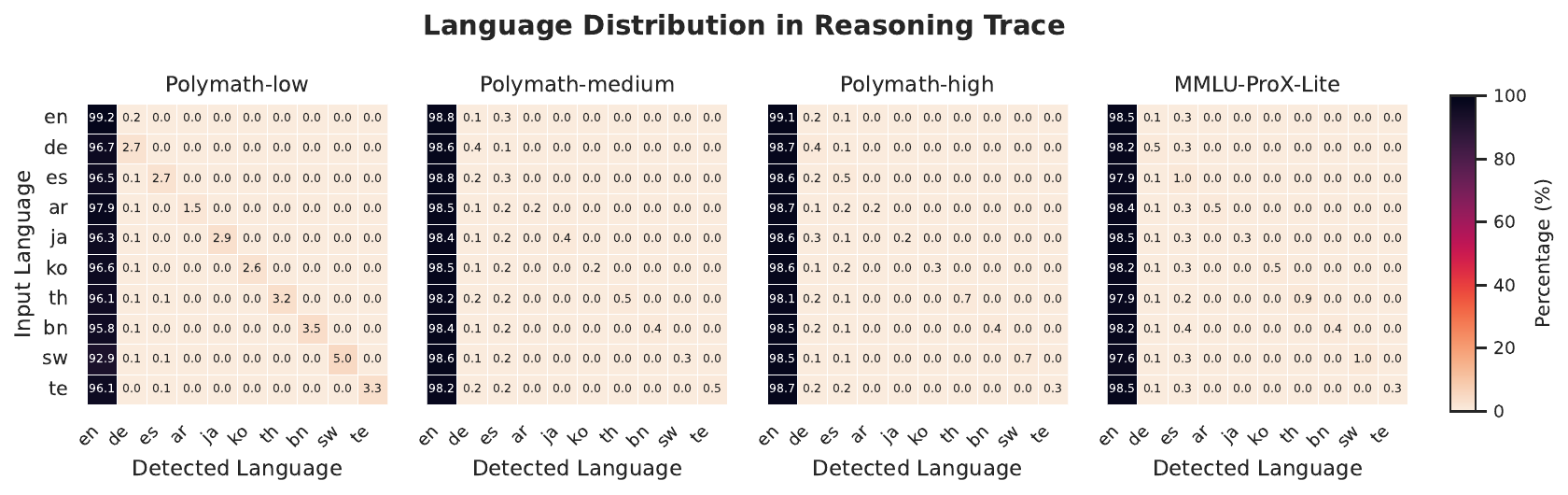}
        \label{fig:ld_4b_tr}
    \end{subfigure}
    \begin{subfigure}{\textwidth}
        \centering
        \includegraphics[width=1.0\textwidth]{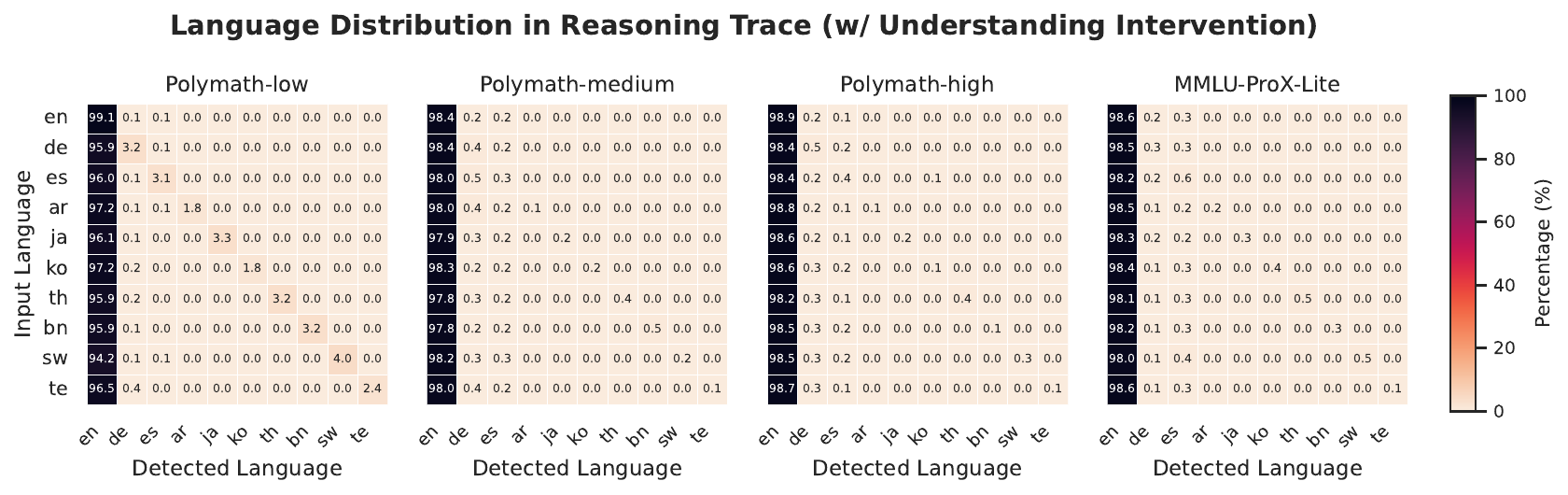}
        \label{fig:ld_4b_tr_intv}
    \end{subfigure}
    \begin{subfigure}{\textwidth}
        \centering
        \includegraphics[width=1.0\textwidth]{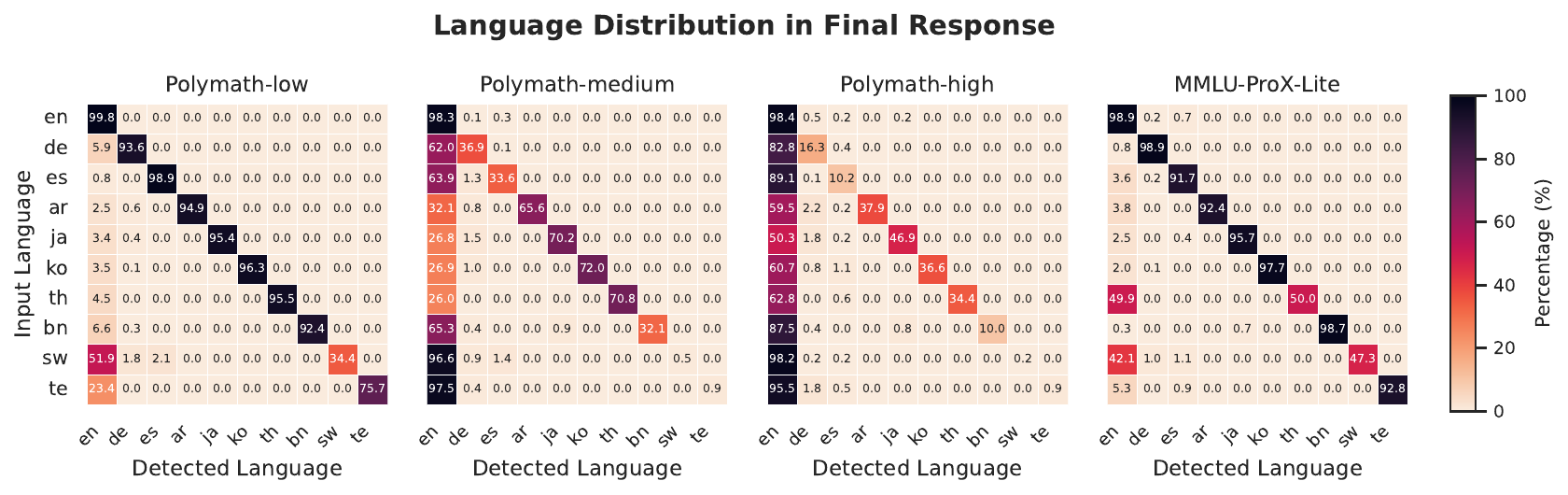}
        \label{fig:ld_4b_final}
    \end{subfigure}
    \begin{subfigure}{\textwidth}
        \centering
        \includegraphics[width=1.0\textwidth]{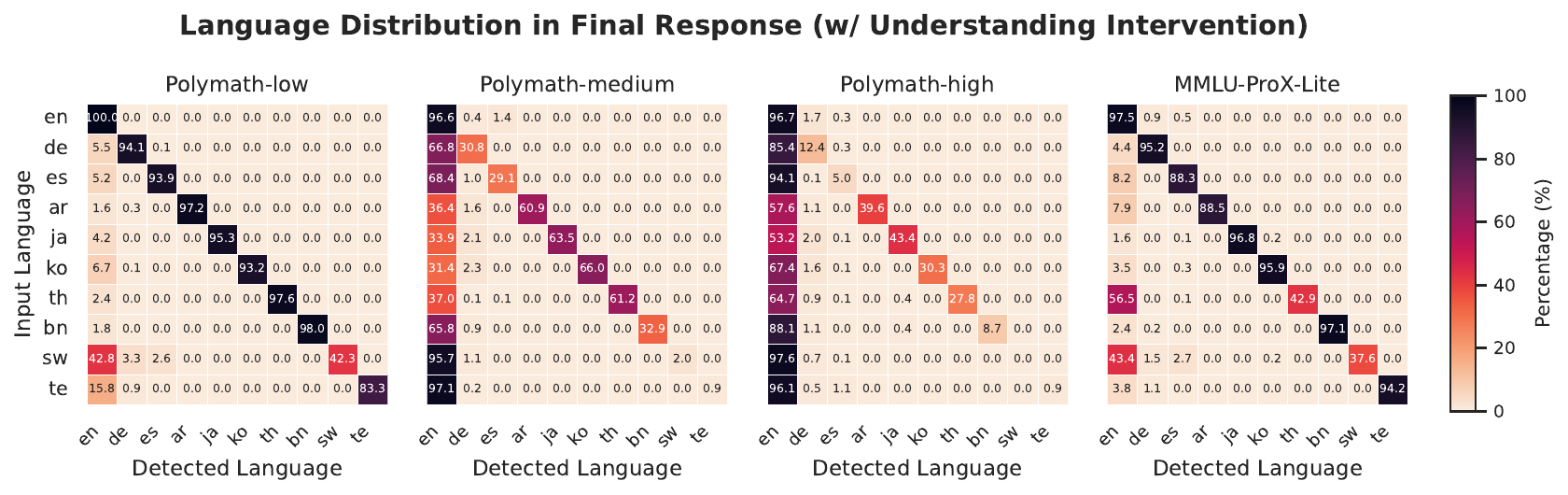}
        \label{fig:ld_4b_final_intv}
    \end{subfigure}
    \caption{Language distributions of reasoning traces and final responses for Qwen3-4B across Polymath (low/medium/high) and MMLU-ProX-Lite datasets, with and without the understanding intervention. }
    \label{fig:ld_4b}
\end{figure*}

\begin{figure*}[!ht]
    \centering
    \begin{subfigure}{\textwidth}
        \centering
        \includegraphics[width=1.0\textwidth]{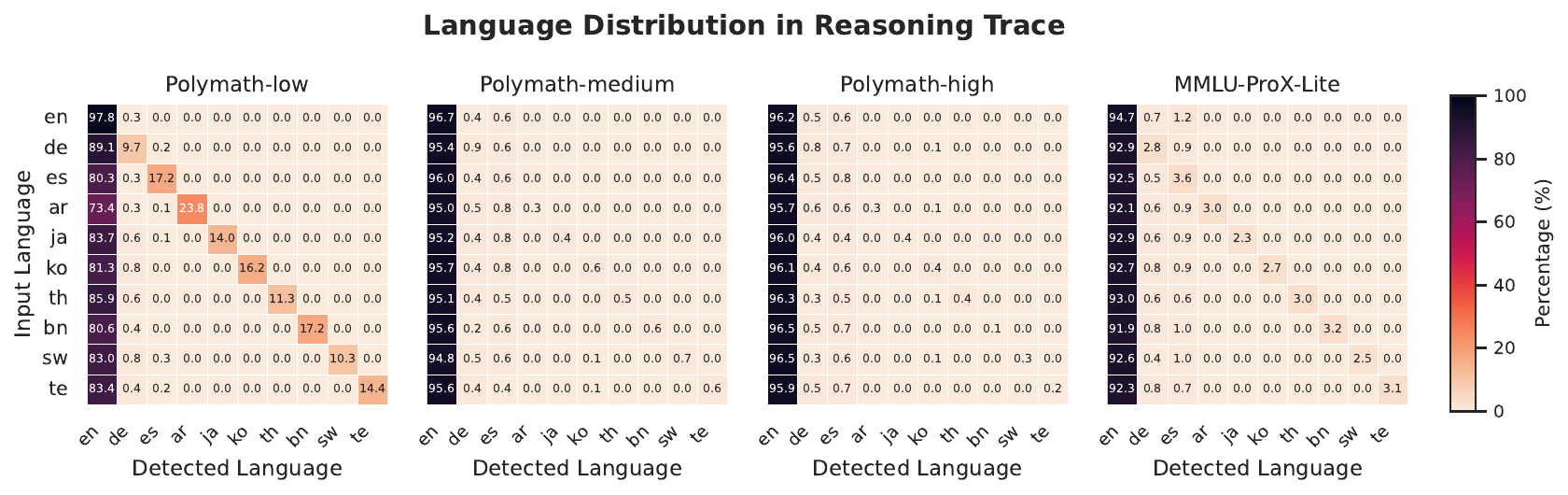}
    \end{subfigure}
    \begin{subfigure}{\textwidth}
        \centering
        \includegraphics[width=1.0\textwidth]{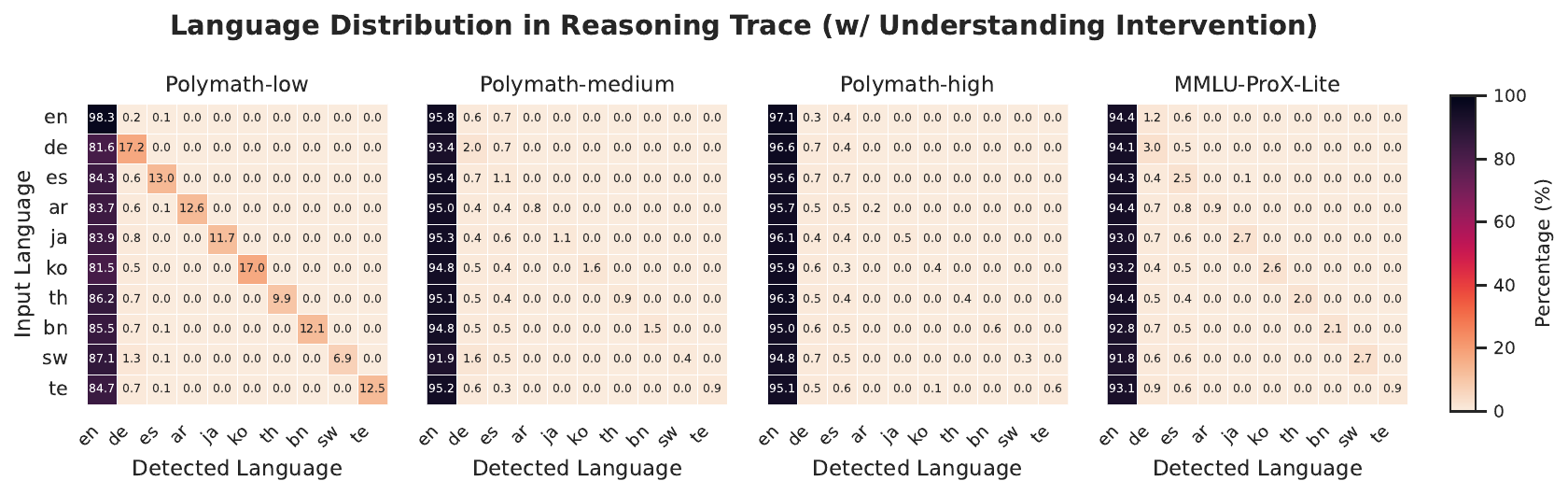}
    \end{subfigure}
    \begin{subfigure}{\textwidth}
        \centering
        \includegraphics[width=1.0\textwidth]{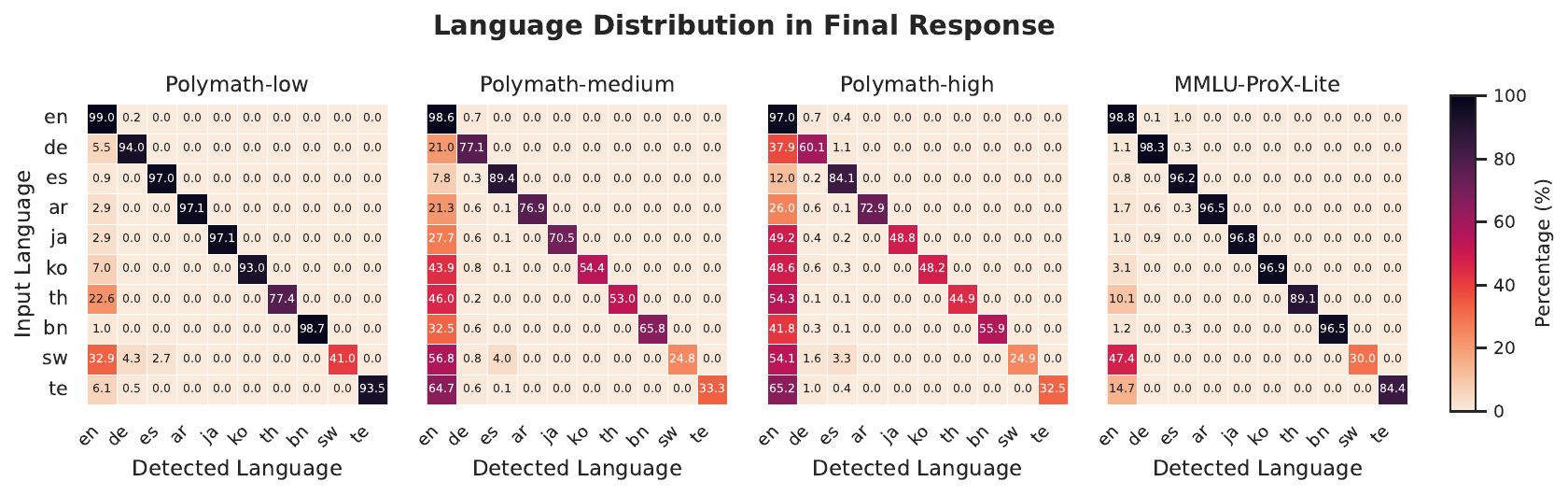}
    \end{subfigure}
    \begin{subfigure}{\textwidth}
        \centering
        \includegraphics[width=1.0\textwidth]{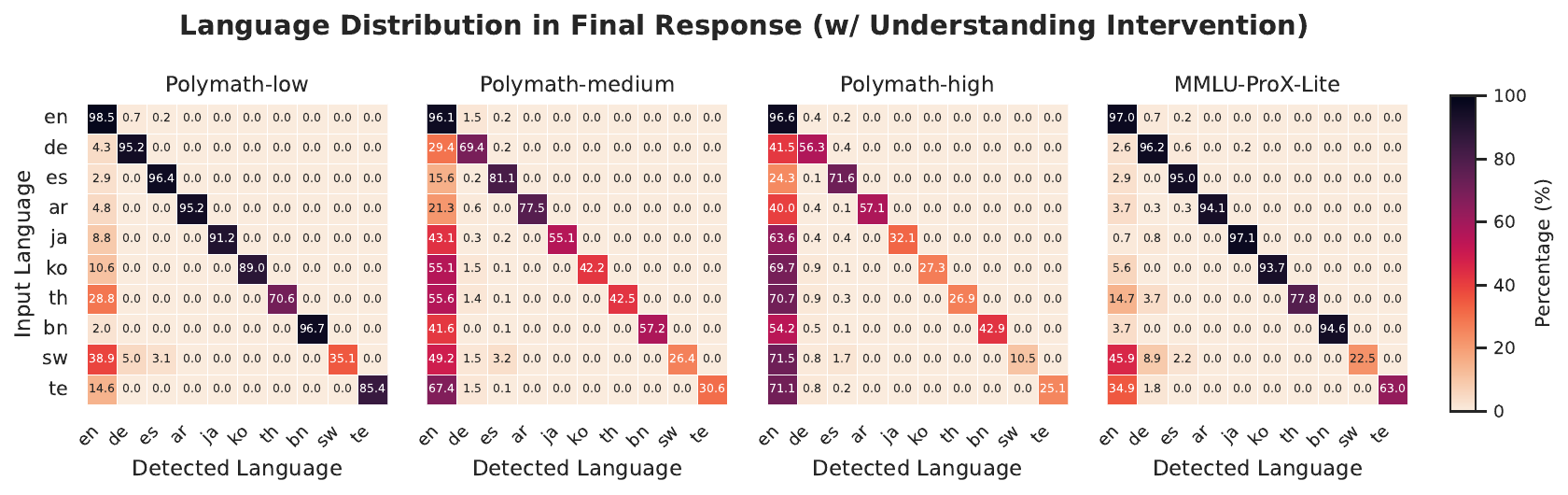}
    \end{subfigure}
    \caption{Language distributions of reasoning traces and final responses for gpt-oss-20b across Polymath (low/medium/high) and MMLU-ProX-Lite datasets, with and without the understanding intervention. }
    \label{fig:ld_oss}
\end{figure*}

\begin{figure*}[!ht]
    \centering
    \begin{subfigure}{\textwidth}
        \centering
        \includegraphics[width=1.0\textwidth]{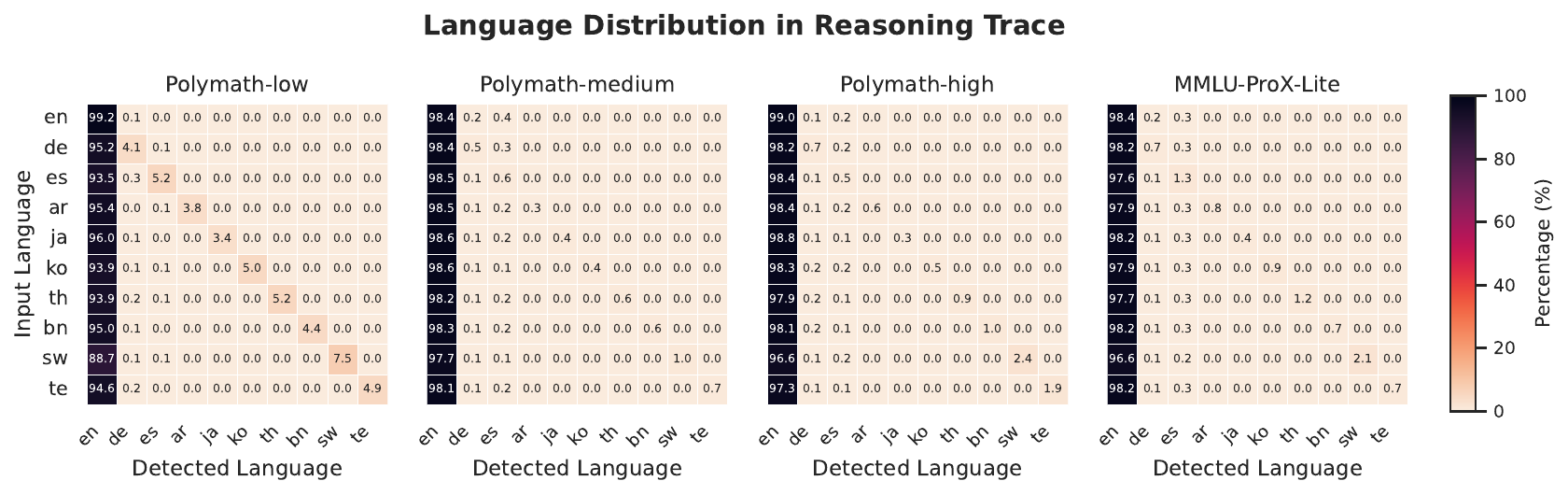}
    \end{subfigure}
    \begin{subfigure}{\textwidth}
        \centering
        \includegraphics[width=1.0\textwidth]{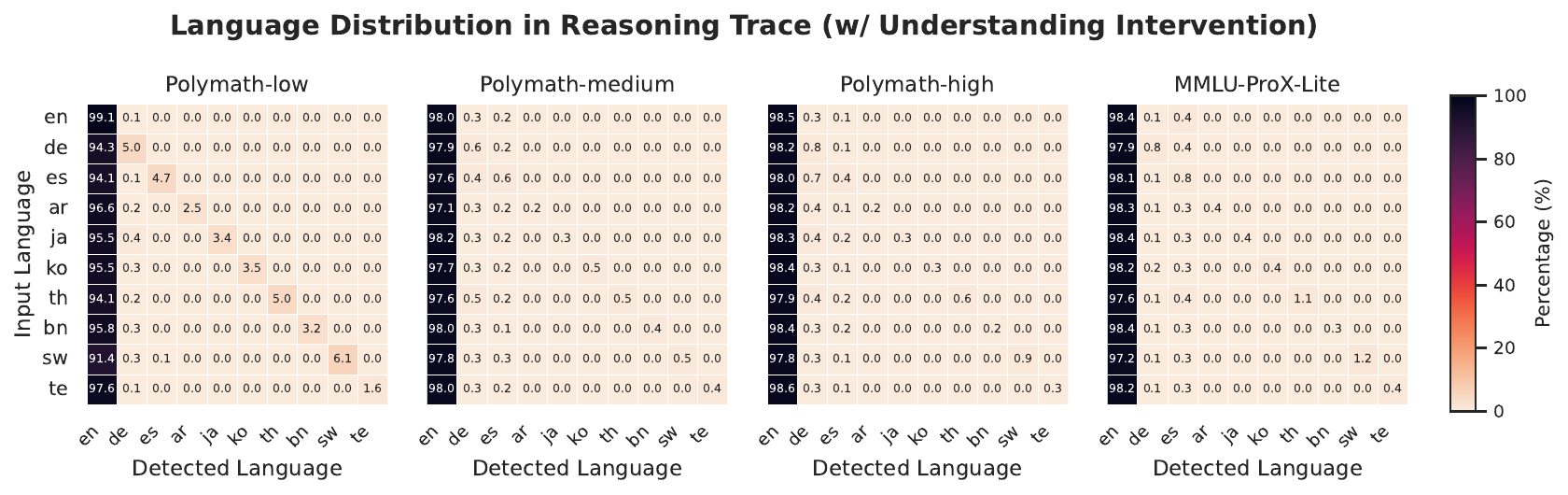}
    \end{subfigure}
    \begin{subfigure}{\textwidth}
        \centering
        \includegraphics[width=1.0\textwidth]{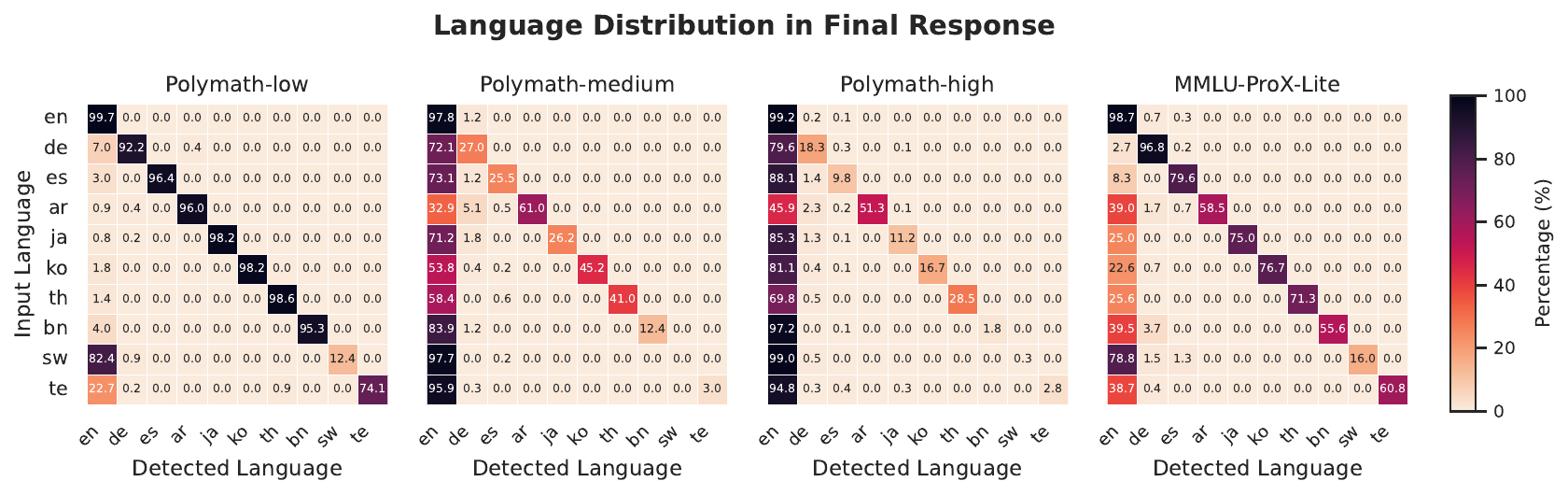}
    \end{subfigure}
    \begin{subfigure}{\textwidth}
        \centering
        \includegraphics[width=1.0\textwidth]{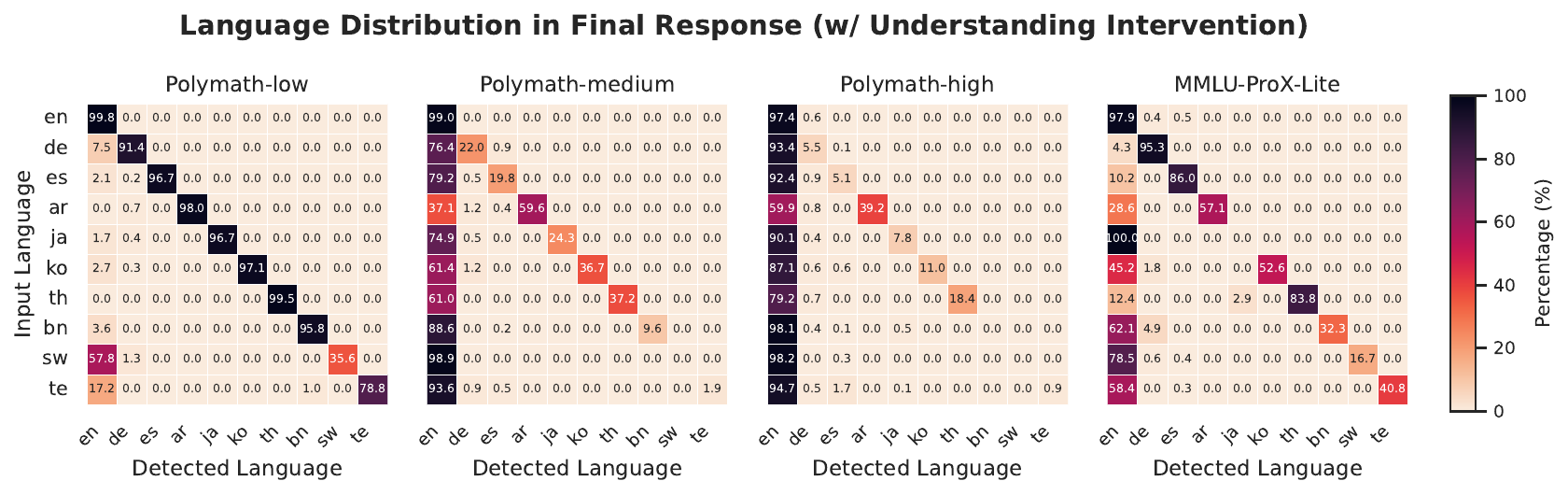}
    \end{subfigure}
    \caption{Language distributions of reasoning traces and final responses for Qwen3-1.7B across Polymath (low/medium/high) and MMLU-ProX-Lite datasets, with and without the understanding intervention. }
    \label{fig:ld_1.7b}
\end{figure*}

\begin{figure*}[!ht]
    \centering
    \begin{subfigure}{\textwidth}
        \centering
        \includegraphics[width=1.0\textwidth]{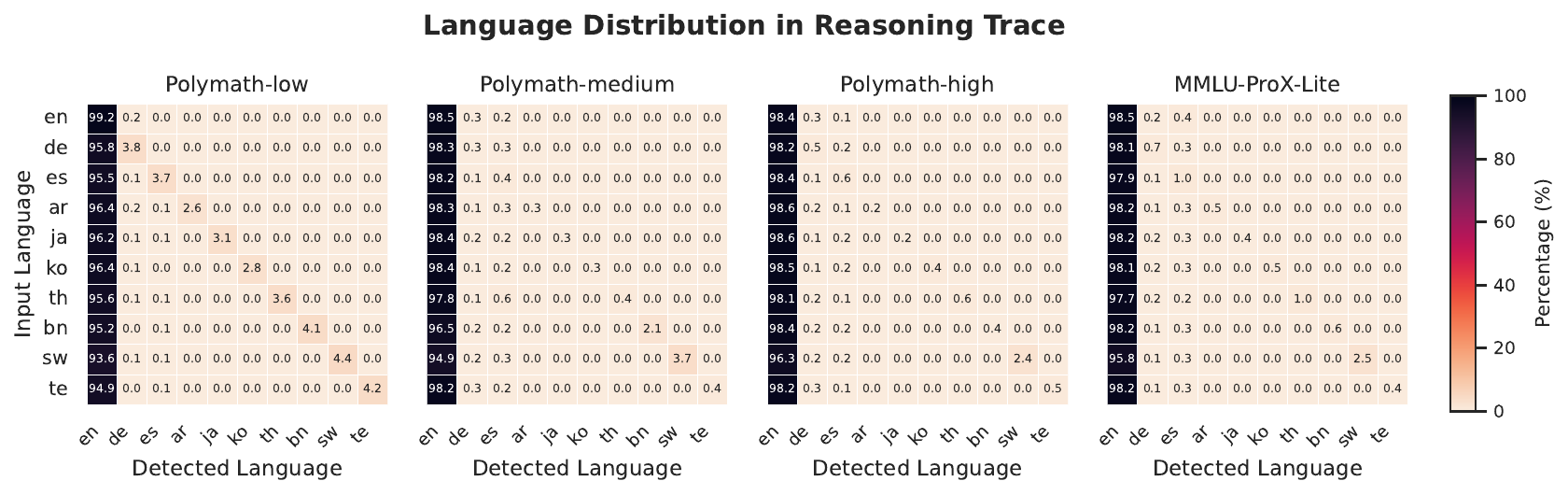}
    \end{subfigure}
    \begin{subfigure}{\textwidth}
        \centering
        \includegraphics[width=1.0\textwidth]{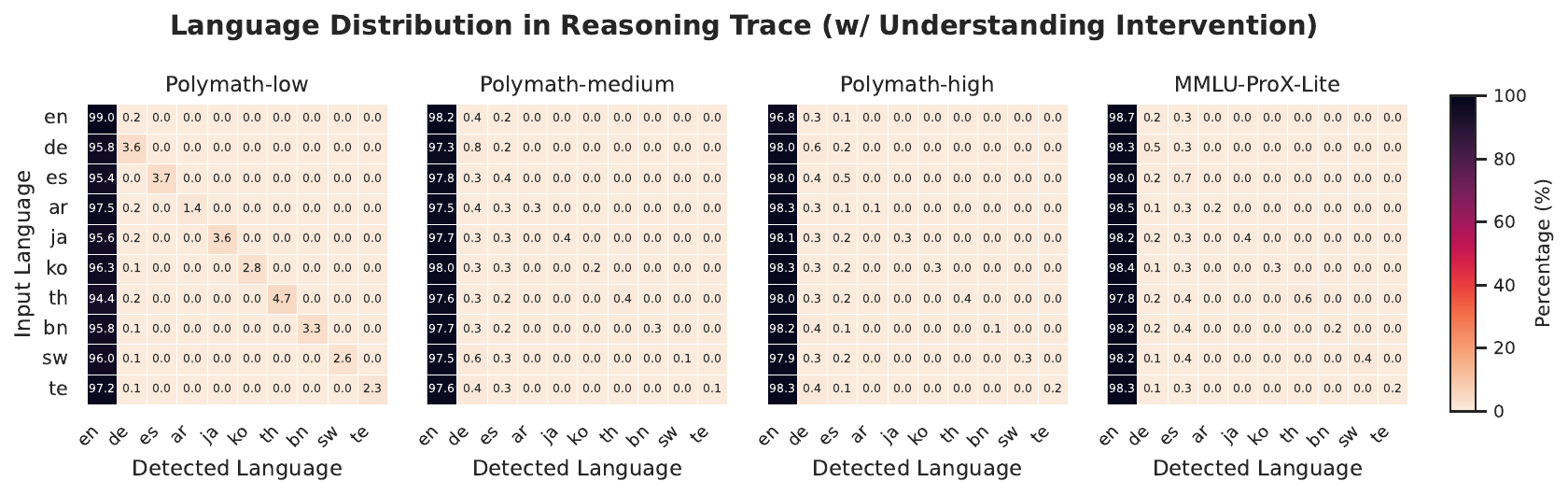}
    \end{subfigure}
    \begin{subfigure}{\textwidth}
        \centering
        \includegraphics[width=1.0\textwidth]{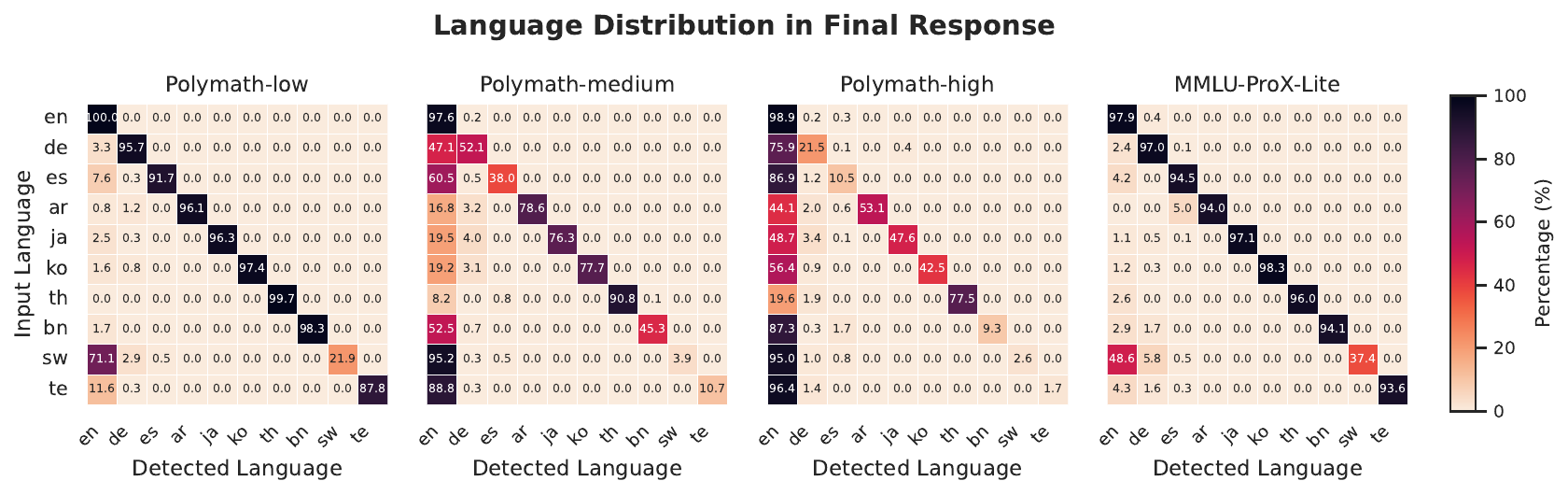}
    \end{subfigure}
    \begin{subfigure}{\textwidth}
        \centering
        \includegraphics[width=1.0\textwidth]{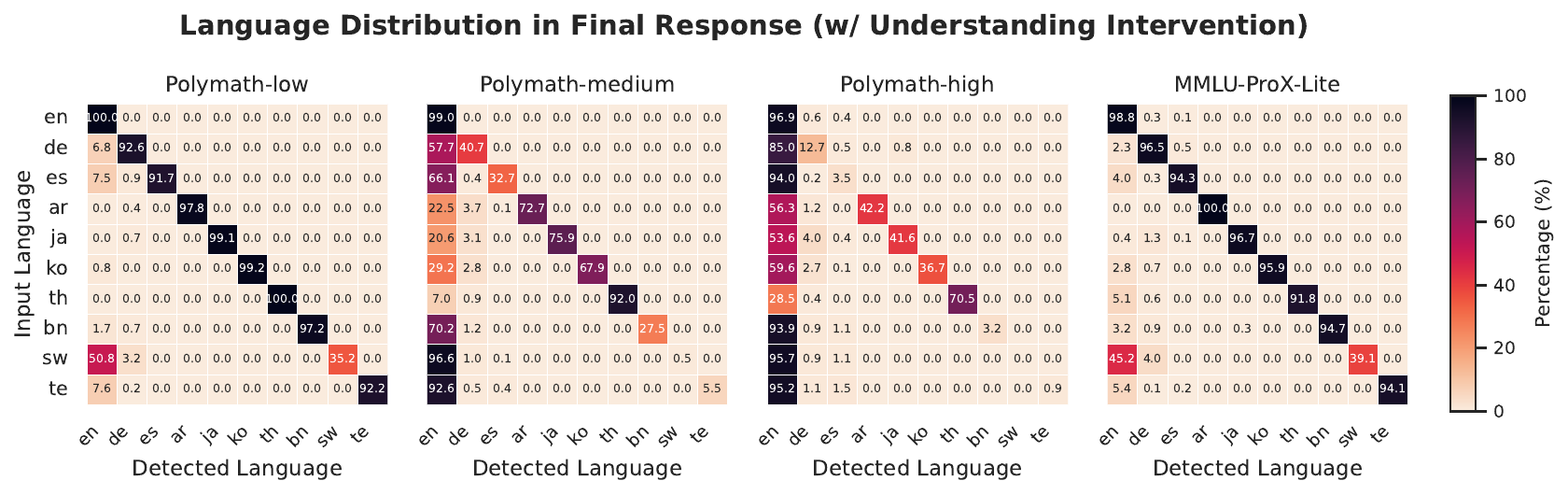}
    \end{subfigure}
    \caption{Language distributions of reasoning traces and final responses for Qwen3-8B across Polymath (low/medium/high) and MMLU-ProX-Lite datasets, with and without the understanding intervention. }
    \label{fig:ld_8b}
\end{figure*}

\begin{figure*}[!ht]
    \centering
    \begin{subfigure}{\textwidth}
        \centering
        \includegraphics[width=1.0\textwidth]{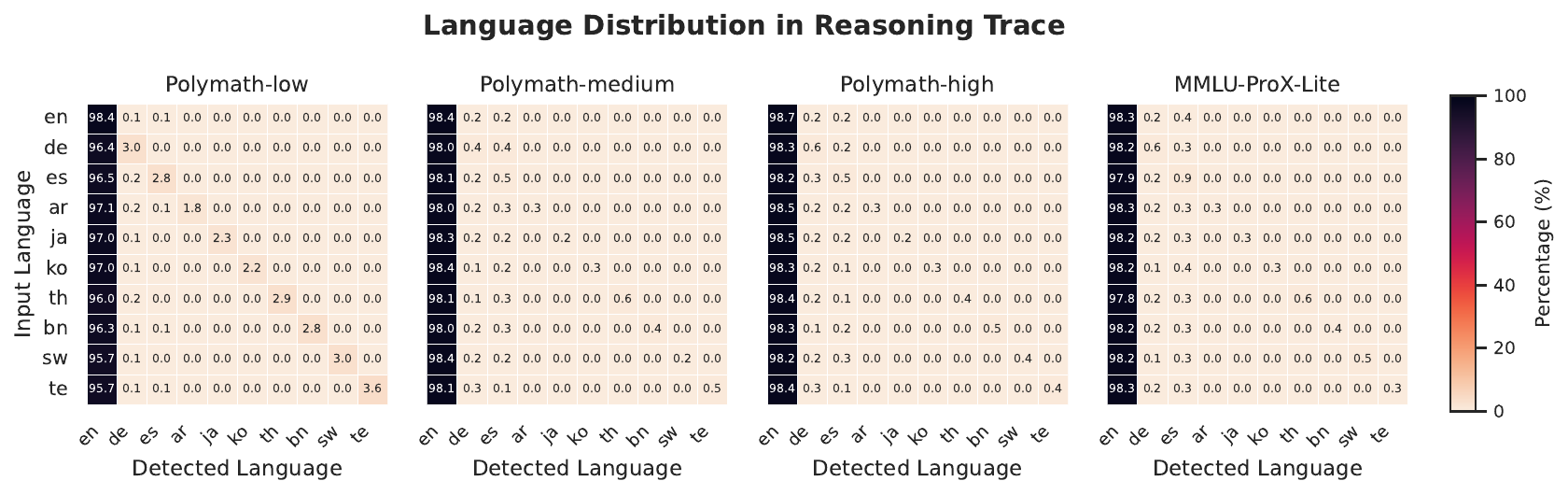}
    \end{subfigure}
    \begin{subfigure}{\textwidth}
        \centering
        \includegraphics[width=1.0\textwidth]{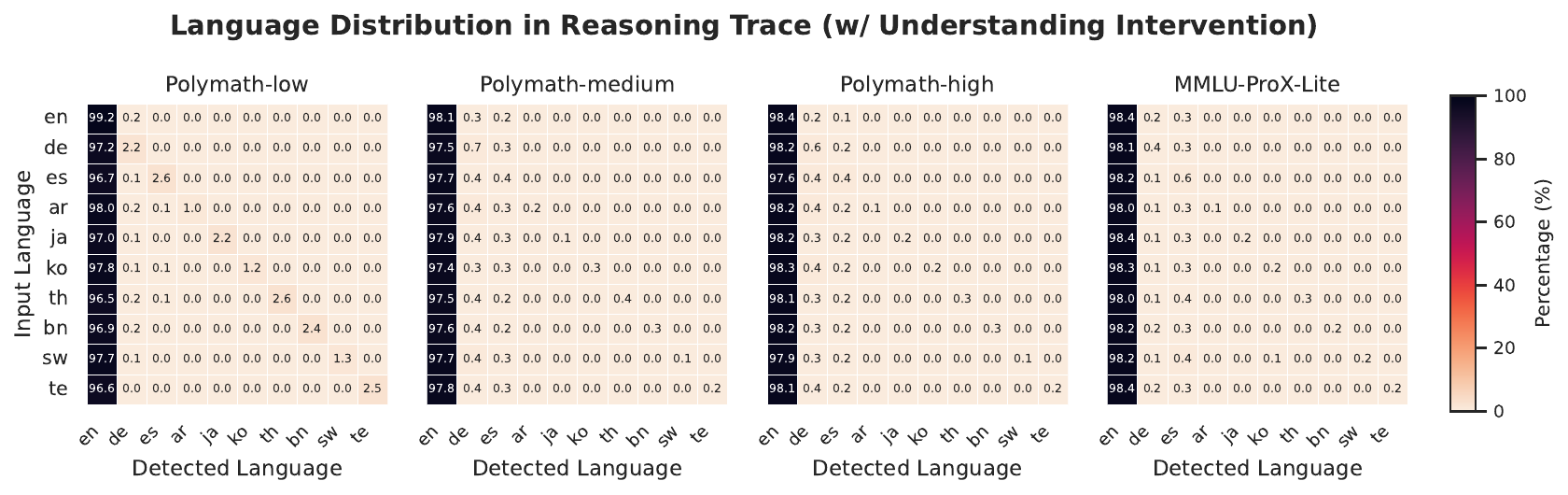}
    \end{subfigure}
    \begin{subfigure}{\textwidth}
        \centering
        \includegraphics[width=1.0\textwidth]{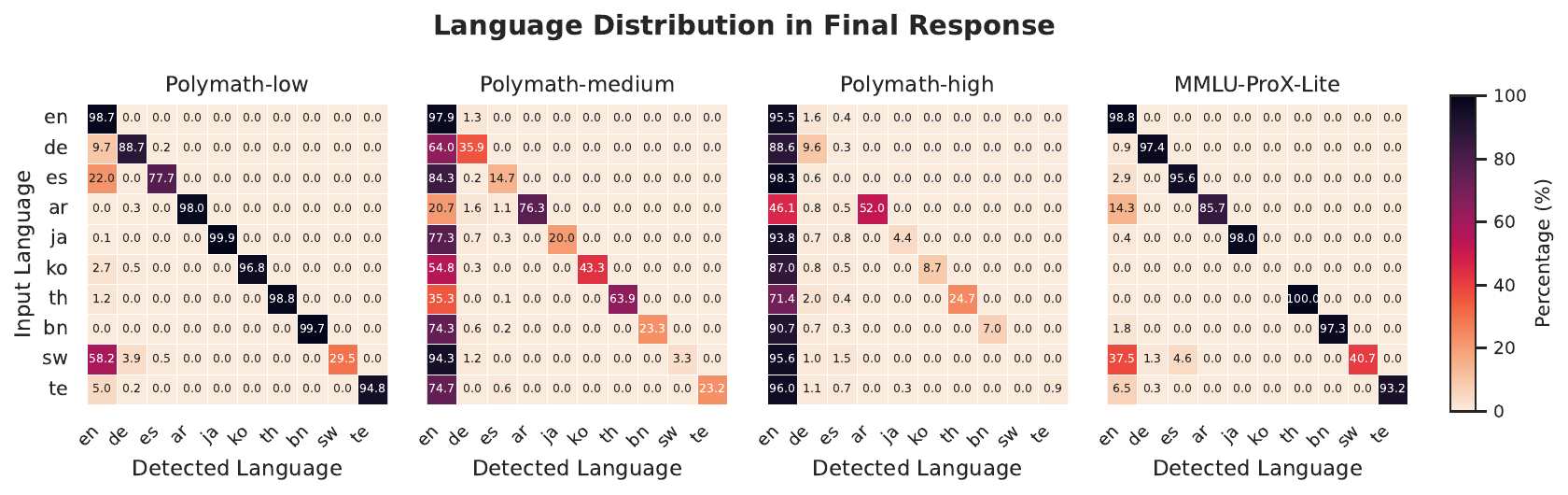}
    \end{subfigure}
    \begin{subfigure}{\textwidth}
        \centering
        \includegraphics[width=1.0\textwidth]{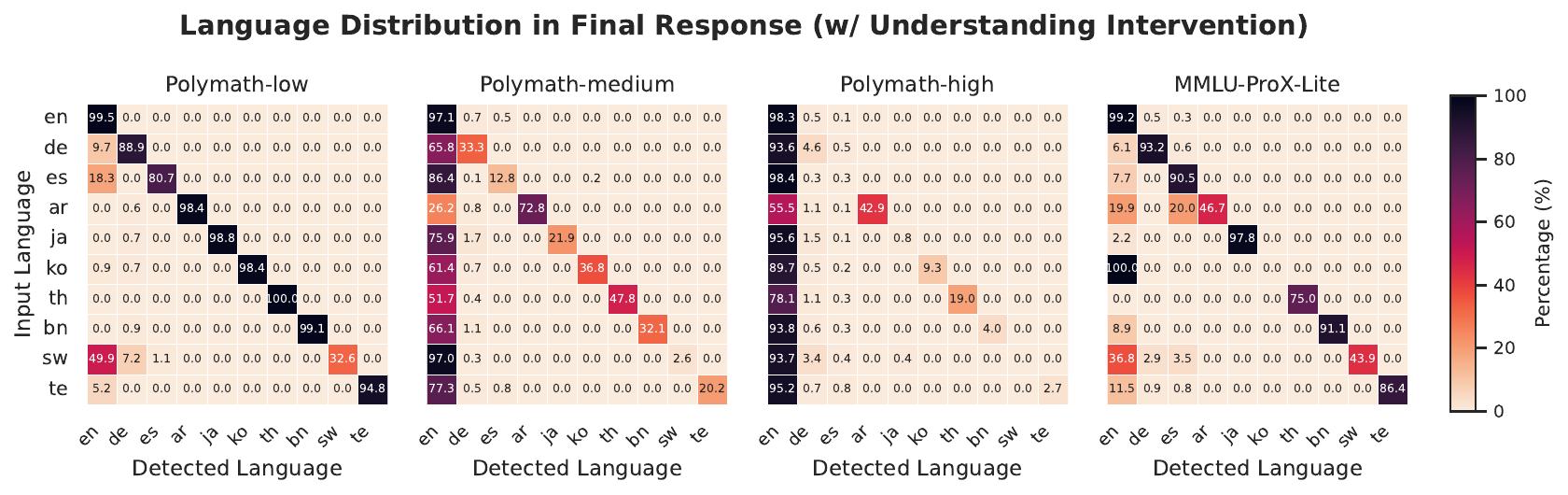}
    \end{subfigure}
    \caption{Language distributions of reasoning traces and final responses for Qwen3-14B across Polymath (low/medium/high) and MMLU-ProX-Lite datasets, with and without the understanding intervention. }
    \label{fig:ld_14b}
\end{figure*}

We evaluate five large reasoning models—Qwen3-1.7B, Qwen3-4B, Qwen3-8B, Qwen3-14B, and gpt-oss-20b—across four datasets: Polymath-Low, Polymath-Medium, Polymath-High, and MMLU-ProX-Lite.
For each model, we visualize the language distribution of reasoning traces and final responses, including the corresponding variants obtained under the \textit{Understanding Intervention}.
Figures~\ref{fig:ld_4b}–\ref{fig:ld_14b} present the results.
\textbf{Across all settings, reasoning traces are consistently dominated by English, whereas final responses tend to be produced in the input language.} While the proportion of the input language in final responses varies across languages and datasets, a substantial portion of responses are still generated in the input language. This tendency remains largely unchanged after applying the \textit{Understanding Intervention}.

While reasoning traces remain English-dominant across all settings, the proportion of final responses in the input language is higher in Polymath-Low and MMLU-ProX-Lite, but relatively lower in Polymath-Medium and Polymath-High.
We hypothesize that this pattern arises from two factors: increased reasoning difficulty and the higher density of mathematical expressions in the more challenging splits. The first factor concerns reasoning difficulty.~\citet{wang2025language} shows that as task difficulty increases, reasoning language models rely more on the Latin script in their internal reasoning representations.
This increased reliance on Latin-based internal reasoning—presumably English—may influence the language distribution of final responses. The second factor relates to the density of mathematical expressions. After removing LaTeX and numerical expressions, the share of pure natural language content is 88.13\% for Low, 46.29\% for Medium, and 60.69\% for High, indicating that the latter two splits are more expression-heavy. Although we mitigate such effects in our language distribution measurement using regular-expression filtering, equations and variable-like expressions occasionally appear in plain text, leading to potential false detections as English due to alphabetic variable names. Future work may  explore why such language shifts occur in final responses.

\section{Additional Evaluation Results}
\subsection{Intervention results on other models}\label{sec:appendix_other_model_intv_res}
\begin{table*}[!t]
\centering
\footnotesize
\resizebox{\textwidth}{!}{
\begin{tabular}{l|l|cccccc|cc|cc|c}
\toprule
\multicolumn{13}{c}{\textbf{Qwen3-1.7B}}\\
\midrule
Dataset & Method & en & de & es & ar & ja & ko & th & bn & sw & te & \textbf{Avg} \\
\midrule
\multirow{4}{*}{Polymath-Low} & Base & 90.1 & 78.7 & 85.6 & 79.7 & 71.5 & 72.3 & 76.8 & 60.3 & 7.2 & 41.1 & \textbf{66.3} \\
 & w/ T & 90.1 & 78.1 & 85.6 & 78.1 & 70.9 & 70.1 & 76.3 & 60.8 & 8.3 & \cellcolor[rgb]{0.88, 0.95, 0.88}42.7 & \textbf{66.1} \\
 & w/ U & 90.7 & \cellcolor[rgb]{0.75 0.88 0.98}82.4 & 87.5 & \cellcolor[rgb]{0.88, 0.95, 0.88}84.8 & \cellcolor[rgb]{0.75 0.88 0.98}81.1 & \cellcolor[rgb]{0.75 0.88 0.98}85.1 & \cellcolor[rgb]{0.75 0.88 0.98}85.3 & \cellcolor[rgb]{0.75 0.88 0.98}83.5 & \cellcolor[rgb]{0.75 0.88 0.98}67.7 & \cellcolor[rgb]{0.75 0.88 0.98}78.1 & \cellcolor[rgb]{0.75 0.88 0.98}\textbf{82.6} \\
 & w/ U+T & 90.1 & \cellcolor[rgb]{0.75 0.88 0.98}82.7 & 87.5 & 84.0 & \cellcolor[rgb]{0.75 0.88 0.98}81.3 & \cellcolor[rgb]{0.75 0.88 0.98}84.8 & \cellcolor[rgb]{0.75 0.88 0.98}84.8 & \cellcolor[rgb]{0.75 0.88 0.98}85.6 & \cellcolor[rgb]{0.75 0.88 0.98}79.5 & \cellcolor[rgb]{0.75 0.88 0.98}83.7 & \cellcolor[rgb]{0.75 0.88 0.98}\textbf{84.4} \\
\midrule
\multirow{4}{*}{Polymath-Medium} & Base & 55.7 & 54.9 & 56.8 & 53.3 & 57.6 & 50.7 & 48.8 & 47.2 & 32.3 & 40.3 & \textbf{49.8} \\
 & w/ T & 56.3 & 54.4 & 57.3 & 53.6 & 56.8 & 51.5 & 49.3 & 48.8 & 32.0 & 40.8 & \textbf{50.1} \\
 & w/ U & 58.7 & 53.6 & 55.7 & 55.2 & 57.1 & 54.4 & 56.5 & \cellcolor[rgb]{0.75 0.88 0.98}55.2 & \cellcolor[rgb]{0.75 0.88 0.98}55.5 & \cellcolor[rgb]{0.75 0.88 0.98}57.9 & \cellcolor[rgb]{0.75 0.88 0.98}\textbf{56.0} \\
 & w/ U+T & 59.5 & 53.9 & 55.7 & 55.5 & 56.5 & \cellcolor[rgb]{0.75 0.88 0.98}55.2 & 56.5 & \cellcolor[rgb]{0.75 0.88 0.98}54.9 & \cellcolor[rgb]{0.75 0.88 0.98}55.5 & \cellcolor[rgb]{0.75 0.88 0.98}58.1 & \cellcolor[rgb]{0.75 0.88 0.98}\textbf{56.1} \\
\midrule
\multirow{4}{*}{Polymath-High} & Base & 33.3 & 32.8 & 32.5 & 29.6 & 33.3 & 30.9 & 28.8 & 22.4 & 14.4 & 21.3 & \textbf{27.9} \\
 & w/ T & 34.1 & 32.0 & 32.8 & 28.8 & 33.6 & 31.5 & 29.1 & 22.4 & 14.4 & 22.1 & \textbf{28.1} \\
 & w/ U & 33.1 & 33.3 & 36.3 & \cellcolor[rgb]{0.75 0.88 0.98}34.1 & 32.0 & 33.1 & 32.0 & \cellcolor[rgb]{0.75 0.88 0.98}31.7 & \cellcolor[rgb]{0.75 0.88 0.98}28.8 & \cellcolor[rgb]{0.75 0.88 0.98}31.7 & \cellcolor[rgb]{0.75 0.88 0.98}\textbf{32.6} \\
 & w/ U+T & 33.1 & 33.1 & 35.5 & \cellcolor[rgb]{0.75 0.88 0.98}34.4 & 31.7 & 32.5 & 32.0 & \cellcolor[rgb]{0.75 0.88 0.98}32.5 & \cellcolor[rgb]{0.75 0.88 0.98}28.8 & \cellcolor[rgb]{0.75 0.88 0.98}31.7 & \cellcolor[rgb]{0.75 0.88 0.98}\textbf{32.5} \\
\midrule
\multirow{4}{*}{MMLU-ProX-Lite} & Base & 67.6 & 63.2 & 65.6 & 56.4 & 61.0 & 57.1 & 59.4 & 52.4 & 31.8 & 52.5 & \textbf{56.7} \\
 & w/ T & 67.2 & 62.8 & 65.2 & \cellcolor[rgb]{0.88, 0.95, 0.88}57.2 & 60.6 & 57.5 & 60.3 & 53.3 & 32.2 & 53.3 & \cellcolor[rgb]{0.75 0.88 0.98}\textbf{57.0} \\
 & w/ U & 64.9 & 65.2 & 64.6 & \cellcolor[rgb]{0.75 0.88 0.98}63.8 & \cellcolor[rgb]{0.88, 0.95, 0.88}63.9 & \cellcolor[rgb]{0.75 0.88 0.98}64.9 & \cellcolor[rgb]{0.88, 0.95, 0.88}63.2 & \cellcolor[rgb]{0.75 0.88 0.98}63.2 & \cellcolor[rgb]{0.75 0.88 0.98}52.4 & \cellcolor[rgb]{0.75 0.88 0.98}62.3 & \cellcolor[rgb]{0.75 0.88 0.98}\textbf{62.8} \\
 & w/ U+T & 65.0 & 65.1 & 64.6 & \cellcolor[rgb]{0.75 0.88 0.98}63.8 & \cellcolor[rgb]{0.75 0.88 0.98}64.1 & \cellcolor[rgb]{0.75 0.88 0.98}64.5 & \cellcolor[rgb]{0.75 0.88 0.98}63.4 & \cellcolor[rgb]{0.75 0.88 0.98}63.8 & \cellcolor[rgb]{0.75 0.88 0.98}60.2 & \cellcolor[rgb]{0.75 0.88 0.98}64.2 & \cellcolor[rgb]{0.75 0.88 0.98}\textbf{63.9} \\
\midrule
\multicolumn{13}{c}{\textbf{Qwen3-8B}}\\
\midrule
Dataset & Method & en & de & es & ar & ja & ko & th & bn & sw & te & \textbf{Avg} \\
\midrule
\multirow{4}{*}{Polymath-Low} & Base & 96.3 & 88.5 & 93.9 & 90.9 & 86.1 & 90.4 & 90.7 & 86.4 & 61.3 & 76.8 & \textbf{86.1} \\
 & w/ T & 96.3 & 88.5 & 94.1 & 90.9 & 85.9 & 90.7 & 90.7 & \cellcolor[rgb]{0.88, 0.95, 0.88}87.5 & 62.1 & 76.5 & \textbf{86.3} \\
 & w/ U & 96.5 & 89.3 & 96.3 & \cellcolor[rgb]{0.75 0.88 0.98}94.4 & \cellcolor[rgb]{0.75 0.88 0.98}90.9 & \cellcolor[rgb]{0.75 0.88 0.98}94.4 & \cellcolor[rgb]{0.75 0.88 0.98}94.4 & \cellcolor[rgb]{0.75 0.88 0.98}95.5 & \cellcolor[rgb]{0.75 0.88 0.98}93.3 & \cellcolor[rgb]{0.75 0.88 0.98}93.1 & \cellcolor[rgb]{0.75 0.88 0.98}\textbf{93.8} \\
 & w/ U+T & 96.3 & 89.3 & 96.0 & 93.3 & \cellcolor[rgb]{0.75 0.88 0.98}90.9 & \cellcolor[rgb]{0.88, 0.95, 0.88}93.9 & \cellcolor[rgb]{0.75 0.88 0.98}94.4 & \cellcolor[rgb]{0.75 0.88 0.98}94.7 & \cellcolor[rgb]{0.75 0.88 0.98}94.1 & \cellcolor[rgb]{0.75 0.88 0.98}92.0 & \cellcolor[rgb]{0.75 0.88 0.98}\textbf{93.5} \\
\midrule
\multirow{4}{*}{Polymath-Medium} & Base & 72.5 & 74.4 & 73.9 & 70.4 & 72.5 & 71.5 & 70.7 & 68.3 & 54.4 & 65.3 & \textbf{69.4} \\
 & w/ T & 73.6 & 74.1 & 74.7 & 70.4 & 72.5 & 72.5 & 70.1 & 68.3 & 54.7 & 66.4 & \textbf{69.7} \\
 & w/ U & 74.7 & 72.5 & 72.8 & 72.5 & 74.7 & 73.9 & 72.8 & \cellcolor[rgb]{0.75 0.88 0.98}74.4 & \cellcolor[rgb]{0.75 0.88 0.98}71.7 & \cellcolor[rgb]{0.75 0.88 0.98}74.9 & \cellcolor[rgb]{0.75 0.88 0.98}\textbf{73.5} \\
 & w/ U+T & 73.9 & 72.3 & 72.3 & 73.1 & 74.1 & 74.1 & 72.5 & \cellcolor[rgb]{0.88, 0.95, 0.88}74.1 & \cellcolor[rgb]{0.75 0.88 0.98}72.0 & \cellcolor[rgb]{0.75 0.88 0.98}74.4 & \cellcolor[rgb]{0.75 0.88 0.98}\textbf{73.3} \\
\midrule
\multirow{4}{*}{Polymath-High} & Base & 54.4 & 54.1 & 54.4 & 52.8 & 57.1 & 54.7 & 51.5 & 50.1 & 38.1 & 45.9 & \textbf{51.3} \\
 & w/ T & 55.5 & 54.4 & 54.4 & 51.7 & 56.8 & 54.4 & 51.5 & 51.5 & 37.9 & 45.9 & \textbf{51.4} \\
 & w/ U & 53.9 & 56.3 & 51.7 & 54.4 & 53.6 & 54.9 & 52.8 & 52.0 & \cellcolor[rgb]{0.75 0.88 0.98}54.4 & \cellcolor[rgb]{0.75 0.88 0.98}51.2 & \cellcolor[rgb]{0.75 0.88 0.98}\textbf{53.5} \\
 & w/ U+T & 54.1 & 56.3 & 51.7 & 54.1 & 53.9 & 55.5 & 53.3 & 52.0 & \cellcolor[rgb]{0.75 0.88 0.98}54.4 & \cellcolor[rgb]{0.75 0.88 0.98}52.3 & \cellcolor[rgb]{0.75 0.88 0.98}\textbf{53.8} \\
\midrule
\multirow{4}{*}{MMLU-ProX-Lite} & Base & 82.9 & 79.5 & 80.8 & 79.5 & 79.1 & 78.3 & 79.4 & 77.0 & 58.5 & 76.0 & \textbf{77.1} \\
 & w/ T & 82.4 & 79.6 & 81.6 & 79.5 & 79.9 & 78.6 & 79.8 & 77.3 & \cellcolor[rgb]{0.88, 0.95, 0.88}61.5 & 76.5 & \textbf{77.7} \\
 & w/ U & 82.2 & 81.1 & 80.5 & \cellcolor[rgb]{0.75 0.88 0.98}82.0 & \cellcolor[rgb]{0.88, 0.95, 0.88}81.7 & \cellcolor[rgb]{0.75 0.88 0.98}81.3 & \cellcolor[rgb]{0.75 0.88 0.98}81.6 & \cellcolor[rgb]{0.75 0.88 0.98}80.5 & \cellcolor[rgb]{0.75 0.88 0.98}79.0 & \cellcolor[rgb]{0.75 0.88 0.98}79.9 & \cellcolor[rgb]{0.75 0.88 0.98}\textbf{81.0} \\
 & w/ U+T & 82.0 & 80.9 & 80.7 & \cellcolor[rgb]{0.75 0.88 0.98}81.7 & \cellcolor[rgb]{0.88, 0.95, 0.88}82.4 & \cellcolor[rgb]{0.88, 0.95, 0.88}81.5 & \cellcolor[rgb]{0.75 0.88 0.98}82.1 & \cellcolor[rgb]{0.75 0.88 0.98}80.3 & \cellcolor[rgb]{0.75 0.88 0.98}81.7 & \cellcolor[rgb]{0.75 0.88 0.98}80.5 & \cellcolor[rgb]{0.75 0.88 0.98}\textbf{81.4} \\
\bottomrule
\end{tabular}
}
\caption{
Accuracy comparison of methods across languages on additional models (Qwen3-1.7B, Qwen3-8B).
\textbf{Base} denotes the original model without intervention, \textbf{w/ T} applies Answer Extraction from Reasoning Trace, \textbf{w/ U} applies Understanding Intervention, and \textbf{w/ U + T} applies both.
Pairwise comparisons against the Base use Welch’s t-test: \colorbox[rgb]{0.75,0.88,0.98}{blue} cells mark statistically significant improvements (p < 0.05), and \colorbox[rgb]{0.88,0.95,0.88}{green} cells indicate notable improvements (p < 0.1). The Average row is \textbf{bolded}.
Consistent with Qwen3-4B (Table~\ref{tab:main_result}), understanding intervention yields substantial gains, particularly for low-resource languages.
}
\label{tab:intv_additional_results_1}
\end{table*}

\begin{table*}[!t]
\centering
\footnotesize
\resizebox{\textwidth}{!}{
\begin{tabular}{l|l|cccccc|cc|cc|c}
\toprule
\multicolumn{13}{c}{\textbf{Qwen3-14B}}\\
\midrule
Dataset & Method & en & de & es & ar & ja & ko & th & bn & sw & te & \textbf{Avg} \\
\midrule
\multirow{4}{*}{Polymath-Low} & Base & 95.2 & 89.6 & 93.6 & 90.4 & 89.3 & 89.3 & 91.7 & 89.3 & 73.3 & 82.1 & \textbf{88.4} \\
 & w/ T & 95.5 & 87.2 & 93.3 & 89.9 & 89.3 & 87.2 & 89.9 & 88.8 & 72.3 & 81.6 & \textbf{87.5} \\
 & w/ U & 95.7 & \cellcolor[rgb]{0.88, 0.95, 0.88}92.3 & \cellcolor[rgb]{0.75 0.88 0.98}95.5 & \cellcolor[rgb]{0.88, 0.95, 0.88}93.9 & \cellcolor[rgb]{0.75 0.88 0.98}91.5 & 93.3 & \cellcolor[rgb]{0.88, 0.95, 0.88}95.2 & \cellcolor[rgb]{0.75 0.88 0.98}94.7 & \cellcolor[rgb]{0.75 0.88 0.98}92.0 & \cellcolor[rgb]{0.75 0.88 0.98}94.1 & \cellcolor[rgb]{0.75 0.88 0.98}\textbf{93.8} \\
 & w/ U+T & 95.7 & \cellcolor[rgb]{0.88, 0.95, 0.88}92.8 & \cellcolor[rgb]{0.75 0.88 0.98}95.7 & 92.3 & 90.4 & 92.3 & \cellcolor[rgb]{0.88, 0.95, 0.88}94.1 & \cellcolor[rgb]{0.88, 0.95, 0.88}93.3 & \cellcolor[rgb]{0.75 0.88 0.98}91.5 & \cellcolor[rgb]{0.75 0.88 0.98}94.4 & \cellcolor[rgb]{0.75 0.88 0.98}\textbf{93.3} \\
\midrule
\multirow{4}{*}{Polymath-Medium} & Base & 76.0 & 72.8 & 78.1 & 73.6 & 76.0 & 74.7 & 72.0 & 69.1 & 62.9 & 70.1 & \textbf{72.5} \\
 & w/ T & 76.5 & 73.9 & 78.7 & 73.6 & 76.0 & 75.7 & 72.0 & 69.1 & 63.2 & 70.9 & \textbf{73.0} \\
 & w/ U & 77.3 & 77.6 & 77.9 & 73.9 & 76.5 & 75.2 & \cellcolor[rgb]{0.75 0.88 0.98}76.0 & \cellcolor[rgb]{0.75 0.88 0.98}77.9 & \cellcolor[rgb]{0.75 0.88 0.98}76.3 & \cellcolor[rgb]{0.88, 0.95, 0.88}76.8 & \cellcolor[rgb]{0.75 0.88 0.98}\textbf{76.5} \\
 & w/ U+T & 77.3 & 77.1 & 77.3 & 73.9 & 76.5 & 75.5 & \cellcolor[rgb]{0.75 0.88 0.98}76.0 & \cellcolor[rgb]{0.75 0.88 0.98}76.8 & \cellcolor[rgb]{0.75 0.88 0.98}75.5 & \cellcolor[rgb]{0.75 0.88 0.98}78.9 & \cellcolor[rgb]{0.75 0.88 0.98}\textbf{76.5} \\
\midrule
\multirow{4}{*}{Polymath-High} & Base & 60.3 & 60.0 & 59.5 & 58.4 & 58.7 & 56.5 & 55.5 & 56.0 & 45.1 & 51.7 & \textbf{56.2} \\
 & w/ T & 59.2 & 58.9 & 58.7 & 57.9 & 57.6 & 55.2 & 53.3 & 54.7 & 43.7 & 50.9 & \textbf{55.0} \\
 & w/ U & 60.3 & 57.3 & 59.5 & 57.6 & 61.1 & 58.1 & 57.3 & 58.1 & \cellcolor[rgb]{0.75 0.88 0.98}58.4 & 55.2 & \cellcolor[rgb]{0.88, 0.95, 0.88}\textbf{58.3} \\
 & w/ U+T & 60.0 & 56.5 & 59.5 & 56.8 & 59.7 & 58.4 & 57.1 & 56.8 & \cellcolor[rgb]{0.75 0.88 0.98}57.6 & 54.1 & \textbf{57.7} \\
\midrule
\multirow{4}{*}{MMLU-ProX-Lite} & Base & 82.9 & 80.8 & 80.8 & 81.2 & 80.9 & 80.5 & 79.2 & 81.1 & 69.0 & 79.9 & \textbf{79.6} \\
 & w/ T & 83.7 & 81.3 & 80.5 & 81.5 & 81.1 & 80.3 & 79.4 & 80.9 & 69.1 & 79.8 & \textbf{79.8} \\
 & w/ U & 83.1 & 82.0 & 81.5 & \cellcolor[rgb]{0.88, 0.95, 0.88}83.3 & 80.7 & 82.5 & 81.3 & 81.5 & \cellcolor[rgb]{0.75 0.88 0.98}81.5 & 80.5 & \cellcolor[rgb]{0.75 0.88 0.98}\textbf{81.8} \\
 & w/ U+T & 83.7 & 82.0 & 81.5 & \cellcolor[rgb]{0.88, 0.95, 0.88}83.3 & 80.8 & 82.4 & 81.2 & 81.2 & \cellcolor[rgb]{0.75 0.88 0.98}81.5 & 80.8 & \cellcolor[rgb]{0.75 0.88 0.98}\textbf{81.8} \\
\midrule
\multicolumn{13}{c}{\textbf{gpt-oss-20b}}\\
\midrule
Dataset & Method & en & de & es & ar & ja & ko & th & bn & sw & te & \textbf{Avg} \\
\midrule
\multirow{4}{*}{Polymath-Low} & Base & 96.0 & 85.6 & 89.9 & 90.9 & 86.1 & 91.2 & 90.4 & 90.1 & 78.7 & 81.1 & \textbf{88.0} \\
 & w/ T & 95.7 & 86.7 & \cellcolor[rgb]{0.88, 0.95, 0.88}92.0 & 91.7 & 86.4 & 90.7 & 90.7 & 90.7 & 80.0 & 81.1 & \cellcolor[rgb]{0.88, 0.95, 0.88}\textbf{88.6} \\
 & w/ U & 95.7 & 84.5 & \cellcolor[rgb]{0.88, 0.95, 0.88}92.3 & 91.2 & 87.5 & 92.5 & 89.3 & 92.3 & \cellcolor[rgb]{0.75 0.88 0.98}88.8 & \cellcolor[rgb]{0.75 0.88 0.98}89.6 & \cellcolor[rgb]{0.88, 0.95, 0.88}\textbf{90.4} \\
 & w/ U+T & 91.7 & 83.7 & 91.7 & 92.3 & 87.2 & 92.0 & 88.8 & 91.2 & \cellcolor[rgb]{0.75 0.88 0.98}89.1 & 87.7 & \textbf{89.5} \\
\midrule
\multirow{4}{*}{Polymath-Medium} & Base & 67.7 & 65.1 & 68.5 & 63.7 & 65.3 & 67.2 & 66.9 & 66.9 & 58.9 & 65.3 & \textbf{65.6} \\
 & w/ T & 61.9 & 60.8 & 58.7 & 58.1 & 62.9 & 63.7 & 61.3 & 61.3 & 52.3 & 61.6 & \textbf{60.3} \\
 & w/ U & 66.1 & 66.1 & 66.4 & \cellcolor[rgb]{0.75 0.88 0.98}66.7 & 68.5 & 69.1 & 65.6 & 65.6 & \cellcolor[rgb]{0.75 0.88 0.98}64.0 & 68.3 & \textbf{66.6} \\
 & w/ U+T & 69.6 & \cellcolor[rgb]{0.88, 0.95, 0.88}68.8 & 67.7 & \cellcolor[rgb]{0.75 0.88 0.98}68.3 & \cellcolor[rgb]{0.75 0.88 0.98}69.9 & \cellcolor[rgb]{0.88, 0.95, 0.88}72.3 & 65.3 & 68.0 & \cellcolor[rgb]{0.75 0.88 0.98}66.1 & 69.6 & \cellcolor[rgb]{0.75 0.88 0.98}\textbf{68.6} \\
\midrule
\multirow{4}{*}{Polymath-High} & Base & 58.1 & 57.3 & 54.7 & 52.8 & 51.5 & 53.9 & 53.1 & 54.1 & 47.5 & 56.3 & \textbf{53.9} \\
 & w/ T & 54.4 & 52.0 & 53.6 & 50.9 & 52.5 & 52.5 & 51.2 & 54.1 & 47.2 & 51.5 & \textbf{52.0} \\
 & w/ U & 57.1 & 57.1 & 56.8 & 53.9 & \cellcolor[rgb]{0.88, 0.95, 0.88}56.8 & 58.9 & \cellcolor[rgb]{0.75 0.88 0.98}58.9 & 57.3 & \cellcolor[rgb]{0.75 0.88 0.98}54.9 & 60.0 & \cellcolor[rgb]{0.75 0.88 0.98}\textbf{57.2} \\
 & w/ U+T & 58.1 & 59.5 & 59.2 & 56.0 & \cellcolor[rgb]{0.88, 0.95, 0.88}57.3 & \cellcolor[rgb]{0.88, 0.95, 0.88}61.3 & \cellcolor[rgb]{0.88, 0.95, 0.88}60.8 & 59.2 & \cellcolor[rgb]{0.75 0.88 0.98}57.1 & \cellcolor[rgb]{0.88, 0.95, 0.88}60.3 & \cellcolor[rgb]{0.75 0.88 0.98}\textbf{58.9} \\
\midrule
\multirow{4}{*}{MMLU-ProX-Lite} & Base & 77.4 & 77.8 & 77.3 & 74.1 & 77.4 & 73.9 & 74.6 & 75.6 & 66.9 & 77.0 & \textbf{75.2} \\
 & w/ T & 77.3 & 77.7 & 77.3 & 74.1 & 77.6 & 74.1 & 74.4 & 75.6 & 66.9 & 77.2 & \textbf{75.2} \\
 & w/ U & \cellcolor[rgb]{0.75 0.88 0.98}78.6 & 77.0 & 79.0 & 78.3 & 77.3 & \cellcolor[rgb]{0.75 0.88 0.98}77.2 & 76.4 & 77.4 & \cellcolor[rgb]{0.75 0.88 0.98}77.0 & 76.8 & \cellcolor[rgb]{0.75 0.88 0.98}\textbf{77.5} \\
 & w/ U+T & \cellcolor[rgb]{0.88, 0.95, 0.88}78.1 & 76.8 & 78.3 & 77.7 & 76.7 & \cellcolor[rgb]{0.75 0.88 0.98}76.9 & 75.5 & 76.9 & \cellcolor[rgb]{0.75 0.88 0.98}77.2 & 75.7 & \cellcolor[rgb]{0.75 0.88 0.98}\textbf{77.0} \\
\bottomrule
\end{tabular}
}
\caption{
Accuracy comparison of methods across languages on Qwen3-14B, and gpt-oss-20b.
\textbf{Base} denotes the original model without intervention, \textbf{w/ T} applies Answer Extraction from Reasoning Trace, \textbf{w/ U} applies Understanding Intervention, and \textbf{w/ U + T} applies both.
Pairwise comparisons against the Base use Welch’s t-test: \colorbox[rgb]{0.75,0.88,0.98}{blue} cells mark statistically significant improvements (p < 0.05), and \colorbox[rgb]{0.88,0.95,0.88}{green} cells indicate notable improvements (p < 0.1). The Average row is \textbf{bolded}.
Consistent with Qwen3-4B (Table~\ref{tab:main_result}), understanding intervention yields substantial gains, particularly for low-resource languages.
}
\label{tab:intv_additional_results_2}
\end{table*}

Table~\ref{tab:intv_additional_results_1}, ~\ref{tab:intv_additional_results_2} extends the analysis in Table~\ref{tab:main_result} to additional reasoning language models, including Qwen3-1.7B, Qwen3-8B, Qwen3-14B, and gpt-oss-20b.
Across all models, the Understanding Intervention (w/ U) consistently yields notable accuracy gains over the Base model, particularly in low-resource languages. In contrast, the Answer Extraction from Reasoning Trace (w/ T) method shows minimal impact, with performance remaining close to the Base setting.\footnote{One exception is observed for gpt-oss-20b on Polymath-Medium, where performance slightly decreases. Through manual inspection of several sampled cases, we found that answers marked incorrect in reasoning traces—but correct in final responses—were in fact semantically equivalent to the ground truth, suggesting that the observed drop results from phrasing noise in automatic verification rather than a genuine reasoning error.} 
\subsection{Language-specific Stage-wise Attribution Analysis}\label{sec:appendix_language_gap}
\begin{figure*}[!ht]
  \centering
  \includegraphics[width=\textwidth]{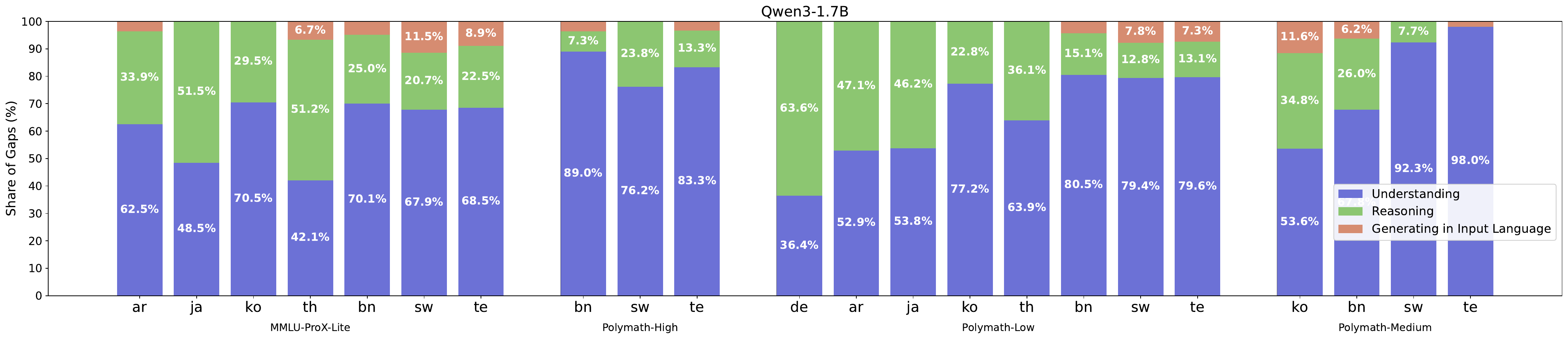}
  \caption{Language-specific Stage-wise Attribution Analysis for Qwen3-1.7B.}
  \label{fig:lang-gap-qwen1.7b}
\end{figure*}

\begin{figure*}[!ht]
  \centering
  \includegraphics[width=\textwidth]{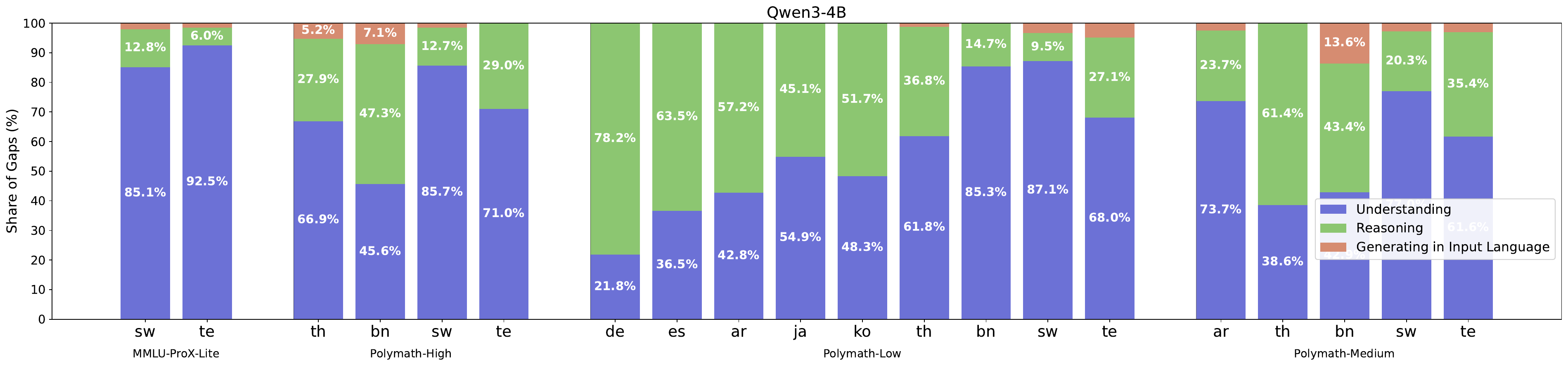}
  \caption{Language-specific Stage-wise Attribution Analysis for Qwen3-4B.}
  \label{fig:lang-gap-qwen4b}
\end{figure*}

\begin{figure*}[!ht]
  \centering
  \includegraphics[width=\textwidth]{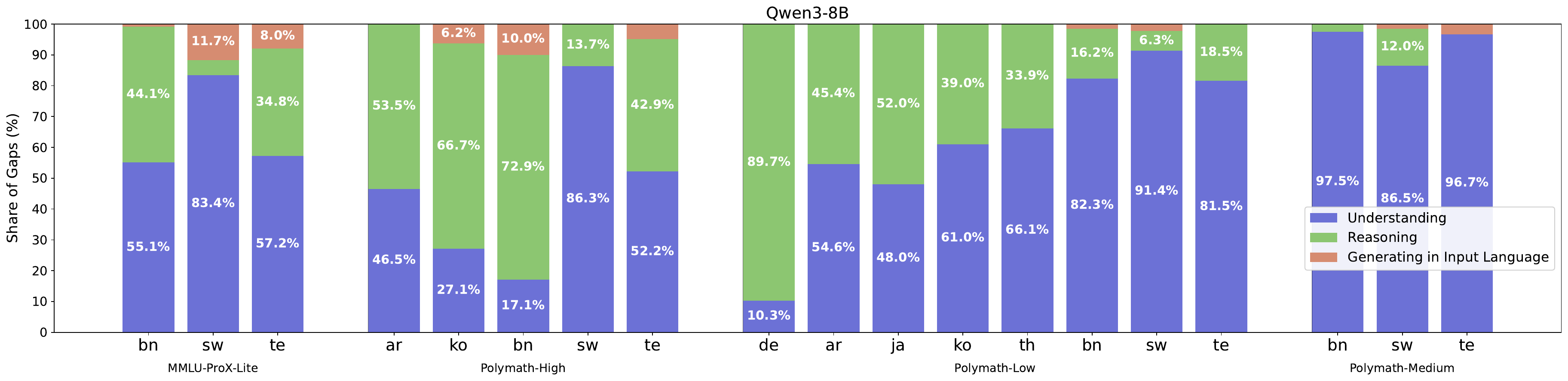}
  \caption{Language-specific Stage-wise Attribution Analysis for Qwen3-8B.}
  \label{fig:lang-gap-qwen8b}
\end{figure*}

\begin{figure*}[!ht]
  \centering
  \includegraphics[width=\textwidth]{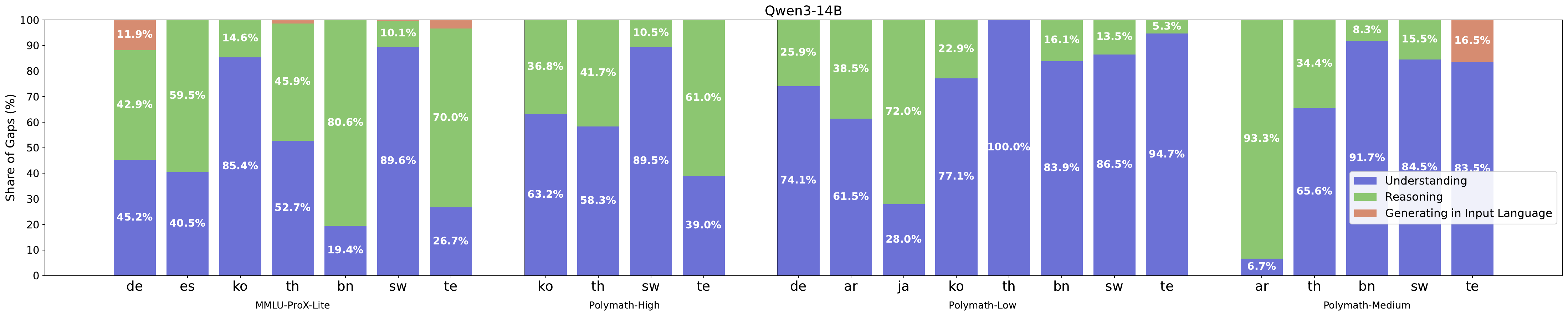}
  \caption{Language-specific Stage-wise Attribution Analysis for Qwen3-14B.}
  \label{fig:lang-gap-qwen14b}
\end{figure*}

\begin{figure*}[!ht]
  \centering
  \includegraphics[width=\textwidth]{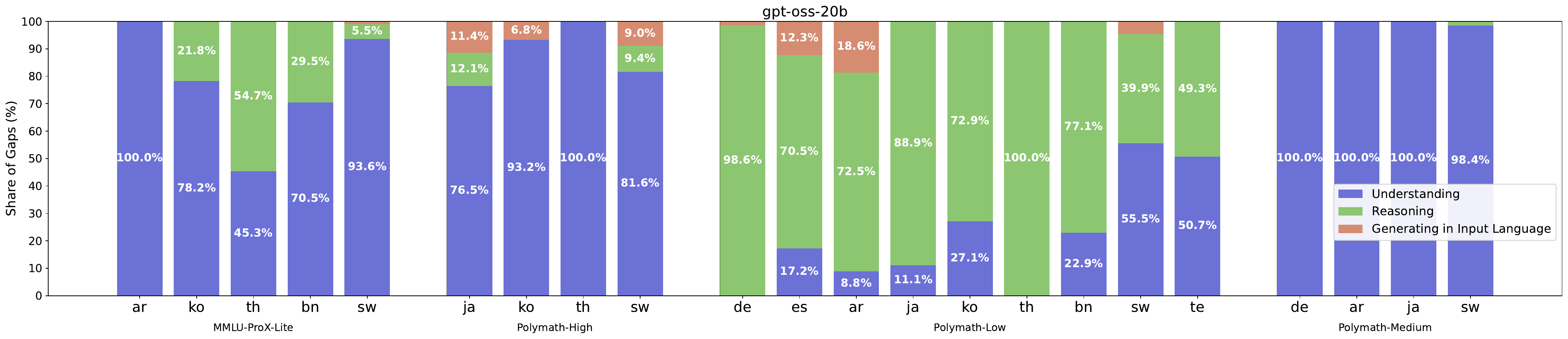}
  \caption{Language-specific Stage-wise Attribution Analysis for gpt-oss-20b.}
  \label{fig:lang-gap-oss}
\end{figure*}

Figures~\ref{fig:lang-gap-qwen1.7b}, \ref{fig:lang-gap-qwen4b}, \ref{fig:lang-gap-qwen8b}, \ref{fig:lang-gap-qwen14b}, and \ref{fig:lang-gap-oss} 
present the language-specific stage-wise attribution analysis results for Qwen3-1.7B, Qwen3-4B, Qwen3-8B, Qwen3-14B, and gpt-oss-20b, respectively. 
Each figure breaks down the multilingual reasoning gap into contributions from \emph{Understanding}, \emph{Reasoning}, and \emph{Generating in Input Language} for each language. 
Across all models, \emph{Understanding} failures consistently emerge as the primary source of the gap. 
However, their relative dominance varies with language resource level. The effect is particularly pronounced in low-resource languages such as Swahili (sw) and Telugu (te), whereas in high-resource languages, where models already achieve strong Base performance, the influence of \emph{Understanding} becomes relatively smaller.

\begin{figure*}[!th]
  \centering
  \includegraphics[width=\textwidth]{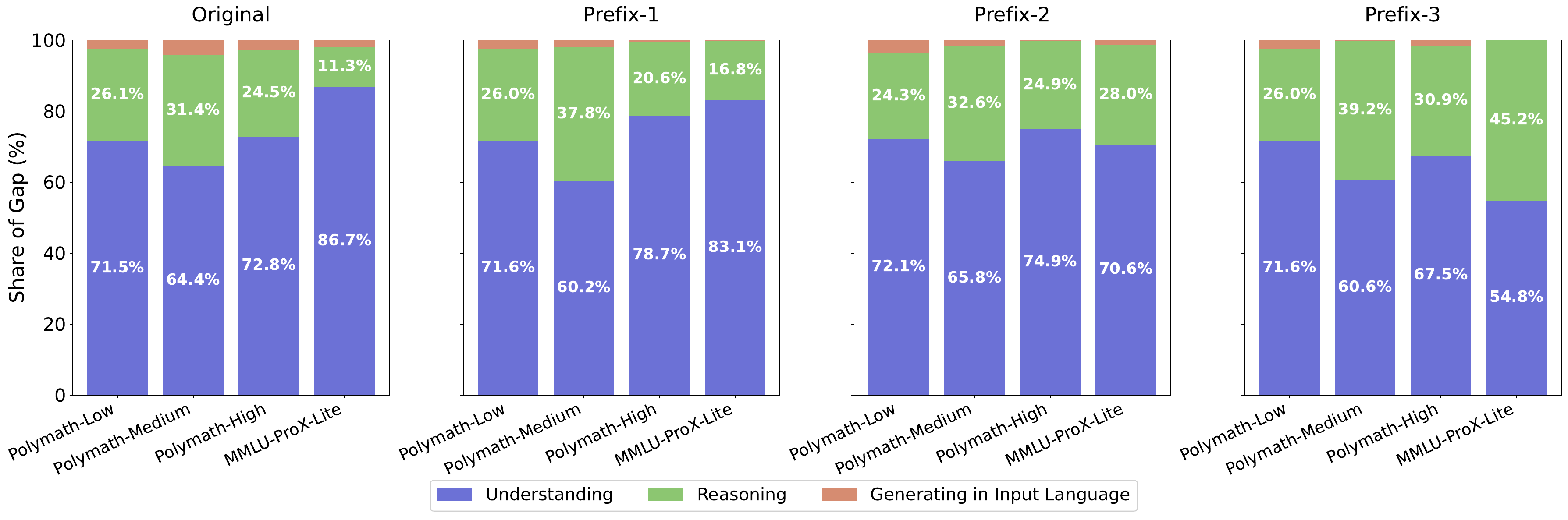}
  \caption{Weighted shares of \textit{Understanding}, \textit{Reasoning}, and \textit{Generation} in the input language to the overall multilingual reasoning gap on Qwen3-4B. Across different prefix variants, failures in \textit{Understanding} dominate the gap.}
  \label{fig:variant_gap_agg}
\end{figure*}

\subsection{Extending Attribution Analysis to Additional Models}
\begin{figure}[!th]
  \centering
  \includegraphics[width=\columnwidth]{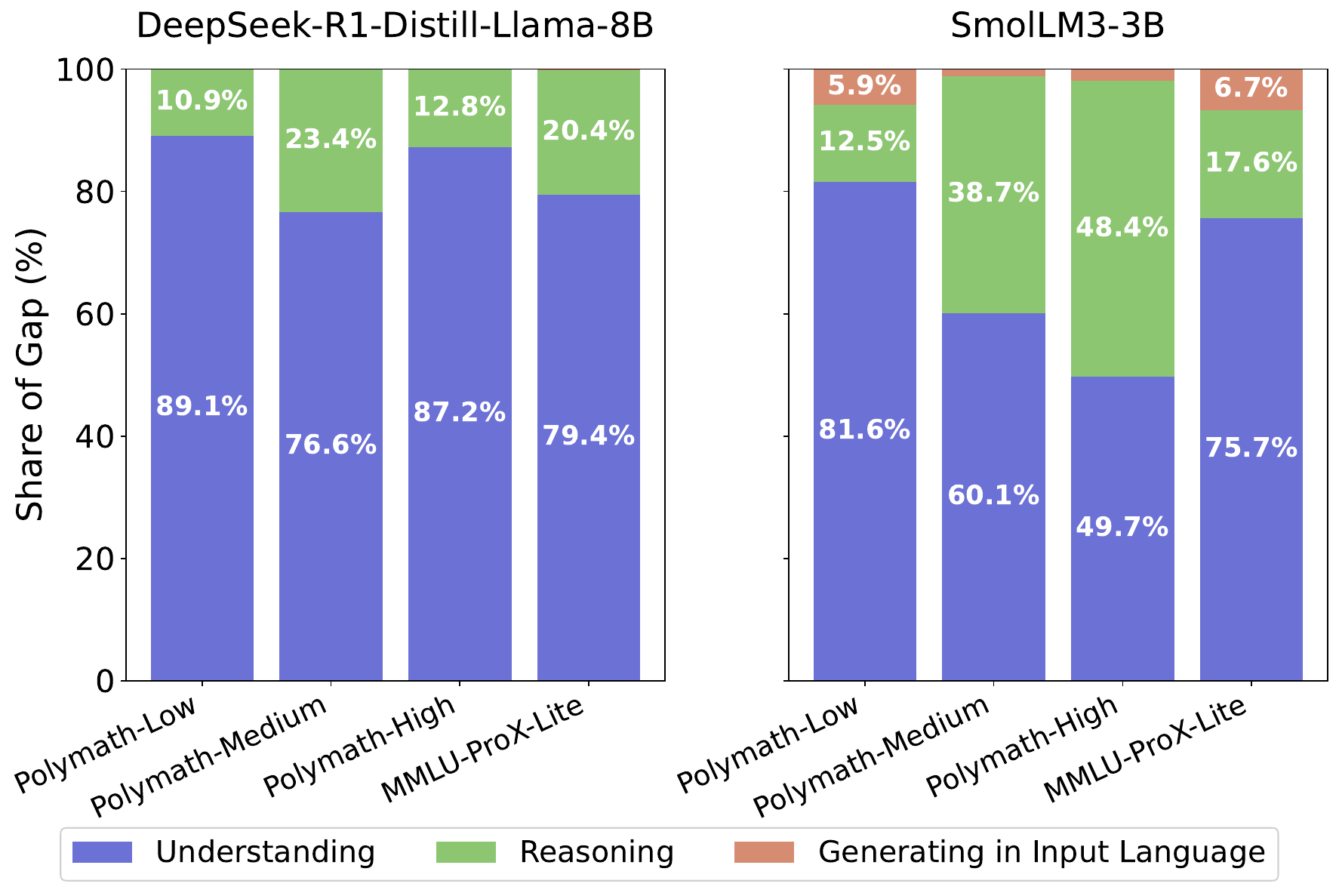}
  \caption{Weighted shares of \textit{Understanding}, \textit{Reasoning}, and \textit{Generation} in the input language to the overall multilingual reasoning gap across additional models.}
  \label{fig:appendix_model_generalization}
\end{figure}
To further examine the generality of our findings, we extend the attribution analysis in Section~\ref{sec:method_gap} to additional open-source reasoning language models: DeepSeek-R1-Distill-Llama-8B~\cite{guo2025deepseek} and SmolLM3-3B~\cite{bakouch2025smollm3}. Figure~\ref{fig:appendix_model_generalization} shows the headroom-weighted average share attributed to the \textit{understanding} stage across datasets. Across models and datasets, understanding failures consistently account for a substantial portion of the multilingual reasoning gap, showing that our findings generalize across diverse open-source RLM families.

\subsection{Robustness to Understanding Prefix Variants}\label{sec:appendix_prefix_variants}
We evaluate the robustness of our stage-wise attribution analysis to the phrasing of the understanding intervention by repeating the analysis with alternative prefixes using Qwen3-4B. The alternative understanding prefix variants are listed below.
\begin{tcolorbox}[
  colback=gray!5,
  colframe=gray!80!black,
  colbacktitle=gray!80!black,
  coltitle=white,
  title={Understanding Prefix Variants},
  boxrule=0.4pt,
  arc=2mm,
  left=2mm, right=2mm, top=1mm, bottom=1mm,
  before upper=\small,
  breakable,
  enhanced
]

\textbf{Prefix-1}\\
\texttt{$\pi(x_{\mathrm{dom}})$ = ``Okay, I understand the question as: '\{$x_{\mathrm{dom}}$\}'. I will solve the problem based on this understanding.''}

\vspace{0.4em}
\textbf{Prefix-2}\\
\texttt{$\pi(x_{\mathrm{dom}})$ = ``Okay, my understanding of the question in English is: '\{$x_{\mathrm{dom}}$\}'. I will proceed using this interpretation.''}

\vspace{0.4em}
\textbf{Prefix-3}\\
\texttt{$\pi(x_{\mathrm{dom}})$ = ``English meaning of the question: '\{$x_{\mathrm{dom}}$\}'. I'll solve the problem based on this understanding.''}

\end{tcolorbox}
The original setting uses the same fixed prefix as in the main text. As shown in Figure~\ref{fig:variant_gap_agg}, understanding failure consistently accounts for the largest share of the multilingual reasoning gap across all prefixes, demonstrating that our conclusion generalizes beyond a specific understanding prefix.

\subsection{Language-specific results on Understanding Failure Detection}\label{sec:appendix_language_specific_detection}
\begin{table*}[!ht]
\centering
\renewcommand{\arraystretch}{0.85}
\resizebox{\textwidth}{!}{
\begin{tabular}{l|ccccc|cc|cc}
\toprule
\multicolumn{10}{c}{\textbf{Polymath-Low}} \\
\cmidrule(lr){1-10}
\textbf{Method} & \textbf{de} & \textbf{es} & \textbf{ar} & \textbf{ja} & \textbf{ko} & \textbf{th} & \textbf{bn} & \textbf{sw} & \textbf{te} \\
\midrule
Random baseline & 16.7 ± 28.9 & 0.0 ± 0.0 & 5.6 ± 9.6 & 3.2 ± 5.5 & 0.0 ± 0.0 & 9.4 ± 9.1 & 11.3 ± 10.5 & 68.0 ± 2.4 & 15.5 ± 5.6 \\
\cmidrule(lr){1-10}
Avg confidence & 32.2 ± 20.7 & 16.5 ± 5.3 & 24.7 ± 8.0 & 32.9 ± 11.2 & 10.4 ± 11.7 & 41.1 ± 5.7 & 43.2 ± 3.8 & 79.5 ± 2.6 & 64.5 ± 2.5 \\
Min Confidence & 25.0 ± 13.8 & 20.4 ± 7.1 & 28.7 ± 11.6 & 25.1 ± 16.3 & 14.3 ± 10.2 & 35.1 ± 7.3 & 36.7 ± 5.3 & 71.4 ± 6.0 & 52.7 ± 2.4 \\
Input NLL & 0.0 ± 0.0 & 0.0 ± 0.0 & 0.0 ± 0.0 & 9.2 ± 3.7 & 0.0 ± 0.0 & 0.0 ± 0.0 & 0.0 ± 0.0 & 69.5 ± 0.6 & 0.0 ± 0.0 \\
\cmidrule(lr){1-10}
Self-reflection & 22.9 ± 20.6 & 50.0 ± 16.7 & 7.4 ± 12.8 & 15.6 ± 14.5 & 0.0 ± 0.0 & 14.7 ± 1.2 & 28.4 ± 14.5 & 45.3 ± 4.4 & 42.7 ± 9.1 \\
LLM-based detector & 0.0 ± 0.0 & 33.3 ± 33.3 & 28.8 ± 18.4 & 5.1 ± 8.9 & 11.1 ± 19.2 & 25.6 ± 11.6 & 25.2 ± 3.8 & 77.5 ± 1.5 & 34.1 ± 12.8 \\
\cmidrule(lr){1-10}
mmBERT detector & 10.3 ± 17.8 & 52.8 ± 21.0 & 36.0 ± 17.4 & \textbf{49.8 ± 12.0} & \textbf{37.7 ± 8.2} & \textbf{51.1 ± 7.4} & \textbf{45.9 ± 0.8} & \textbf{82.7 ± 1.1} & \textbf{67.4 ± 8.9} \\
Prober & \textbf{36.2 ± 19.6} & \textbf{57.8 ± 21.8} & \textbf{40.8 ± 4.9} & 40.1 ± 13.7 & 20.6 ± 18.0 & 51.0 ± 15.2 & 40.6 ± 2.7 & 81.7 ± 1.6 & 61.8 ± 7.5 \\
\midrule[0.4pt]
\vspace{-0.5em} \\[-0.8em]
\midrule[0.4pt]
\multicolumn{10}{c}{\textbf{MMLU-ProX-Lite}} \\
\cmidrule(lr){2-10}
\textbf{Method} & \textbf{de} & \textbf{es} & \textbf{ar} & \textbf{ja} & \textbf{ko} & \textbf{th} & \textbf{bn} & \textbf{sw} & \textbf{te} \\
\midrule
Random baseline & 7.1 ± 7.8 & 2.3 ± 4.0 & 14.1 ± 9.2 & 13.5 ± 8.4 & 5.7 ± 9.9 & 13.9 ± 10.2 & 13.1 ± 6.9 & 29.4 ± 2.4 & 13.2 ± 5.5 \\
\cmidrule(lr){1-10}
Avg confidence & 15.2 ± 7.0 & 30.1 ± 8.1 & 34.6 ± 8.0 & 19.4 ± 1.6 & 30.2 ± 20.6 & \textbf{39.7 ± 14.3} & 23.0 ± 15.2 & 48.5 ± 5.9 & 38.4 ± 7.3 \\
Min Confidence & 25.6 ± 9.2 & 22.0 ± 5.9 & 32.7 ± 5.5 & 28.4 ± 7.0 & 32.3 ± 2.4 & 28.0 ± 4.2 & 32.6 ± 1.8 & 46.6 ± 8.3 & 35.9 ± 3.6 \\
Input NLL & 0.0 ± 0.0 & 0.0 ± 0.0 & 0.0 ± 0.0 & 0.0 ± 0.0 & 2.5 ± 4.3 & 0.0 ± 0.0 & 0.0 ± 0.0 & 42.4 ± 13.0 & 0.0 ± 0.0 \\
\cmidrule(lr){1-10}
Self-reflection & 12.0 ± 12.5 & 18.2 ± 5.9 & 31.9 ± 4.9 & 17.0 ± 9.4 & 32.6 ± 11.8 & 32.8 ± 7.5 & 29.3 ± 9.0 & 32.6 ± 6.4 & 29.7 ± 8.1 \\
LLM-based detector & 4.2 ± 7.2 & 7.4 ± 12.8 & 2.8 ± 4.8 & 6.3 ± 11.0 & 2.8 ± 4.8 & 0.0 ± 0.0 & 6.3 ± 5.5 & 43.3 ± 1.5 & 12.0 ± 5.3 \\
\cmidrule(lr){1-10}
mmBERT detector & 4.4 ± 7.7 & 9.8 ± 9.2 & 16.2 ± 7.6 & 11.5 ± 12.6 & 10.2 ± 10.3 & 16.2 ± 15.0 & 11.0 ± 9.7 & 52.8 ± 0.8 & 21.7 ± 3.0 \\
Prober & \textbf{34.6 ± 10.8} & \textbf{39.2 ± 15.4} & \textbf{46.8 ± 1.1} & \textbf{41.3 ± 10.6} & \textbf{45.1 ± 4.6} & 35.9 ± 8.6 & \textbf{36.6 ± 7.5} & \textbf{56.5 ± 3.4} & \textbf{40.7 ± 7.8} \\
\bottomrule
\end{tabular}
}
\caption{Per-language performance of understanding failure detection methods on Qwen3-4B. F1 values are reported as mean ± stdev, and the best performance for each language is highlighted in bold.}
\label{tab:per_language_f1_qwen3_4b}
\end{table*}

\begin{table*}[!ht]
\centering
\renewcommand{\arraystretch}{0.85}
\resizebox{\textwidth}{!}{
\begin{tabular}{l|ccccc|cc|cc}
\toprule
\multicolumn{10}{c}{\textbf{Polymath-Low}} \\
\cmidrule(lr){1-10}
\textbf{Method} & \textbf{de} & \textbf{es} & \textbf{ar} & \textbf{ja} & \textbf{ko} & \textbf{th} & \textbf{bn} & \textbf{sw} & \textbf{te} \\
\midrule
Random baseline & 0.0 ± 0.0 & 7.4 ± 12.8 & 0.0 ± 0.0 & 14.8 ± 25.7 & 4.4 ± 7.7 & 8.3 ± 14.4 & 10.3 ± 9.0 & 16.9 ± 1.0 & 5.8 ± 10.0 \\
\cmidrule(lr){1-10}
Avg confidence & 14.6 ± 9.8 & 20.1 ± 1.9 & 19.0 ± 8.8 & 14.4 ± 1.4 & 16.2 ± 4.5 & 19.4 ± 6.2 & 23.5 ± 9.3 & \textbf{42.9 ± 5.1} & 47.8 ± 3.8 \\
Min Confidence & 6.7 ± 11.5 & 31.8 ± 8.0 & 15.8 ± 15.4 & 15.9 ± 3.0 & 21.7 ± 18.8 & 6.4 ± 5.7 & 30.0 ± 12.6 & 27.6 ± 8.8 & 44.0 ± 0.5 \\
Input NLL & 4.5 ± 4.3 & 10.5 ± 2.5 & 16.7 ± 3.3 & 0.0 ± 0.0 & 11.1 ± 19.2 & 0.0 ± 0.0 & 0.0 ± 0.0 & 28.3 ± 3.0 & 0.0 ± 0.0 \\
\cmidrule(lr){1-10}
Self-reflection & 0.0 ± 0.0 & 0.0 ± 0.0 & 0.0 ± 0.0 & 0.0 ± 0.0 & 0.0 ± 0.0 & 0.0 ± 0.0 & 15.7 ± 13.7 & 0.0 ± 0.0 & 6.1 ± 10.5 \\
LLM-based detector & 0.0 ± 0.0 & 9.5 ± 16.5 & \textbf{29.1 ± 9.6} & 0.0 ± 0.0 & 0.0 ± 0.0 & 0.0 ± 0.0 & 6.7 ± 11.5 & 12.7 ± 9.0 & 0.0 ± 0.0 \\
\cmidrule(lr){1-10}
mmBERT detector & 16.7 ± 28.9 & \textbf{54.5 ± 31.0} & 7.8 ± 13.6 & \textbf{29.3 ± 28.6} & 16.3 ± 18.5 & \textbf{26.5 ± 28.8} & \textbf{43.5 ± 6.3} & 34.6 ± 13.3 & 38.8 ± 13.1 \\
Prober & \textbf{16.9 ± 18.3} & 46.9 ± 13.6 & 19.5 ± 14.0 & 24.2 ± 4.8 & \textbf{26.7 ± 17.7} & 24.9 ± 7.5 & 37.0 ± 13.0 & 35.4 ± 5.4 & \textbf{52.4 ± 15.7} \\
\midrule[0.4pt]
\vspace{-0.5em} \\[-0.8em]
\midrule[0.4pt]
\multicolumn{10}{c}{\textbf{MMLU-Prox-Lite}} \\
\cmidrule(lr){1-10}
\textbf{Method} & \textbf{de} & \textbf{es} & \textbf{ar} & \textbf{ja} & \textbf{ko} & \textbf{th} & \textbf{bn} & \textbf{sw} & \textbf{te} \\
\midrule
Random baseline & 3.9 ± 6.8 & 3.5 ± 3.0 & 6.9 ± 7.6 & 1.9 ± 3.2 & 7.7 ± 1.7 & 8.5 ± 7.8 & 3.4 ± 5.9 & 17.0 ± 2.8 & 4.7 ± 4.1 \\
\cmidrule(lr){1-10}
Avg confidence & 9.7 ± 1.3 & 13.7 ± 2.6 & 15.9 ± 6.1 & 11.9 ± 2.2 & 15.4 ± 3.0 & 14.6 ± 1.7 & 13.0 ± 2.4 & 33.5 ± 4.0 & 10.8 ± 2.9 \\
Min Confidence & 13.2 ± 3.4 & 18.3 ± 3.8 & 19.8 ± 8.6 & 17.0 ± 1.6 & 18.7 ± 3.5 & 20.3 ± 3.2 & 22.0 ± 4.9 & 36.8 ± 6.3 & 17.9 ± 3.7 \\
Input NLL & 4.1 ± 7.1 & 6.2 ± 5.5 & 17.2 ± 10.3 & 0.0 ± 0.0 & 10.6 ± 10.1 & 17.8 ± 1.3 & 10.3 ± 4.8 & 30.2 ± 2.7 & 8.6 ± 7.7 \\
\cmidrule(lr){1-10}
Self-reflection & 6.1 ± 10.5 & 15.4 ± 8.4 & 10.8 ± 9.6 & 18.4 ± 16.0 & 14.0 ± 4.8 & 8.4 ± 7.3 & 10.5 ± 9.7 & 11.0 ± 7.9 & \textbf{28.7 ± 11.5} \\
LLM-based detector & 0.0 ± 0.0 & 0.0 ± 0.0 & 8.5 ± 9.1 & 0.0 ± 0.0 & 0.0 ± 0.0 & 3.9 ± 6.8 & 5.6 ± 9.6 & 12.7 ± 2.1 & 0.0 ± 0.0 \\
\cmidrule(lr){1-10}
mmBERT detector & 10.6 ± 9.7 & \textbf{28.4 ± 2.3} & \textbf{23.6 ± 25.1} & 22.7 ± 6.0 & \textbf{24.3 ± 12.3} & 21.8 ± 10.3 & 23.4 ± 10.3 & 36.1 ± 2.8 & 11.8 ± 10.2 \\
\cmidrule(lr){1-10}
Prober & \textbf{15.6 ± 14.5} & 21.7 ± 6.1 & 22.4 ± 2.8 & \textbf{23.8 ± 3.9} & 20.7 ± 6.3 & \textbf{29.3 ± 3.2} & \textbf{32.9 ± 9.9} & \textbf{36.9 ± 11.3} & 9.5 ± 9.5 \\
\bottomrule
\end{tabular}
}
\caption{Per-language performance of understanding failure detection methods on gpt-oss-20b. F1 values are reported as mean ± stdev, and the best performance for each language is highlighted in bold.}
\label{tab:per_language_f1_oss_20b}
\end{table*}

\begin{table*}[!ht]
\centering
\renewcommand{\arraystretch}{0.85}
\tiny
\begin{tabular}{l|ccc}
\toprule
\multicolumn{4}{c}{\textbf{Polymath-Low}} \\
\cmidrule(lr){1-4}
\textbf{Method} & \textbf{fr} & \textbf{mr} & \textbf{wo} \\
\midrule
\cmidrule(lr){1-4}
Avg confidence & 17.9 ± 15.6 & 0.0 ± 0.0 & 33.5 ± 19.2  \\
\cmidrule(lr){1-4}
LLM-based detector & 0.0 ± 0.0 & 15.3 ± 6.0 & 40.4 ± 3.6 \\
\cmidrule(lr){1-4}
mmBERT detector & \textbf{26.2 ± 25.0} & \textbf{38.7 ± 16.3} & \textbf{86.1 ± 5.3} \\
Prober & 9.5 ± 16.5 & 25.0 ± 25.0 & 76.8 ± 13.2 \\
\midrule[0.4pt]
\vspace{-0.5em} \\[-0.8em]
\midrule[0.4pt]
\multicolumn{4}{c}{\textbf{MMLU-ProX-Lite}} \\
\cmidrule(lr){1-4}
\textbf{Method} & \textbf{fr} & \textbf{mr} & \textbf{wo} \\
\midrule
\cmidrule(lr){1-4}
Avg confidence & 31.3 ± 9.6 & \textbf{37.3 ± 7.8} & 51.9 ± 5.0  \\
\cmidrule(lr){1-4}
LLM-based detector & 4.7 ± 8.2 & 4.9 ± 4.2 & 13.5 ± 5.0  \\
\cmidrule(lr){1-4}
mmBERT detector & 4.8 ± 8.2 & 15.6 ± 6.7 & 63.6 ± 8.8 \\
Prober & \textbf{44.2 ± 16.0} & 35.2 ± 11.1 & \textbf{70.5 ± 5.8} \\
\bottomrule
\end{tabular}
\caption{Per-language performance of understanding failure detection methods on \textit{unseen} languages (fr, mr, wo) on Qwen3-4B. F1 values are reported as mean ± stdev, and the best performance for each language is highlighted in bold.}
\label{tab:per_language_f1_unseen}
\end{table*}

Table~\ref{tab:per_language_f1_qwen3_4b} and Table~\ref{tab:per_language_f1_oss_20b} report the F1 scores of each detection method across languages for Qwen3-4B and gpt-oss-20b, respectively.
Overall, supervised approaches such as the mmBERT detector and the Prober generally outperform other baselines, while the Avg-Confidence method surpasses them on a few individual languages.
Table~\ref{tab:per_language_f1_unseen} presents results on \textit{unseen} languages for Qwen3-4B, where similar trends are observed.\footnote{For the Polymath dataset, we use the translated dataset generated by gpt-4.1 using the prompt described in Section~\ref{sec:appendix_selective_translation}.}
These per-language breakdowns indicate that supervised approaches are not biased toward specific languages but instead exhibit robust performance across diverse languages.
\subsection{Selective Translation Results on gpt-oss-20b}\label{sec:appendix_st_oss_20b}
\begin{table*}[!t]
\centering
\resizebox{\textwidth}{!}{
\begin{tabular}{l|l|cccccc|cc|cc|c}
\toprule
& \textbf{Method} & en & de & es & ar & ja & ko & th & bn & sw & te & \textbf{Avg. (Translator usage)} \\
\midrule
\multirow{3}{*}{\textbf{Polymath-Low}} 
& Base                  & 96.0 & 85.6 & 89.9 & 90.9 & 86.1 & 91.2 & 90.4 & 90.1 & 78.7 & 81.1 & \textbf{88.0 (0.0\%)} \\
& Selective translation & 
\makecell{96.8 \\ (6.1\%)} &
\makecell{85.9 \\ (8.5\%)} &
\makecell{91.5 \\ (11.2\%)} &
\makecell{90.4 \\ (16.3\%)} &
\makecell{85.9 \\ (9.6\%)} &
\makecell{90.9 \\ (12.8\%)} &
\makecell{90.9 \\ (12.3\%)} &
\makecell{88.5 \\ (21.1\%)} &
\makecell{84.8 \\ (50.4\%)} &
\makecell{83.5 \\ (18.7\%)} &
\textbf{88.9 (16.7\%)}  \\
\cmidrule(lr){2-13}
& Full translation      & 93.9 & 84.5 & 88.3 & 90.4 & 86.4 & 88.5 & 87.7 & 86.9 & 86.4 & 85.6 & \textbf{87.9 (100.0\%)} \\
\midrule
\multirow{3}{*}{\textbf{MMLU-ProX-Lite}} 
& Base                  & 77.4 & 77.8 & 77.3 & 74.1 & 77.4 & 73.9 & 74.6 & 75.6 & 66.9 & 77.0 & \textbf{75.2 (0.0\%)} \\
& Selective translation &
\makecell{76.7 \\ (3.6\%)} &
\makecell{77.0 \\ (5.4\%)} &
\makecell{77.4 \\ (4.7\%)} &
\makecell{74.6 \\ (7.1\%)} &
\makecell{76.0 \\ (7.9\%)} &
\makecell{74.1 \\ (8.2\%)} &
\makecell{75.0 \\ (7.3\%)} &
\makecell{75.5 \\ (6.4\%)} &
\makecell{68.9 \\ (17.8\%)} &
\makecell{76.1 \\ (3.9\%)} &
\textbf{75.1 (7.2\%)} \\
\cmidrule(lr){2-13}
& Full translation      & 78.2 & 77.7 & 78.6 & 77.8 & 76.0 & 76.3 & 75.7 & 76.1 & 74.4 & 76.9 & \textbf{76.8 (100.0\%)} \\
\bottomrule
\end{tabular}
}
\caption{Performance of translation strategies with gpt-oss-20b.
\textbf{Avg. (Translator usage)} reports average accuracy across languages, with the overall translator usage shown in parentheses.
For \textbf{Selective translation}, per-language translator usage (\%) is shown below each accuracy score.}
\label{tab:selective_translation_results_oss_20b}
\end{table*}

Table~\ref{tab:selective_translation_results_oss_20b} presents the Selective Translation results on gpt-oss-20b. In Polymath-Low, Selective Translation is highly effective, outperforming both the Base model (88.0\%) and Full Translation (87.9\%) while requiring translation for only 16.7\% of the inputs. While performance improves for specific languages such as Swahili, the overall impact on average accuracy is minimal on MMLU-ProX-Lite. We attribute this to the relatively narrow initial gap between the base ceiling language (English) and other languages. Consequently, the detector triggers translation for only 7.2\% of the inputs, resulting in a minimal impact on the average score. Nevertheless, these results demonstrate that Selective Translation acts as an adaptive strategy, invoking translation only when necessary and generalizing to larger model scales.

\subsection{Selective translation with Early Detection}\label{sec:appendix_st_early_detection}
\begin{table*}[!t]
\centering
\resizebox{\textwidth}{!}{
\begin{tabular}{l|l|cccccc|cc|cc|c}
\toprule
& \textbf{Method} & en & de & es & ar & ja & ko & th & bn & sw & te & \textbf{Avg. (Translator usage)} \\
\midrule
\multirow{8}{*}{\textbf{Polymath-Low}}
& Base                         & 96.5 & 88.0 & 93.9 & 89.6 & 85.3 & 90.7 & 85.1 & 83.2 & 29.3 & 69.9 & \textbf{81.1 (0.0\%)} \\
\cmidrule(lr){2-13}
& Selective translation (512)  &
\makecell{96.5 \\ (4.0\%)} &
\makecell{88.3 \\ (4.5\%)} &
\makecell{93.6 \\ (2.7\%)} &
\makecell{89.9 \\ (3.7\%)} &
\makecell{85.6 \\ (11.5\%)} &
\makecell{91.2 \\ (3.2\%)} &
\makecell{85.6 \\ (7.7\%)} &
\makecell{84.3 \\ (21.9\%)} &
\makecell{76.5 \\ (77.9\%)} &
\makecell{73.3 \\ (28.3\%)} &
\textbf{86.5 (19.6\%)} \\
& Selective translation (1024) &
\makecell{96.3 \\ (5.6\%)} &
\makecell{88.5 \\ (2.7\%)} &
\makecell{94.1 \\ (3.7\%)} &
\makecell{89.9 \\ (2.7\%)} &
\makecell{85.6 \\ (5.9\%)} &
\makecell{91.2 \\ (2.1\%)} &
\makecell{86.4 \\ (6.9\%)} &
\makecell{84.0 \\ (20.0\%)} &
\makecell{78.7 \\ (78.7\%)} &
\makecell{74.7 \\ (26.7\%)} &
\textbf{86.9 (15.5\%)} \\
& Selective translation (2048) &
\makecell{96.5 \\ (4.3\%)} &
\makecell{88.3 \\ (2.4\%)} &
\makecell{94.7 \\ (1.6\%)} &
\makecell{90.1 \\ (2.9\%)} &
\makecell{86.1 \\ (5.3\%)} &
\makecell{90.9 \\ (2.9\%)} &
\makecell{86.7 \\ (7.2\%)} &
\makecell{85.9 \\ (17.6\%)} &
\makecell{79.5 \\ (85.9\%)} &
\makecell{77.1 \\ (34.7\%)} &
\textbf{87.6 (25.6\%)} \\
& Selective translation (4096) &
\makecell{96.5 \\ (2.1\%)} &
\makecell{88.3 \\ (0.8\%)} &
\makecell{94.7 \\ (1.1\%)} &
\makecell{90.1 \\ (1.3\%)} &
\makecell{86.4 \\ (4.3\%)} &
\makecell{90.9 \\ (0.8\%)} &
\makecell{87.2 \\ (7.5\%)} &
\makecell{86.9 \\ (12.3\%)} &
\makecell{76.3 \\ (69.1\%)} &
\makecell{76.0 \\ (17.9\%)} &
\textbf{87.3 (21.5\%)} \\
& Selective translation (8192) &
\makecell{96.3 \\ (5.3\%)} &
\makecell{88.8 \\ (2.7\%)} &
\makecell{94.9 \\ (2.7\%)} &
\makecell{90.4 \\ (4.0\%)} &
\makecell{85.9 \\ (6.1\%)} &
\makecell{91.5 \\ (5.1\%)} &
\makecell{87.5 \\ (8.5\%)} &
\makecell{86.1 \\ (13.1\%)} &
\makecell{78.1 \\ (73.9\%)} &
\makecell{76.3 \\ (21.1\%)} &
\textbf{87.6 (23.0\%)} \\
& Selective translation (full) &
\makecell{96.3 \\ (1.6\%)} &
\makecell{88.3 \\ (3.7\%)} &
\makecell{94.4 \\ (3.7\%)} &
\makecell{90.4 \\ (4.8\%)} &
\makecell{86.1 \\ (13.1\%)} &
\makecell{91.5 \\ (5.9\%)} &
\makecell{88.3 \\ (9.6\%)} &
\makecell{86.7 \\ (26.1\%)} &
\makecell{81.3 \\ (86.4\%)} &
\makecell{77.1 \\ (37.9\%)} &
\textbf{88.0 (19.3\%)} \\
\cmidrule(lr){2-13}
& Full translation             & 96.0 & 88.3 & 93.3 & 90.9 & 87.5 & 92.5 & 89.6 & 90.4 & 85.3 & 80.5 & \textbf{89.4 (100.0\%)} \\

\bottomrule
\end{tabular}
}
\caption{
Selective Translation with Early-detection on \textbf{Polymath-Low} with Qwen3-4B.
For each cutoff length (512, 1024, …, full), we allow the detector to read signals from only the first $N$ tokens of its own reasoning trace before deciding whether to translate.
Each cell shows accuracy (top) and per-language translator usage (\%) below.
\textbf{Avg. (Translator usage)} reports overall average accuracy across languages, with overall translator usage in parentheses.
}
\label{tab:early_detection_selective_translation}
\end{table*}

Table~\ref{tab:early_detection_selective_translation} presents the results of applying early detection to Selective Translation on Polymath-low on Qwen3-4B. 
Here, the detector is implemented using the Prober model, which takes the last \textit{hidden state} (from the residual stream of the last layer of the model) of the reasoning trace  as input and observes only a prefix of the trace—up to $N$ tokens—before deciding whether translation is needed. The full reasoning trace extends up to 32,768 tokens. Remarkably, using only 2,048 tokens of the reasoning trace yields performance nearly identical to the full-trace setting (\textbf{87.6 vs.\ 88.0} average accuracy). 
This trend persists across longer cutoffs, indicating that signals of misunderstanding are already detectable within the early stages of reasoning.

These results demonstrate that \textbf{Selective Translation remains highly effective even under early detection}, suggesting that it does not need to wait until the model completes its entire reasoning trace. 
Instead, the decision can be accurately made using only the initial hidden states from the beginning, making the method substantially more efficient while maintaining comparable accuracy to the full-trace version.

\subsection{Latency and Detector Overhead of Selective Translation}
\label{sec:appendix_efficiency}

We analyze the efficiency of Selective Translation (ST), focusing on the overhead introduced by the understanding failure detector. Specifically, we measure the latency of the prober used in our method. On an RTX 6000 Ada GPU, the prober adds less than 0.01 seconds per question on average across both MMLU-ProX-Lite and Polymath, showing that \textbf{the detector itself introduces negligible overhead.}

We further report the average per-question latency of Base and Selective Translation in Table~\ref{tab:st_efficiency}, using Qwen3-4B as the reasoning language model (RLM) on an RTX 6000 Ada GPU. While ST incurs additional latency, this increase is primarily due to re-running inference on queries where understanding failures are detected. Importantly, this overhead is not fixed and is expected to decrease as the RLM becomes more multilingual and triggers translation less frequently.
\begin{table}[t]
\centering
\small
\begin{tabular}{lccc}
\toprule
\multicolumn{4}{c}{\textbf{Polymath-Low}} \\
\midrule
Method & Latency (s) & Input tokens & Output tokens \\
\midrule
Base & 2.45 & 168.6 & 2588.1 \\
ST & 3.30 & 224.4 & 3184.5 \\
\midrule
\multicolumn{4}{c}{\textbf{MMLU-ProX-Lite}} \\
\midrule
Method & Latency (s) & Input tokens & Output tokens \\
\midrule
Base & 9.27 & 403.3 & 5489.1 \\
ST & 13.1 & 558.5 & 7187.5 \\
\bottomrule
\end{tabular}
\caption{End-to-end latency and token usage for Base and Selective Translation (ST) using Qwen3-4B.}
\label{tab:st_efficiency}
\end{table}
\section{Examples of Understanding Failures in Reasoning Language Models}\label{sec:appendix_uf_case_study}
In this section, we present examples showing how understanding failures manifest in the reasoning traces of reasoning language models. At the beginning of reasoning, the model attempts to reinterpret the given question into the dominant language of its reasoning trace (in this case, English). This process may occur implicitly or be explicitly verbalized (e.g., “It’s in Swahili. Let’s translate.”). During this \textit{understanding} stage, \textbf{models sometimes exhibit uncertainty—repeating translation multiple times, or explicitly expressing confusion} (e.g., “This is confusing.”). We highlight in red the parts of the reasoning traces where such uncertainty is particularly evident, and present each example along with a brief observation. We collected several representative cases across different models and tasks to illustrate these behaviors. The examples are shown below. 

\begin{tcolorbox}[
  colback=gray!5,
  colframe=gray!80!black,
  colbacktitle=gray!80!black,
  coltitle=white,
  title={[Korean / Polymath-Low] Qwen3-1.7B — Understanding Failure},
  boxrule=0.4pt,
  arc=2mm,
  left=2mm, right=2mm, top=1mm, bottom=1mm,
  before upper=\small,
  breakable,
  enhanced
]
\textbf{Question:}  
\begin{CJK*}{UTF8}{mj}
양초는 타는 동안 매 시간마다 2센티미터씩 녹습니다. 양초가 오후 1시부터 오후 5시까지 타면 몇 센티미터가 짧아질까요? \\
\end{CJK*} \\
\noindent\textit{(Translated: \underline{A candle melts by 2 centimeters every hour} that it burns. How many centimeters shorter will a candle be after burning from 1:00 PM to 5:00 PM?)}

\vspace{2mm}
\textbf{Reasoning Trace (excerpt):}  
<think>Okay, let's see. The problem is about a matchstick (yangchu) that burns for a certain period, and \underline{each minute it gets shorter by 2 centimeters.} We need to find out how much it shortens from 1 PM to 5 PM. Hmm. First, I need to figure out how many minutes are there between 1 PM and 5 PM. Let me think. From 1 PM to 5 PM is 4 hours. Since there are 60 minutes in an hour, 4 hours would be 4 * 60 = 240 minutes. Wait, but is that right? Let me check again. 1 PM to 2 PM is 60 minutes, 2 PM to 3 PM is another 60, 3 PM to 4 PM, and 4 PM to 5 PM. So yeah, that's 4 hours, which is 240 minutes. So the total time is 240 minutes. Now, the problem says that each minute, the yangchu shortens by 2 centimeters. So if it's burning for 240 minutes, then the total shortening would be 2 centimeters per minute multiplied by 240 minutes. Let me calculate that: 2 * 240 = 480 centimeters. \textcolor{red}{Wait, but that seems like a lot. Is that possible? Wait, but maybe I need to check if the question is asking for the total shortening or something else. The question says, 
\begin{CJK*}{UTF8}{mj}
"몇 센티미터가 짧아질까요?"
\end{CJK*}
which translates to "how many centimeters will be shortened?"} So it's asking for the total shortening. So if each minute it shortens 2 cm, over 240 minutes, it's 2*240=480 cm. But that seems like a very large number. \textcolor{red}{Is there a possibility that I'm misunderstanding the question?...}
\vspace{2mm} 

\textbf{Observation:}  
The model misinterprets the unit “per hour” as “per minute,” repeatedly expresses uncertainty, and questions its own understanding, indicating an understanding failure.
\end{tcolorbox}

\begin{tcolorbox}[
  colback=gray!5,
  colframe=gray!80!black,
  colbacktitle=gray!80!black,
  coltitle=white,
  title={[Swahili / Polymath-Medium] Qwen3-4B — Understanding Failure},
  breakable,
  boxrule=0.4pt,
  arc=2mm,
  left=2mm, right=2mm, top=1mm, bottom=1mm,
  enhanced,
  before upper=\small
]
\textbf{Question:}  
Uwakilishi wa desimali wa $m/n,$ ambapo $m$ na $n$ ni nambari chanya ambazo hazina kipengele cha pamoja isipokuwa 1 na $m < n,$ una tarakimu 2, 5 na 1 zinazofuatana katika mpangilio huo. Tafuta thamani ndogo kabisa ya $n$ ambayo hili linawezekana. \\
\\
\noindent\textit{(Translated: The decimal representation of $m/n,$ where $m$ and $n$ are relatively prime positive integers and $m < n,$ contains the digits 2, 5, and 1 consecutively, and in that order. Find the smallest value of $n$ for which this is possible.)}

\vspace{2mm}
\textbf{Reasoning Trace (excerpt):}  
<think> Okay, so I need to find the smallest possible value of n such that the decimal expansion of m/n has a repeating cycle of length 2, 5, or 1. Wait, the problem says "tarakimu 2, 5 na 1 zinazofuatana katika mpangilio huo." Hmm, maybe that means the decimal has a period of 2, 5, or 1? Or maybe the decimal expansion has periods of 2, 5, and 1? But that doesn't make much sense. Wait, maybe it's saying that the decimal expansion has a period of 2, 5, or 1? But periods are usually the length of the repeating part. So if the decimal has a period of 1, 2, or 5, then n must be such that the decimal expansion of m/n has a period that is one of those numbers. \textcolor{red}{But the question says "tarakimu 2, 5 na 1 zinazofuatana katika mpangilio huo." Maybe it's saying that the decimal expansion has periods 2, 5, and 1? But that seems conflicting. Wait, maybe the original problem is in Swahili, and the translation is a bit off. Let me try to parse it again. "Uwakilishi wa desimali wa m/n, ambapo m na n ni nambari chanya ambazo hazina kipengele cha pamoja isipokuwa 1 na m < n, una tarakimu 2, 5 na 1 zinazofuatana katika mpangilio huo. Tafuta thamani ndogo kabisa ya n ambayo hili linawezekana."} So, the decimal expansion of m/n, where m and n are coprime integers with m < n, has periods 2, 5, and 1 in the given context. \textcolor{red}{Wait, maybe "tarakimu" is "periods" and "zinazofuatana katika mpangilio huo" is "occur in this expansion"...} 

\vspace{2mm}
\textbf{Observation:}  
The model repeatedly attempts to reinterpret the Swahili condition, and incorrectly treats “the digits 2, 5, and 1 consecutively” as a statement about repeating period length (e.g., “a period of 2, 5, or 1”), indicating an understanding failure and sustained uncertainty about the problem statement.
\end{tcolorbox}

\begin{tcolorbox}[
  colback=gray!5,
  colframe=gray!80!black,
  colbacktitle=gray!80!black,
  coltitle=white,
  title={[Swahili / MMLU-ProX-Lite] gpt-oss-20b — Understanding Failure},
  breakable,
  boxrule=0.4pt,
  arc=2mm,
  left=2mm, right=2mm, top=1mm, bottom=1mm,
  enhanced,
  before upper=\small
]
\textbf{Question:}  
Katika nafasi $x_0$ = kitengo 1 kwenye skrini, kazi ya mawimbi ya kwanta kwa boriti ya elektroni ina thamani +1 kitengo, na katika kipindi cha $\Delta$x na karibu na $x_0$ = 1 kuna miale 100 kwa dakika. Je, ni nini nguvu ya miale katika $x_0$ = 2, 3, na 4 vitengo ambapo i ina thamani +4, +2, na -2 vitengo, mtawalia? \\
\\
\noindent\textit{(Translated: At the position $x_0$ = 1 unit on a screen, the wave function for an electron beam has the value +1 unit, and in an interval $\Delta$x and around $x_0$ = 1 there are observed 100 light flashes per minute. What is the intensity of flashes at $x_0$ = 2, 3, and 4 units where it has the values +4, +2, and -2 units, respectively?)}

\vspace{2mm}
\textbf{Reasoning Trace (excerpt):}  
We need to parse the question. It's in Swahili. Let's translate: "Yafuatayo ni maswali ya chaguo-nyingi (yenye majibu) kuhusu fizikia. Fikiria hatua kwa hatua kisha malizia jibu lako kwa 'Jibu ni (X)' ambapo X ni herufi ya chaguo sahihi. Swali: Katika nafasi $x_0$ = kitengo 1 kwenye skrini, kazi ya mawimbi ya kwanta kwa boriti ya elektroni ina thamani +1 kitengo, na katika kipindi cha $\Delta$x na karibu na $x_0$ = 1 kuna miale 100 kwa dakika. Je, ni nini nguvu ya miale katika $x_0$ = 2, 3, na 4 vitengo ambapo i ina thamani +4, +2, na -2 vitengo, mtawalia? \textcolor{red}{\textbf{This is confusing}: "kazi ya mawimbi ya kwanta kwa boriti ya elektroni ina thamani +1 kitengo" maybe refers to potential energy? "miale 100 kwa dakika" maybe "force"? Actually "miale" maybe "current"? Wait "miale" maybe "ampere"? Might be "current" in physics? Let's parse: "kazi ya mawimbi ya kwanta" maybe "work done by the wave" ??? "boriti ya elektroni" is "electron's potential"?} Hmm. We might need to interpret the physics concept...
\vspace{2mm}

\textbf{Observation:}  
The model struggles to ground physical concepts expressed in Swahili, repeatedly expressing uncertainty about key terms (e.g., miale, kazi ya mawimbi ya kwanta) and speculating multiple conflicting meanings (“current,” “energy,” “intensity”). It even explicitly verbalizes its confusion (e.g., “This is confusing”), illustrating a clear instance of understanding failure.
\end{tcolorbox}

\section{Further Details on Multilingual Reasoning Gap Analysis}
\subsection{Dataset Details}\label{appendix:dataset_details}
\paragraph{Polymath} Polymath~\cite{wang2025polymath} is a multilingual mathematical reasoning benchmark covering 18 languages and four difficulty levels—low, medium, high, and top. The four levels are organized as follows:
\begin{itemize}
    \item Low: Covers K–12 mathematics such as math word problems. Includes 125 multilingual samples from MGSM~\cite{shilanguage} and four additional languages from P-MMeval~\cite{zhang2024p}.
    \item Medium: Consists of exam-style problems from college mathematics, China’s Gaokao, and postgraduate entrance exams, along with entry-level competition questions from AMC and CNMO provincial contests, all collected from official sources.
    \item High: Focuses on mid- to high-difficulty competition problems, primarily from AIME and CNMO.
    \item Top: Comprises Top Olympiad problems from IMO, IMO Shortlist, and national contests such as USAMO, CMO, and Putnam, as well as 25 research-level problems from HLE~\cite{phan2025humanity}.
\end{itemize}
After collecting original problems from diverse public sources, each problem was translated into 18 languages using GPT-4o~\cite{hurst2024gpt} and subsequently refined by language experts with mathematical expertise, ensuring the precise translation of mathematical terminology. This process resulted in high-quality, consistent problems across all languages, with \textbf{125 problems per language} at each difficulty level.

In this study, we focus on the low, medium, and high levels because (1) these three levels sufficiently capture performance trends across different reasoning difficulties, and (2) the top-level problems, which involve Olympiad and frontier mathematics, are too challenging for the evaluated reasoning models, leading to uniformly low accuracy that prevents statistically meaningful multilingual gap analysis.
We use the official Polymath evaluation template for all experiments.
The prompt template is shown below:
\begin{tcolorbox}[
  colback=gray!5,
  colframe=gray!80!black,
  colbacktitle=gray!80!black,
  coltitle=white,
  title=Polymath Evaluation Prompt Template
]
\textbf{Prompt format:} Question in language \texttt{L} + Instruction prompt in language \texttt{L}
\label{box:polymath-prompt}
\end{tcolorbox}
The instruction prompt varies by language.
For English, the instruction is
\texttt{“Note: Please put the final answer in the \textbackslash boxed\{\}.”}.
For other languages, the instruction prompt is a translation of the English version, and all language-specific instructions are available in the official Polymath repository \footnote{\url{https://github.com/QwenLM/PolyMath/}}.

\paragraph{MMLU-ProX-Lite}
MMLU-ProX~\cite{xuanmmluprox} is a multilingual benchmark extending the reasoning-focused English benchmark MMLU-Pro~\cite{wang2024mmlu} to 29 typologically diverse languages, enabling cross-linguistic comparison of reasoning ability with fully parallel question sets.
Each language version of the full benchmark contains 11{,}829 multiple-choice questions covering 57 subjects, with each question providing up to 10 answer options. A Lite version (MMLU-ProX-Lite) was released alongside the full benchmark, consisting of 658 uniformly sampled questions per language while maintaining balanced subject coverage and reasoning difficulty. In this study, we adopt MMLU-ProX-Lite for efficient experiments and restrict evaluation to \textbf{five STEM-related subjects}—\texttt{math}, \texttt{physics}, \texttt{chemistry}, \texttt{computer science}, and \texttt{engineering}—to focus on reasoning-intensive domains rather than factual recall. After applying this filtering, the resulting subset contains \textbf{257 samples per language.}

We adopt the official MMLU-ProX-Lite evaluation template implemented in the \texttt{lm-evaluation-harness}~\cite{eval-harness} for all experiments.
Each prompt consists of (i) a short instruction introducing the subject, (ii) the question body, and (iii) a list of options labeled from A to J.
The model is instructed to reason step by step and to conclude with a language-specific closing phrase indicating the selected option; for English, this takes the form \texttt{"the answer is (X)"} (where \texttt{X} denotes the correct letter), while other languages use the corresponding translation. All instructional phrases and subject names are likewise translated into the target language to ensure semantic consistency and fairness in multilingual reasoning evaluation.
The English prompt format is shown below:

\begin{tcolorbox}[
colback=gray!5,
colframe=gray!80!black,
colbacktitle=gray!80!black,
coltitle=white,
title=MMLU-ProX-Lite Evaluation Prompt Template,
listing only,           
listing options={basicstyle=\ttfamily\small,breaklines=true,columns=fullflexible}
]
\textbf{Prompt format (English example):}
The following are multiple choice questions (with answers) about {{subject}}.
Think step by step and then finish your answer with "the answer is (X)"
where X is the correct letter choice.

Question: \{question\} \\
Options: \\
(A) \{option 1\} \\
(B) \{option 2\} \\
... \\
(J) \{option 10\} \\

Answer: Let's think step by step.
\end{tcolorbox}

\subsection{Evaluation Details}\label{appendix:eval_details}
\paragraph{Answer Extraction.}
We first perform rule-based answer extraction based on regular expressions. 
For Polymath, the final answer is identified from the expression enclosed in \texttt{\textbackslash boxed\{\}}, 
while for MMLU-ProX-Lite, it is extracted from the language-specific closing phrase 
(e.g., \texttt{r"answer is \(?([A--J])\)?"}) by selecting the last matching string. 
However, since models do not always follow the specified output format and the final answers often appear in diverse forms, we use Qwen2.5-7B-Instruct to parse and extract the final answer when regular-expression matching fails. The prompts used for answer extraction are shown below.

\begin{tcolorbox}[
  colback=gray!5,
  colframe=gray!80!black,
  colbacktitle=gray!80!black,
  coltitle=white,
  title=Polymath Answer Extraction Prompt,
  enhanced,
  breakable,
  width=\linewidth,
  boxsep=4pt
]
You are an answer extractor.\\
Inputs:\\
- Question: \{question\}\\
- Reasoning trace: \{reasoning\_trace\}\\
Task:\\
1) Read the Question and determine the expected final answer type.\\
- Possible types include: Numeric scalar, Comparison/Ordering among variables, Set/List, Interval/Inequality, Coordinate/Tuple, Algebraic expression, or Multiple-choice letter.\\
- Decide the most appropriate type for THIS Question.\\
2) Carefully scan the Reasoning trace and identify the final/conclusive answer consistent with the expected type.\\
- Prefer the final/most conclusive statement (e.g., “Therefore…”, “Thus…”, “Final answer…”, or the last decisive equation).\\
- If multiple candidates appear, choose the last one that is self-consistent.\\
- Ignore exploratory or contradicted intermediate guesses.\\
3) Output EXACTLY in the format: \texttt{\textbackslash boxed\{{FINAL\_ANSWER\}}}\\
Formatting rules:\\
- Put ONLY the final answer inside \texttt{\textbackslash boxed\{\}} (no units, words, or explanations).\\
- Do not include any explanation or extra symbols outside \texttt{\textbackslash boxed\{\}}.\\
- If no conclusive final answer is present in the trace, choose the last consistent candidate stated as final; if still impossible, output \texttt{\textbackslash boxed\{NO\_ANSWER\}}.\\
Output:
\end{tcolorbox}

\begin{tcolorbox}[
  colback=gray!5,
  colframe=gray!80!black,
  colbacktitle=gray!80!black,
  coltitle=white,
  title=MMLU-ProX-Lite Answer Extraction Prompt,
  enhanced,
  breakable,
  width=\linewidth,
  boxsep=4pt
]
You are an answer extractor.\\[4pt]
You will be provided with the following inputs:\\
- Multiple-choice options (corresponding to the Question)\\
- A reasoning trace that shows the step-by-step thought process\\[4pt]
Task:\\
1) Carefully scan the Reasoning trace and identify the final multiple-choice option answer.\\
- Valid answers are only single capital letters from [A-J].\\
- If the final answer in the Reasoning trace is given as option text instead of a letter, use the provided multiple-choice options to map it to the corresponding letter from [A-J].\\
- Prefer the final/most conclusive statement (e.g., "Therefore...", "Thus...", "Final answer...", or the last decisive choice).\\
- If multiple candidates appear, choose the last consistent one.\\
- Ignore exploratory or contradicted intermediate guesses.\\[4pt]
2) Output EXACTLY in the format:\\
Answer: X\\[4pt]
Formatting rules:\\
- Replace X with the chosen letter from [A-J].\\
- Do not include any explanation, units, or extra text.\\[4pt]
Now, the inputs are given below.\\[4pt]
Inputs:\\
- Multiple-choice options (corresponding to the Question): \{options\_block\}\\
- Reasoning trace: \{reasoning\_trace\}\\[4pt]
Output:
\end{tcolorbox}
\paragraph{Answer Verification.}
We evaluate the correctness of extracted answers using \textsc{Math-Verify}~\cite{kydlicek2025mathverify},
a robust mathematical expression evaluation system developed for evaluating Large Language Model outputs in mathematical tasks. For MMLU-ProX-Lite, correctness is evaluated by comparing the extracted and ground-truth answers—both in option format—using case-insensitive exact string matching.

\section{Further Details on Understanding Failure Detection}\label{appendix:uf_details}
\subsection{Dataset statistics}
Table~\ref{tab:dataset_stats} summarizes the number of samples used for calibration and evaluation in our experiments. For methods requiring threshold calibration or supervised training, we use MGSM (filtered) as the calibration data for Polymath-Low experiments and the validation split of MMLU-ProX-Lite for MMLU-ProX-Lite experiments, ensuring that the calibration data remains \textit{in-distribution} with respect to the evaluation set. 
For Polymath-Low, 125 MGSM samples per language are used as evaluation data, and we exclude these from the calibration set; MGSM (filtered) thus refers to the remainder of the MGSM dataset after this exclusion. For MMLU-ProX-Lite, we restrict the validation split to STEM-related subjects, consistent with the composition of the test split.

\begin{table*}[!t]
\centering
\resizebox{\textwidth}{!}{
\begin{tabular}{lccc|cc}
\toprule
\textbf{Model} & \textbf{Seed} & \textbf{MGSM (filtered)} & \textbf{MMLU-ProX-Lite (validation)} & \textbf{Polymath-Low} & \textbf{MMLU-ProX-Lite (test)} \\
\midrule
\multirow{3}{*}{Qwen3-4B} 
& 32 & 1184 & 516 & 1166 & 2091 \\
& 42 & 1189 & 520 & 1163 & 2097 \\
& 52 & 1187 & 514 & 1170 & 2094 \\
\midrule
\multirow{3}{*}{gpt-oss-20b} 
& 32 & 1168 & 590 & 1191 & 2112 \\
& 42 & 1173 & 582 & 1183 & 2083 \\
& 52 & 1173 & 585 & 1184 & 2106 \\
\bottomrule
\end{tabular}
}
\caption{Number of samples used for calibration and evaluation across datasets and random seeds. \textbf{MGSM (filtered)} denotes the MGSM calibration set after excluding the 125 samples per language used in Polymath-Low evaluation.}
\label{tab:dataset_stats}
\end{table*}

\subsection{Implementation details}
\paragraph{LLM-based detector.}
We employ \texttt{GPT-4.1-mini}~\cite{openai2025gpt4.1} as an LLM-based detector due to its strong multilingual understanding ability and cost-effectiveness. On the Multilingual MMLU\footnote{This is obtained by translating MMLU~\cite{hendryckstest2021} benchmark's test set into 14 languages.} benchmark, GPT-4.1-mini demonstrates strong multilingual performance with an accuracy of 78.5\%, closely approaching GPT-4o~\cite{hurst2024gpt} (81.4\%) while being significantly more cost-effective. We prompt the detector with the original input and the reasoning model’s reasoning trace at the base setting to output a binary judgment on whether the model has understood the problem, along with a brief explanation. We then parse the structured JSON output into a binary label for downstream evaluation. The prompt provided to the detector is shown below:

\begin{tcolorbox}[colback=gray!5,colframe=gray!70!black,title=Prompt for LLM-based detector]
You are given a problem (question and possibly options) and a model’s reasoning trace. \\
Your task is to decide whether the model correctly understood the problem. \\
Do not solve the problem yourself.

Return the output strictly in the following JSON format, with no extra text. \\
The "Reason" field should be one or two sentences. \\

\{
  "understood": true/false,
  "Reason": "<one or two sentences explanation of why you judged it this way>"
\} \\

Problem:
\{problem text\}

Reasoning Trace:
\{reasoning trace\}
\end{tcolorbox}

\paragraph{mmBERT detector.}
We also train a lightweight text-based classifier based on mmBERT~\cite{marone2025mmbert}, a state-of-the-art multilingual encoder trained on over 3T tokens across more than 1,800 languages. We adopt the mmBERT-base variant for its strong multilingual classification performance and parameter efficiency (307M). We aggregate samples from all languages and split the combined dataset into training and validation sets with a 9:1 ratio. Each training instance consists of the original input(problem statement) concatenated with the reasoning model’s reasoning trace at the base setting, separated by the special \texttt{[SEP]} token. Labels are assigned as $1$ when the base model’s final answer is incorrect (\textit{not understood}) and $0$ otherwise. 

We fine-tune the model for 30 epochs with a batch size of 8 and a learning rate of $3 \times 10^{-5}$ using supervised fine-tuning. To address label imbalance, we adopt a weighted cross-entropy loss where class weights are set inversely proportional to class frequencies. Concretely, for a dataset $\{(x_i, y_i)\}_{i=1}^N$ with $n_1$ positive and $n_0$ negative samples ($N = n_0 + n_1$), the class weights are defined as:
\[
w_1 = \frac{N}{2 n_1}, \quad w_0 = \frac{N}{2 n_0}
\]
This inverse-frequency weighting ensures that both classes contribute equally to the loss regardless of their original proportions. The resulting objective is:
\[
\mathcal{L} = - \frac{1}{N} \sum_{i=1}^N \Big[ w_1\, y_i \log \hat{p}_i + w_0\, (1 - y_i) \log (1 - \hat{p}_i) \Big]
\label{eq:loss}
\]
where $y_i \in \{0,1\}$ is the ground-truth label and $\hat{p}_i = p_\theta(y_i = 1 \mid x_i)$ is the model’s predicted probability for the positive class. We minimize this loss using the AdamW optimizer~\cite{loshchilov2018decoupled}.
During training, we evaluate the model on the validation set after each epoch and select the checkpoint that achieves the highest F1 score. To prevent overfitting, we use early stopping with a patience of five epochs, stopping training when the validation F1 score fails to improve within this period. All experiments were run on a single A100-80GB GPU, with total training time of about one hour.

\paragraph{Token-probability–based approaches.} 
Given an input prompt $x = (x_1, \dots, x_{L_p})$ and a reasoning trace $r = (r_1, \dots, r_{L_r})$, let $p_\theta$ denote the language model’s conditional probability distribution over the next token given the previous tokens. The log-probability of the ground-truth next token $y_t$ at position $t$ is then defined as:
\[
\ell_t = \log p_\theta(y_t \mid y_{<t})
\]

We define the \textit{token confidence} $c_t$ at each step as the negative log-probability averaged over the top-$K$ most probable tokens (default $K = 20$):
\[
c_t = - \frac{1}{K} \sum_{k=1}^K \log p_\theta(w_k^{(t)} \mid y_{<t})
\]
where $w_k^{(t)}$ denotes the $k$-th most likely token predicted at position $t$. Higher $c_t$ values indicate a more peaked next-token distribution and thus higher certainty, whereas smaller $c_t$ values indicate a more uniform distribution and lower certainty. Based on these token confidences, we consider two detection methods:
\begin{itemize}
    \item \textbf{Average Confidence Method.} Computes the mean token confidence across all reasoning tokens:
    \[
    \text{avg\_confidence} = \frac{1}{L_r} \sum_{t=L_p+1}^{L_p+L_r} c_t
    \]
    \item \textbf{Minimum Confidence Method.} Takes the minimum token confidence across the reasoning trace:
    \[
    \text{min\_confidence} = \min_{t \in [L_p+1,\, L_p+L_r]} c_t
    \]
\end{itemize}

Both methods estimate how confidently the model generates its reasoning steps. Smaller values indicate lower confidence, which we hypothesize to be associated with a higher likelihood of understanding failure.

Finally, we define the \textit{input negative log-likelihood}, which measures how familiar the model is with the input prompt. It is computed as the mean negative log-probability over all input tokens:
\[
\text{Input\_NLL} = - \frac{1}{L_p} \sum_{t=1}^{L_p} \log p_\theta(x_t \mid x_{<t})
\]
A higher input negative log-likelihood suggests that the model assigns lower likelihood to the input sequence itself, which we hypothesize to indicate potential difficulty in understanding the input language or phrasing.

\begin{figure}[!t]
    \centering
    \begin{subfigure}[t]{0.48\textwidth}
        \centering
        \includegraphics[width=\textwidth]{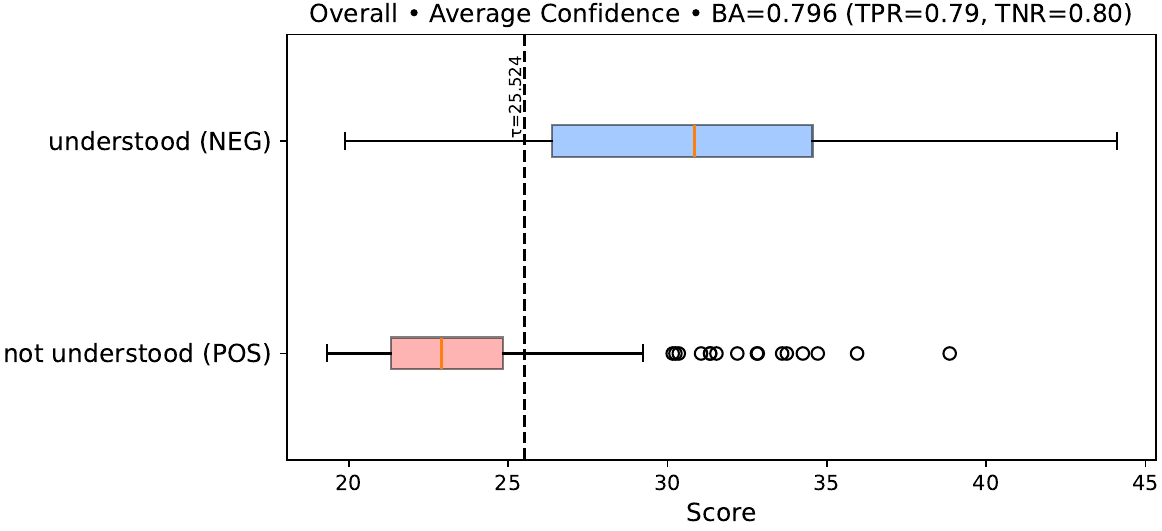}
        \caption{Average token confidence.}
        \label{fig:avg_confidence_boxplot}
    \end{subfigure}
    \vspace{0.6cm} % vertical spacing between subfigures

    \begin{subfigure}[t]{0.48\textwidth}
        \centering
        \includegraphics[width=\textwidth]{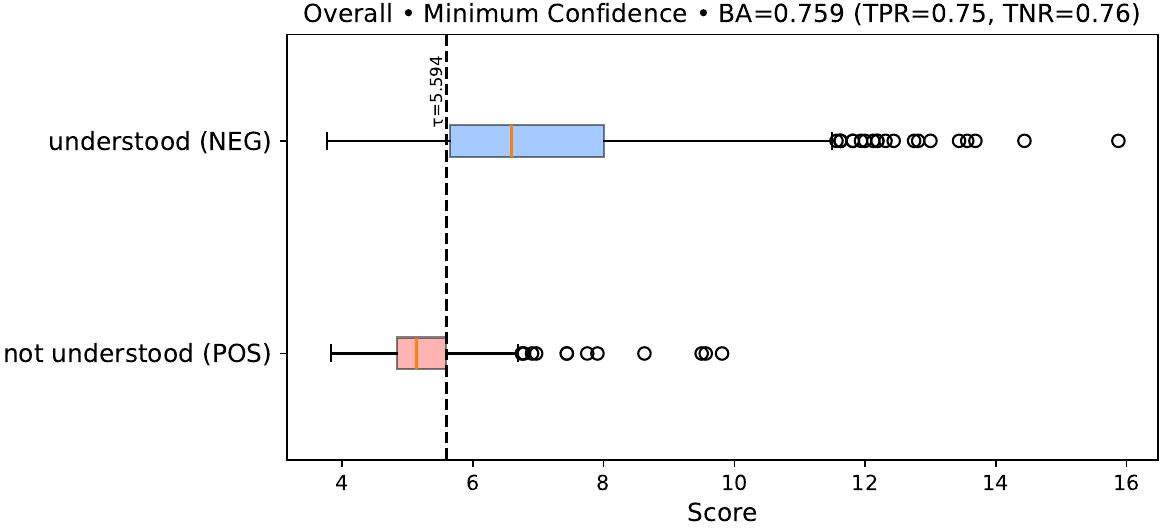}
        \caption{Minimum token confidence.}
        \label{fig:min_confidence_boxplot}
    \end{subfigure}
    \vspace{0.6cm}
    \begin{subfigure}[t]{0.48\textwidth}
        \centering
        \includegraphics[width=\textwidth]{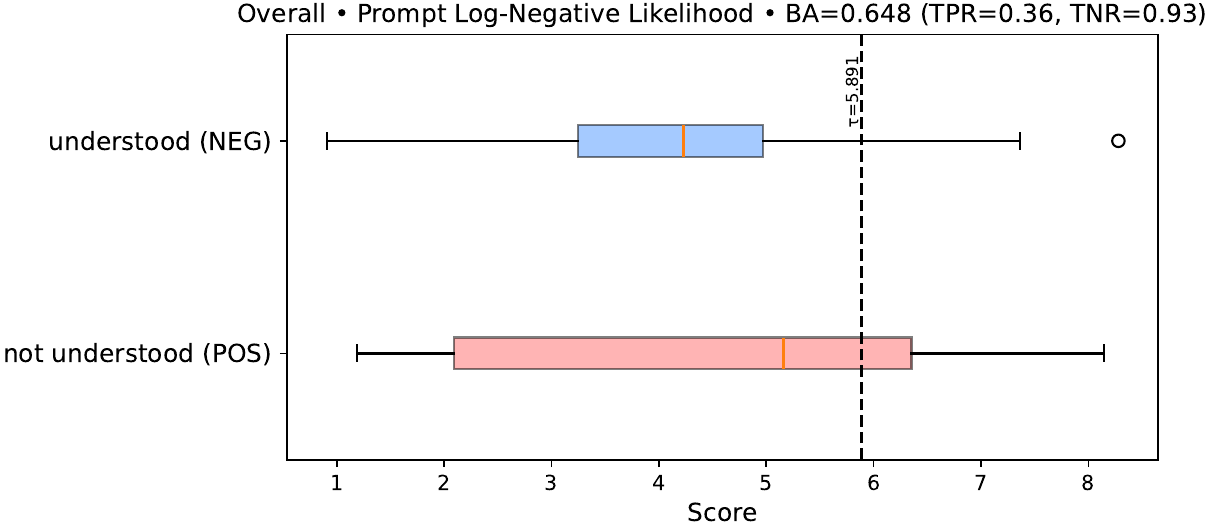}
        \caption{Input negative log-likelihood.}
        \label{fig:prompt_ln_nll_boxplot}
    \end{subfigure}
    \caption{
Distributions of three token-probability–based signals — average confidence, minimum confidence, and input negative log-likelihood — measured by Qwen3-4B on the Polymath-Low dataset. Each box plot illustrates how these signals differ between the \textit{understood} (label = 0) and \textit{not understood} (label = 1) classes, and the vertical line indicates the threshold derived from the calibration dataset.}
    \label{fig:probability_signals_vertical}
\end{figure}

For evaluation, we determine the decision threshold for each signal using the value that maximizes the F1 score on the calibration dataset. Figure~\ref{fig:probability_signals_vertical} visualizes the distributions of these three signals for understood (label = 0) and not-understood (label = 1) samples. As hypothesized, \textit{average confidence} and \textit{minimum confidence} tend to be lower for not-understood samples, indicating their usefulness as understanding-failure signals. In contrast, \textit{input negative log-likelihood} shows no clear correlation with the labels.

\paragraph{Hidden-state--based prober.}
We additionally explore a hidden-state--based approach that trains a lightweight classifier (``prober'') on the internal representations of the reasoning model to detect understanding failures. For each sample, we extract the final-layer hidden state of the last token from the model’s reasoning trace under the \textbf{Base} setting. These hidden representations, paired with binary labels ($1$ for ``not understood'' and $0$ otherwise), constitute the training data for the prober. We aggregate samples across all languages and split the dataset into training and development sets with a 9:1 ratio. The prober is trained to minimize the weighted binary cross-entropy loss (Eq.~\ref{eq:loss}) using the AdamW optimizer~\cite{loshchilov2018decoupled}:
\[
\mathcal{L} = - \frac{1}{N} \sum_{i=1}^N \Big[ w_1\, y_i \log \hat{p}_i + w_0\, (1 - y_i) \log (1 - \hat{p}_i) \Big]
\]
where $y_i \in \{0,1\}$ is the ground-truth label, $\hat{p}_i = p_\theta(y_i = 1 \mid h_i)$ is the prober’s predicted probability of a sample being not understood, and $h_i$ denotes the hidden representation of the $i$-th sample.

We perform a grid search over learning rates $\{10^{-3}, 10^{-4}, 10^{-5}\}$ and hidden layer sizes $\{0, 32, 128, 512, d/2, d\}$, where $d$ is the dimensionality of the original hidden state and $0$ corresponds to a linear probe. Training is conducted with a batch size of 16 for up to 50 epochs, with early stopping using a patience of five epochs, terminating training if the validation F1 score does not improve within this window. We select the checkpoint that achieves the highest F1 score on the development set for evaluation. 
The training was conducted on a single RTX 3090 GPU and completed in under one hour in total.
\paragraph{Self-reflection.}
We further evaluate the self-reflection method, where the reasoning model is prompted to assess whether it has understood the input. For each sample, we reuse the original input and reasoning trace. Then, we remove the end-of-thinking token (e.g., \texttt{</think>}) and append a fixed instruction for self-reflection, which allows the model to continue from its prior reasoning and generate a binary self-assessment. Generation is performed using greedy decoding. The instruction for self-reflection is shown below.
\begin{tcolorbox}[colback=gray!5,colframe=gray!70!black,title=Instruction for self-reflection]
Wait, before proceeding, I will reflect on my prior reasoning to assess my overall 
understanding of the problem. I will respond with <Understandability>: YES or NO 
(YES if I'm fully confident that I understood the problem correctly, NO otherwise).\\
<Understandability>:
\end{tcolorbox}

\section{Further Details on Selective Translation}\label{sec:appendix_selective_translation}
\paragraph{Translation.} We use GPT-4.1 for translation in all experiments. The prompt used for translating Polymath is shown below:
\begin{tcolorbox}[
  colback=gray!5,
  colframe=gray!70!black,
  title=Instruction for Polymath,
  breakable,
  enhanced jigsaw
]
Translate the following mathematical question enclosed within <instruction> and </instruction> into English.  
The text may contain mathematical notation and LaTeX formatting. You must strictly preserve:  \\
- All LaTeX math and commands EXACTLY as written, including inline math \texttt{\$...\$}, display math \texttt{\$\$...\$\$}, \texttt{\textbackslash(...\textbackslash)}, \texttt{\textbackslash[...\textbackslash]}, and any \texttt{\textbackslash begin\{...\}...\textbackslash end\{...\}} environments. \\
- All mathematical symbols, variables, numbers, operators, and equation labels.  \\

Provide only the translated instruction without any additional explanation.  \\
Wrap the translated output with <translated> and </translated> tags.  \\
\\
<instruction>\{instruction\}</instruction>
\end{tcolorbox}

For MMLU-ProX-Lite, each instance is represented as a JSON object containing a question and multiple-choice options (\texttt{option\_0}, \texttt{option\_1}, ...).  
We first construct a \textit{payload} that includes the question and all available options (0–9).  
The translator model is then instructed to translate values in this JSON—while preserving the keys, structure, and mathematical content—into English. The prompt used for translating MMLU-ProX-Lite is shown below:

\begin{tcolorbox}[colback=gray!5,colframe=gray!70!black,title=Instruction for MMLU-ProX-Lite]
You are a professional scientific translator.\\

TASK\\
- Translate ONLY the string **values** in the given JSON object into English.\\
- Do not change the JSON keys, structure, or order.\\
- Preserve all numbers, mathematical expressions, symbols, and units exactly.\\
- Return ONLY the translated JSON object.\\
\\
INPUT:
\{payload\}
\end{tcolorbox}

\paragraph{Selective translation.} 
After translation, the translated input is provided to the reasoning model via the understanding intervention  (Section~\ref{sec:method_gap}), where it corresponds to $x_{\text{dom}}$. We adopt this approach to preserve multilingual consistency, ensuring that the final response remains in the same language as the input.
Understanding intervention allows the model to reason without understanding failures while keeping the final response in the input language. We also experimented with a simpler variant that, instead of applying the understanding intervention, directly feeds the translated input to the model along with an additional instruction, “Respond in X,” where X is the input language, but it underperformed in our preliminary experiments. Therefore, we use the understanding-intervention formulation as the default setting for selective translation.
\section{More Related Work}\label{sec:appendix_additional_related_work} To complement the discussion in the main text, this section extends our related work section to include works on multilingual reasoning in conventional large language models and recent advances in multilingual abstention, further positioning our work within the broader literature of multilingual reasoning and abstention.
\paragraph{Multilingual Reasoning.} LLMs have made significant advances in multilingual reasoning, but performance gaps remain between high-resource and low-resource languages \citep{shilanguage}. To bridge this gap, \citet{qin-etal-2023-cross} exploited cross-lingual prompting to leverage strong reasoning abilities in English, while other works \citep{chen-etal-2024-breaking, zhu-etal-2024-question, ko2025understand} improved multilingual reasoning by fine-tuning on data generated through translations. \citet{she-etal-2024-mapo} and \citet{yanglanguage} exploited the inherent performance imbalance between high- and low-resource languages to construct preference datasets and applied preference optimization. Some prior works introduced external components to boost multilingual reasoning. \citet{yoon-etal-2024-langbridge} and \citet{huang2024mindmerger} merged LLMs with external multilingual models to enhance language understanding, and \citet{shilanguage} and \citet{liu-etal-2025-translation} translated questions posed in low-resource languages into high-resource languages to bypass challenges in language understanding. Among recent works, \citet{ko2025understand} report a similar finding that the performance gap in multilingual reasoning largely stems from models’ difficulty in comprehending non-English inputs, rather than their reasoning ability. However, their study focuses on conventional LLMs and is limited to Korean, whereas we analyze recent reasoning language models across diverse languages. We further show that such comprehension failures can be automatically detected and selectively mitigated through Selective Translation.
\paragraph{Abstention in multilingual setting.}
Abstention enables language models to refuse to answer when uncertain or inappropriate, reducing errors and hallucinations~\cite{tomani2024uncertainty,wen2025know,kirichenko2025abstentionbench}. Approaches include token-level confidence estimation~\cite{jiang2021can}, self-reflection~\cite{kadavath2022language}, probing~\cite{azaria2023internal}, and multi-agent collaboration~\cite{feng2024don}, though most have focused on monolingual or English-centric settings. Recently, \citet{feng2024teaching} presented the first study of multilingual abstention, proposing multilingual feedback to address cross-lingual knowledge disparities. However, their focus is on identifying whether the model possesses sufficient \textit{knowledge} in that language. In contrast, our work targets detecting \textit{understanding} failures, and leverages the results to trigger selective translation.

% For Appendix A 

% Figure for gpt-oss-20b

% Figure for Qwen3-1.7B

% Figure for Qwen3-8B

% Figure for Qwen3-14B

\end{document}